\documentclass[10pt,twocolumn,letterpaper]{article}

\usepackage[pagenumbers]{cvpr} 

\usepackage{xcolor}

\definecolor{cvprblue}{rgb}{0.21,0.49,0.74}
\usepackage[pagebackref,breaklinks,colorlinks,allcolors=cvprblue]{hyperref}
\usepackage{tikzducks}
\usepackage{bm}
\usepackage{multirow}
\usepackage{array}

\usepackage{lipsum}  
\usepackage{caption}
\usepackage{subcaption}
\usetikzlibrary{calc,arrows,arrows.meta,tikzmark,shapes.geometric, decorations.pathreplacing,decorations.markings, spy,matrix,positioning,shadows.blur, shapes.misc,}
\usepackage{siunitx}
\usepackage{mathtools,xparse}
\usepackage{bm}

\pgfdeclarelayer{background} 
\pgfdeclarelayer{shadow} 
\pgfsetlayers{background,shadow,main}

\makeatletter
\newcommand{\definetrim}[2]{%
  \define@key{Gin}{#1}[]{\setkeys{Gin}{trim=#2,clip}}%
}
\makeatother
\usepackage[export]{adjustbox}

\def\paperID{7371} 
\def\confName{CVPR}
\def\confYear{2025}

\newcommand{\sysname}{DiffLocks}

\title{\sysname: Generating 3D Hair \\from a Single Image using Diffusion Models}

\definecolor{net-bg}{RGB}{245, 177, 95} 
\definecolor{net-layer}{RGB}{145, 189, 179} 
\definecolor{net-latent}{RGB}{141, 235, 217} 

\definecolor{ph-purple}{RGB}{129, 39, 232}
\definecolor{ph-blue}{RGB}{5, 131, 227}
\definecolor{ph-gray}{rgb}{0.5, 0.5, 0.5}
\definecolor{ph-orange}{RGB}{227, 127, 5}
\definecolor{ph-green}{RGB}{0, 135, 124}
\definecolor{ph-yellow}{RGB}{235, 201, 52}
\definecolor{ph-light-green}{RGB}{181, 209, 21}
\definecolor{ph-red}{RGB}{250, 101, 60}
\definecolor{ph-gold}{RGB}{255, 223, 0}
\definecolor{ph-metallic-gold}{RGB}{212, 175, 55}
\definecolor{ph-silver}{RGB}{192, 192, 192}
\colorlet{g}{ph-gold!90!black}
\colorlet{s}{ph-silver!60}
\colorlet{ph-orange-light}{ph-orange!70}
\colorlet{ph-blue-light}{ph-blue!70}
\colorlet{ph-purple-light}{ph-purple!70}
\colorlet{ph-green-light}{ph-green!70}
\definecolor{ph-light-gray}{rgb}{0.75, 0.75, 0.75}

\DeclarePairedDelimiter{\abs}{\lvert}{\rvert}
\DeclarePairedDelimiter{\norm}{\lVert}{\rVert}

\newcommand{\Loss}{\mathcal{L}}
\newcommand{\pos}{\mathbf{p}}
\newcommand{\strand}{\mathbf{S}}
\newcommand{\length}{\emph{L}}
\newcommand{\direction}{\mathbf{d}}
\newcommand{\curv}{\bm{\kappa}}
\newcommand{\encoder}{\mathcal{E}}
\newcommand{\generator}{\mathcal{G}}
\newcommand{\latent}{\mathbf{z}}
\newcommand{\scalptexture}{\mathbf{T}}
\newcommand{\density}{\mathbf{D}}
\newcommand{\Hair}{\mathbf{H}}
\newcommand{\denoiser}{\mathcal{D}_\theta}
\newcommand{\denoiserInner}{\mathcal{F}_\theta}
\newcommand{\noise}{\sigma}
\newcommand{\cin}{c_\textrm{in} }
\newcommand{\cout}{c_\textrm{out}}
\newcommand{\cskip}{c_\textrm{skip}}
\newcommand{\cnoise}{c_\textrm{noise}}

\newcommand{\featuremap}{\mathcal{F}_{map}}
\newcommand{\cls}{\mathcal{F}_{cls}}

\newcommand{\Real}{\mathbb{R}}

\newcommand{\R}{\mathbb{R}}

\newcommand{\E}{\mathbb{E}}

\author{
Radu Alexandru Rosu$^{1,}$\thanks{Joint first author\\\makebox[1.0em][l]{} \dag Work done during an internship at Meshcapade} \quad
Keyu Wu$^{2,\ast,\dag}$ \quad
Yao Feng$^{1,3}$ \quad
Youyi Zheng$^2$ \quad
Michael J. Black$^{4}$ \quad \\
\normalsize $^1$Meshcapade \quad
\normalsize $^2$Zhejiang University \quad
\normalsize $^3$Stanford University \quad
\normalsize $^4$Max Planck Institute for Intelligent Systems
}
\begin{document}
\maketitle
\begin{abstract}
We address the task of generating 3D hair geometry from a single image, which is challenging due to the diversity of hairstyles and the lack of paired image-to-3D hair data.
Previous methods are primarily trained on synthetic data and cope with the limited amount of such data by using low-dimensional intermediate representations, such as guide strands and scalp-level embeddings, that require post-processing to decode, upsample, and add realism. 
These approaches fail to reconstruct detailed hair, struggle with curly hair, or are limited to handling only a few hairstyles. 
To overcome these limitations, we propose \mbox{DiffLocks}, a novel framework that enables detailed reconstruction of a wide variety of hairstyles directly from a single image.
First, we address the lack of 3D hair data by automating the creation of the largest synthetic hair dataset to date, containing 40K hairstyles. 
Second, we leverage the synthetic hair dataset to learn an image-conditioned diffusion-transfomer model that generates accurate 3D strands from a single frontal image. 
By using a pretrained image backbone, our method generalizes to in-the-wild images despite being trained only on synthetic data.
Our diffusion model predicts a scalp texture map in which any point in the map contains the latent code for an individual hair strand. 
These codes are directly decoded to 3D strands without post-processing techniques.
Representing individual strands, instead of guide strands, enables the transformer to model the detailed spatial structure of complex hairstyles.
With this, DiffLocks can recover highly curled hair, like afro hairstyles, from a single image for the first time.
Data and code is available at~\url{https://radualexandru.github.io/difflocks/}.

\end{abstract}    
\section{Introduction}\label{intro}

Realistic 3D hair is a crucial component of digital humans \cite{chai2014reduced,chai2015high,hu2017avatar,hadap2007strands,li2015facial,wang2023neuwigs,saito2024relightable}, which have wide applications in games, media and entertainment. 
Accurate hair representation greatly enhances the character's realism and overall visual quality. 
Unfortunately, both generating realistic hair and reconstructing it from an image are challenging due to the complex 3D geometry and the wide variety of hairstyles.

\input{fig/teaser}

Since the release of the first 3D synthetic hair dataset \cite{hu2015single}, numerous methods for hair reconstruction have been proposed. 
Multi-view capture methods \cite{paris2004capture,zhang2018modeling,kuang2022deepmvshair,nam2019strand,zhou2024groomcap} and video-based approaches \cite{sklyarova2023neural_haircut,wu2024monohair,luo2024gaussianhair} are able to reconstruct high-fidelity 3D hair but 
are time-intensive and challenging to deploy due to their reliance on expensive capture equipment or strict capture conditions. 

In contrast, single-view methods \cite{chai2016autohair,saito20183d,yang2019dynamic,wu2022neuralhdhair,zheng2023hairstep}, which reconstruct 3D hair from a single image, are faster, more accessible, and user-friendly. 
These methods, however, often produce low-detail hair and handle only a restricted range of hairstyles. 
Furthermore, these methods struggle to handle male-pattern baldness and very curly hair such as afro-like hairstyles, which have a complex geometric structure (see \cite{HWU2024}).
Previous methods suffer from a lack of diverse and comprehensive hairstyle training datasets. 
To increase training set diversity, GroomGen \cite{zhou2023groomgen}, HAAR \cite{sklyarova2023haar} and Fake It Till You Make It \cite{fakeit} use artist-created hairstyles together with data augmentation techniques to train their methods. 
However, their datasets are not released, do not have paired RGB images, or the creation process requires significant manual effort. 

To cope with limited training data, previous approaches rely on several intermediate representations to simplify the problem.
Most prior methods rely on oriented filters to capture the structure of hair in images.
They also do not model hair strands directly but, instead, use a smaller number of guide strands and then upsample these to generate a hairstyle.
Additionally, several methods use variational autoencoding to compress the guide strands across the scalp into a low-dimensional representation.
All of these approaches are designed to cope with limtied training data but have two effects: (1) they limit the ability to model the complex spatial relationships between individual strands, and (2) they limit realism by working with low-dimensional representations.
Effectively, they work in a compressed low-dimensional space and then need to upsample the results and post-process them to increase realism.

\begin{figure*}

\newcommand\HSWidth{0.162}

\captionsetup[subfigure]{labelformat=empty}
     \centering
     \begin{subfigure}[b]{\HSWidth\textwidth}
         \centering
         \includegraphics[width=\textwidth]{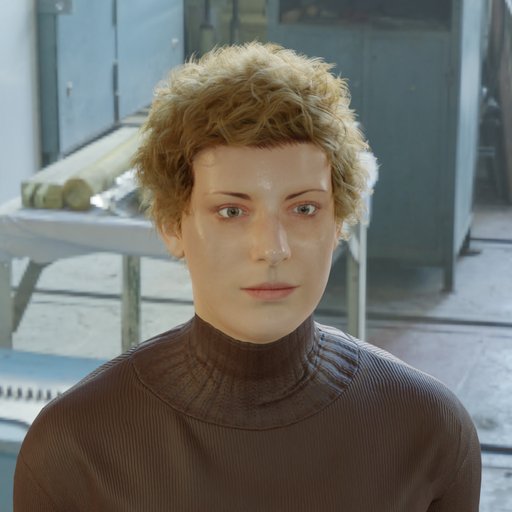}
     \end{subfigure}
     \begin{subfigure}[b]{\HSWidth\textwidth}
         \centering
         \includegraphics[width=\textwidth]{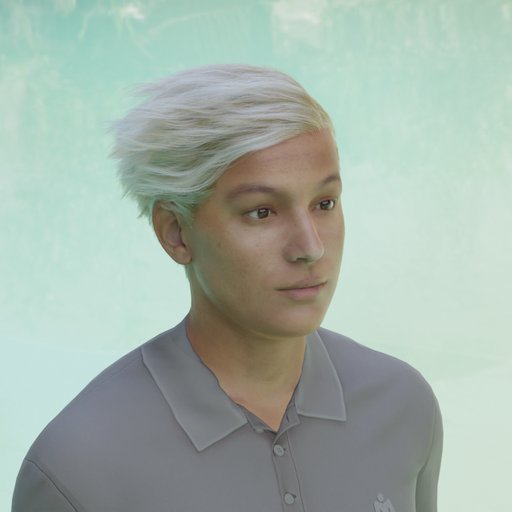}
     \end{subfigure}
     \begin{subfigure}[b]{\HSWidth\textwidth}
         \centering
         \includegraphics[width=\textwidth]{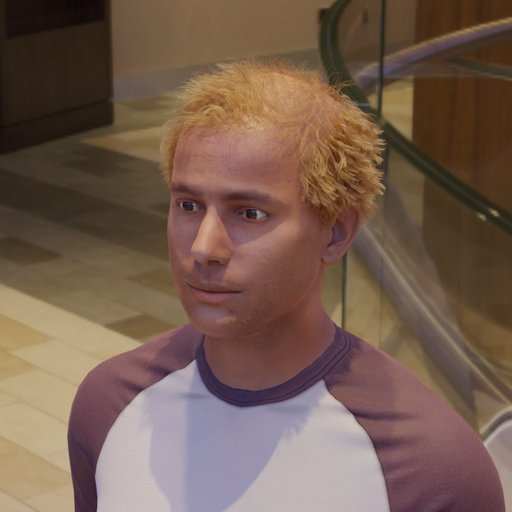}
     \end{subfigure}
     \begin{subfigure}[b]{\HSWidth\textwidth}
         \centering
         \includegraphics[width=\textwidth]{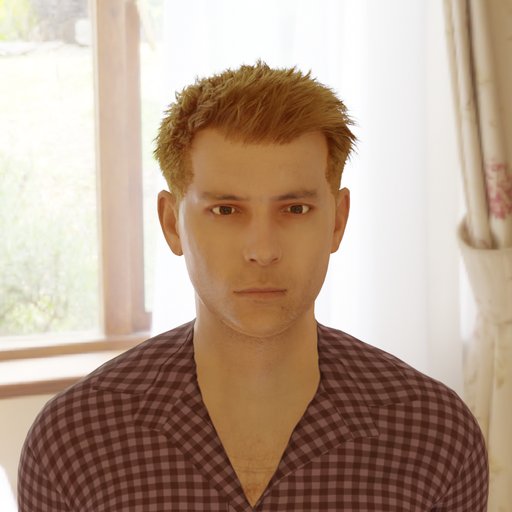}
     \end{subfigure}
     \begin{subfigure}[b]{\HSWidth\textwidth}
         \centering
         \includegraphics[width=\textwidth]{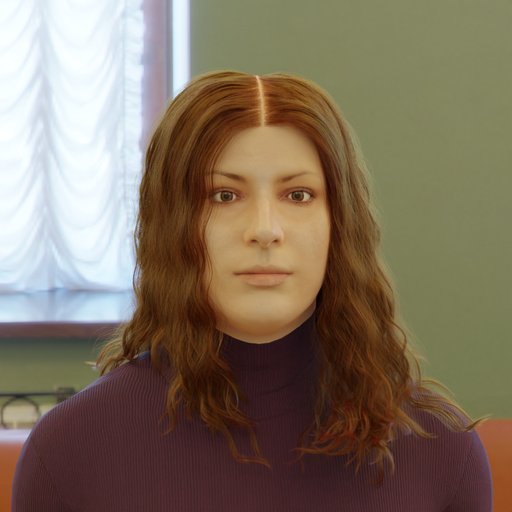}
     \end{subfigure}
     \begin{subfigure}[b]{\HSWidth\textwidth}
         \centering
         \includegraphics[width=\textwidth]{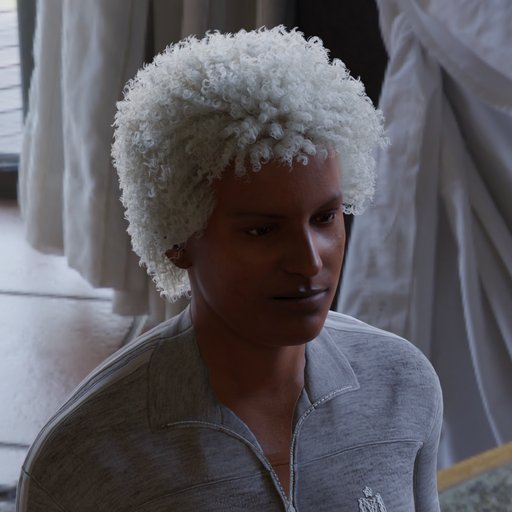}
     \end{subfigure}
     \vspace{-0.2cm}
    \caption{{\textbf{Synthetic data.} Sample RGB images from our synthetic hair dataset. 
    Each sample from the dataset contains an image at $768 \times 768$ resolution together with the corresponding 3D strand geometry of $\approx 100$K strands.} }
    \label{fig:hair_synth_samples}
\end{figure*}

We take a different approach and focus on creating a sufficient amount of training data that we can remove the simplifying assumptions of previous methods.
To that end,  we present a novel 3D synthetic hair dataset consisting of $40$K samples covering a wide variety of hairstyles; see Fig.~\ref{fig:hair_synth_samples}.
Each sample consists of 3D hair strands and a realistic RGB image rendered in Blender through path-tracing. 
We achieve this by automating the creation of the data using a variety of artist-inpsired techniques.
Traditional approaches are designed to create a single hairstyle using a small geometry node network tailored-made for that hairstyle.
In contrast, we design a large and very general geometry node network in Blender that allows us to automatically create a large number of realistic hairstyles by changing the different parameters.

\begin{figure*}[t]

    \def\shadowshift{5pt,-10pt}
    \def\shadowradius{10pt}
    \def\shadowxshift{-0.3ex}
    \def\shadowyshift{0.3ex}
    \def\shadowopacity{60}
    \def\shadowblurradius{0.3ex}

	\centering
	\begin{tikzpicture}[ remember picture, >={Stealth[inset=1pt,length=8pt,angle'=30,round]} ]

    \node[] (origin) at (0,0) {};

    \node[inner sep=0pt, anchor=east] (rgb_input) at (0,0) {\includegraphics[width=.12\textwidth]{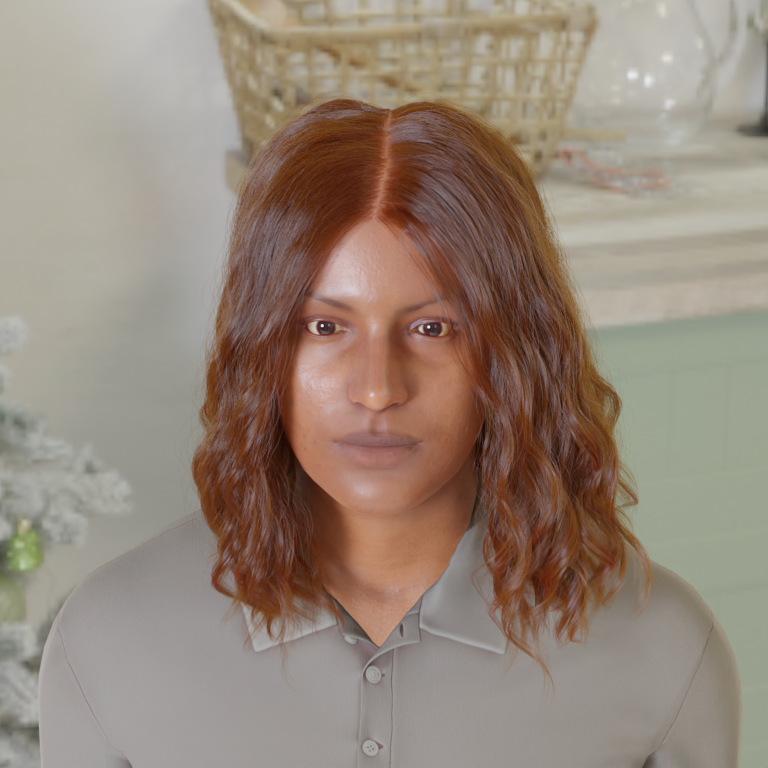}};
    \node[font=\footnotesize\selectfont, align=center, below = -0.0cm of rgb_input] (rgb_input_text) {RGB input};

    \node(dinonet)[anchor=west, name prefix = dino- ]
    at ($(rgb_input.east)+(-0.18cm,0)$){ 
\begin{tikzpicture}


    \newcommand\LayerW{0.1cm}
    \newcommand\LayerH{1.6cm}
    \newcommand\LayerConstriction{0.38cm}
    \newcommand\InterLayerSpace{0.1cm}
    \newcommand\TrapezoidMiddleXOffset{0.25cm}
    \newcommand\TrapezoidMiddleYOffset{0.25cm}
    \newcommand\TrapezoidEncXOffset{0.25cm}
    \newcommand\TrapezoidEncYOffset{0.25cm}



    \node[rectangle,
        very thin,
        minimum width=\LayerW,
        minimum height=\LayerH-\LayerConstriction*0] (enc1) at (-\InterLayerSpace*2,0) {};
    \node[rectangle, 
        very thin,
        minimum width=\LayerW,
        minimum height=\LayerH-\LayerConstriction*1] (enc2) at (-\InterLayerSpace*1,0) {};
    \node[rectangle,
        very thin,
        minimum width=\LayerW,
        minimum height=\LayerH-\LayerConstriction*2] (middle) at (-\InterLayerSpace*0,0) {};


    \node[circle,above right=\TrapezoidMiddleYOffset and  \TrapezoidMiddleXOffset of middle.north, anchor=center] (middle_top) {};
    \node[circle,below right=\TrapezoidMiddleYOffset and  \TrapezoidMiddleXOffset of middle.south, anchor=center] (middle_bottom) {};
    \node[circle,above left=\TrapezoidEncYOffset and \TrapezoidEncXOffset of enc1.north, anchor=center] (enc_top) {};
    \node[circle,below left=\TrapezoidEncYOffset and \TrapezoidEncXOffset of enc1.south, anchor=center] (enc_bottom) {};



    \begin{pgfonlayer}{background}
         \path [fill=net-bg,draw, thin] (enc_top.center) -- (middle_top.center) -- (middle_bottom.center) --  (enc_bottom.center) -- (enc_top.center);
     \end{pgfonlayer}

     \node[text=black, anchor=center] (dinotext) at (enc2.center) {\tiny DINOv2};


\end{tikzpicture}
	};

    \node[inner sep=0pt, anchor=west] (dino-latent) at ($(dinonet.east)+(-0.26cm,0.15cm)$) {\includegraphics[width=.06\textwidth]{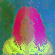}};
    \node[inner sep=1pt, draw, rectangle, anchor=north west, fill=teal, text=white] (dino-cls) at ($(dino-latent.south west)+(-0.0cm,-0.05cm)$) {\tiny CLS token};

    \node[inner sep=0pt, anchor=east, blur shadow= {shadow xshift=\shadowxshift, shadow yshift=\shadowyshift, shadow opacity=\shadowopacity, shadow blur radius=\shadowblurradius, 
    shadow blur extra rounding,
    shadow blur steps=8},] (scalp-dens-noisy) at ($(dinonet.east)+(2.55cm,0.45cm)$)  {\includegraphics[width=.065\textwidth]{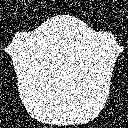}};
    \node[inner sep=0pt, anchor=east, blur shadow= {shadow xshift=\shadowxshift, shadow yshift=\shadowyshift, shadow opacity=\shadowopacity, shadow blur radius=\shadowblurradius, 
    shadow blur extra rounding,
    shadow blur steps=8},] (scalp-noisy) at ($(dinonet.east)+(3.0cm,0.23cm)$)  {\includegraphics[width=.065\textwidth]{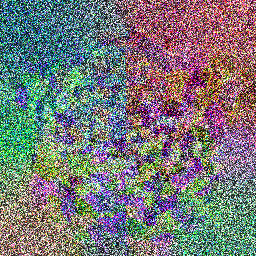}};
    
    \node[inner sep=1pt, draw, fill=teal, anchor=east, text=white] (globalcond) at ($(scalp-noisy.east)+(0.0cm,-1.1cm)$) {
    \tiny Time};
    \node[inner sep=1pt, draw, fill=teal, anchor=east, text=white, font=\tiny\selectfont] (globalcond) at (dino-cls -| scalp-noisy.east) {Global Cond};
    \draw[line width=0.2mm,->] (dino-cls.east) -- (globalcond.west);

    \node(diffnet)[anchor=west, name prefix = diffusion- ]
    at (4.05, 0.0)
    { 
\begin{tikzpicture}


    \newcommand\LayerW{0.17cm}
    \newcommand\LayerH{1.3cm}
    \newcommand\LayerConstriction{0.4cm}
    \newcommand\InterLayerSpace{0.35cm}
    \newcommand\TrapezoidMiddleOffset{0.3cm}
    \newcommand\TrapezoidEncXOffset{0.35cm}
    \newcommand\TrapezoidEncYOffset{0.35cm}



    \node[rectangle,
        rounded corners=2pt,
        draw,
        very thin,
        fill=net-layer, 
        minimum width=\LayerW,
        minimum height=\LayerH-\LayerConstriction*0] (enc1) at (-\InterLayerSpace*2,0) {};
    \node[rectangle, 
        rounded corners=2pt,
        draw,   
        very thin,
        fill=net-layer, 
        minimum width=\LayerW,
        minimum height=\LayerH-\LayerConstriction*1] (enc2) at (-\InterLayerSpace*1,0) {};
    \node[rectangle,
        rounded corners=2pt,
        draw,
        very thin,
        fill=net-layer, 
        minimum width=\LayerW,
        minimum height=\LayerH-\LayerConstriction*2] (middle) at (-\InterLayerSpace*0,0) {};
    \node[rectangle,   
        rounded corners=2pt,
        draw,
        very thin,
        fill=net-layer,
        minimum width=\LayerW,
        minimum height=\LayerH-\LayerConstriction*1] (dec1) at (\InterLayerSpace*1,0) {};
    \node[rectangle, 
        rounded corners=2pt,
        draw,
        very thin,
        fill=net-layer, 
        minimum width=\LayerW,
        minimum height=\LayerH-\LayerConstriction*0] (dec2) at (\InterLayerSpace*2,0) {};


    \node[circle,above=\TrapezoidMiddleOffset of middle.north, anchor=center] (middle_top) {};
    \node[circle,below=\TrapezoidMiddleOffset of middle.south, anchor=center] (middle_bottom) {};
    \node[circle,above left=\TrapezoidEncYOffset and \TrapezoidEncXOffset of enc1.north, anchor=center] (enc_top) {};
    \node[circle,below left=\TrapezoidEncYOffset and \TrapezoidEncXOffset of enc1.south, anchor=center] (enc_bottom) {};
    \node[circle,above right=\TrapezoidEncYOffset and \TrapezoidEncXOffset of dec2.north, anchor=center] (dec_top) {};
    \node[circle,below right=\TrapezoidEncYOffset and \TrapezoidEncXOffset of dec2.south, anchor=center] (dec_bottom) {};



    \begin{pgfonlayer}{background}
         \path [fill=net-bg,draw, thin] (enc_top.center) -- (middle_top.center) -- (dec_top.center) -- (dec_bottom.center) -- (middle_bottom.center) --  (enc_bottom.center) -- (enc_top.center);
     \end{pgfonlayer}

\end{tikzpicture}
 
    };
    \node[font=\footnotesize\selectfont, align=center] (scalp-clean_text) at (diffnet |- rgb_input_text) {Diffusion Model};

    \node[inner sep=0pt, anchor=north west, blur shadow= {shadow xshift=\shadowxshift, shadow yshift=\shadowyshift, shadow opacity=\shadowopacity, shadow blur radius=\shadowblurradius, 
    shadow blur extra rounding,
    shadow blur steps=8},] (scalp-clean) at ($(diffnet.east)+(-0.2cm,1.0cm)$) {\includegraphics[width=.085\textwidth]{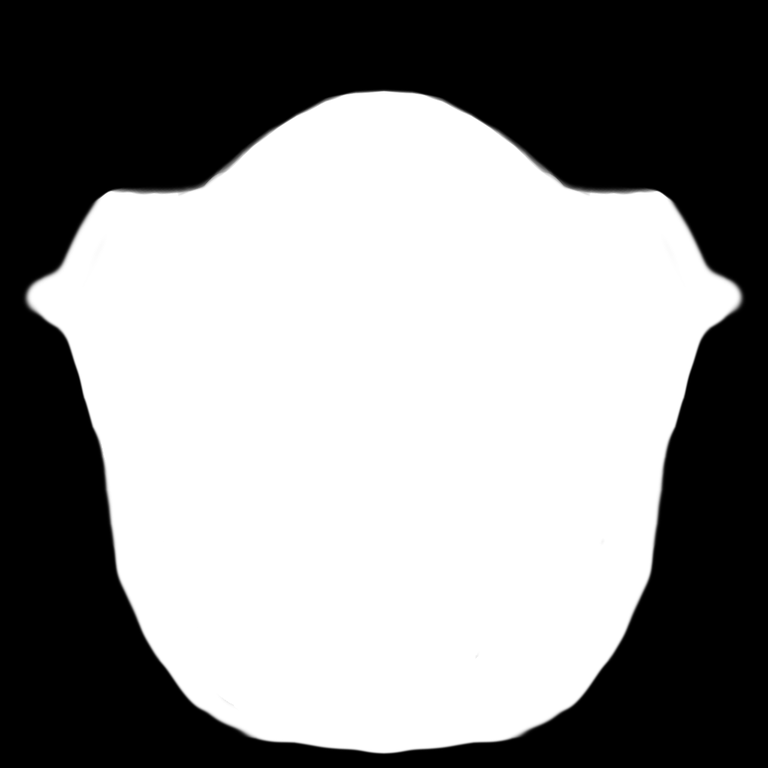}};
    \node[inner sep=0pt, anchor=south east, blur shadow= {shadow xshift=\shadowxshift, shadow yshift=\shadowyshift, shadow opacity=\shadowopacity, shadow blur radius=\shadowblurradius, 
    shadow blur extra rounding,
    shadow blur steps=8},] (scalp-clean) at ($(diffnet.east)+(1.8cm,-1.05cm)$) {\includegraphics[width=.085\textwidth]{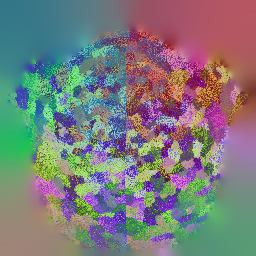}};
    \node[inner sep=0pt, font=\footnotesize\selectfont, align=center, xshift=-0.1cm, yshift=0.12cm, anchor=north] (scalp-clean_text) at (scalp-clean |- rgb_input_text) {Scalp Texture\\+Density};



    \def\yshiftstrands{0.2} 
    \node[inner sep=0pt] (scalp) at (10.0, \yshiftstrands) {\includegraphics[width=.15\textwidth]{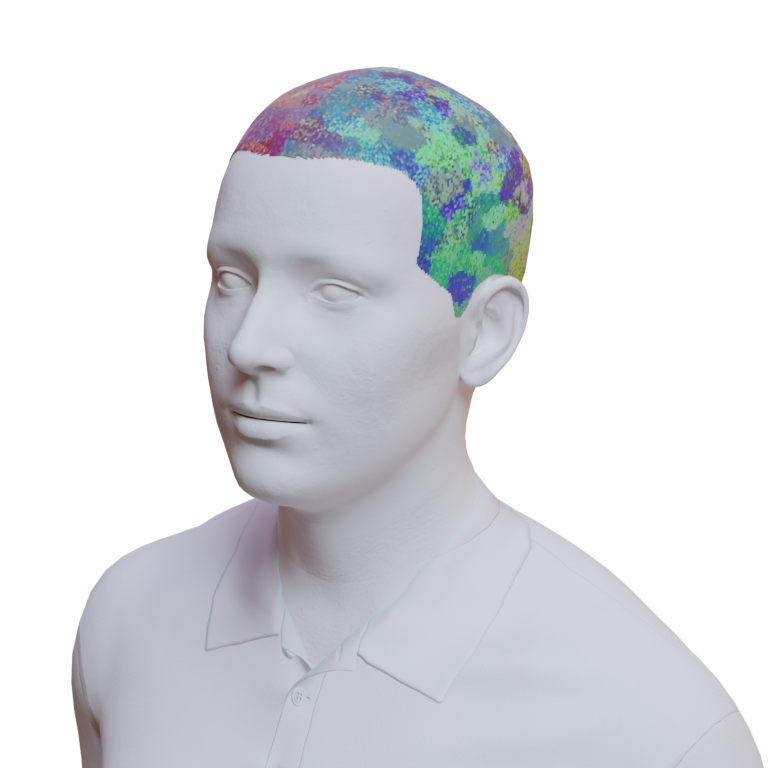}};
    \begin{scope} [ shift={(11.9, \yshiftstrands)}, name prefix = strandec- ]
	\input{fig/overview_2/strand_decoder}
	\end{scope}
    \node[inner sep=0pt] (single-strand) at (12.5, \yshiftstrands) {\includegraphics[width=.05\textwidth]{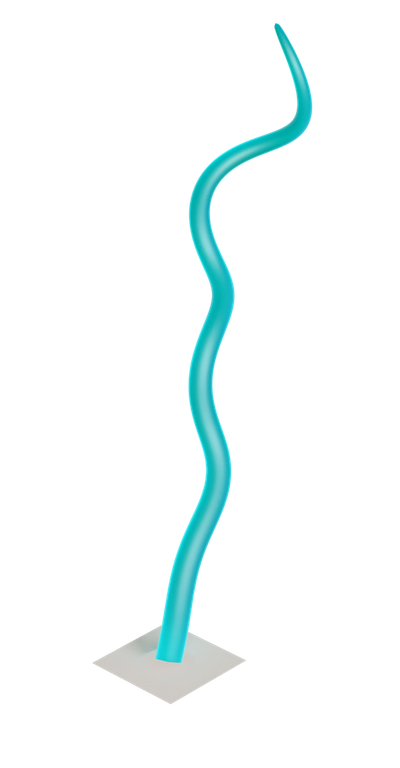}};
    \node[font=\tiny\selectfont, align=center, below = -0.2cm of single-strand] (single-strand_text) {Single\\Strand};

    \node[inner sep=0pt] (full-strands) at (14.0, \yshiftstrands) {\includegraphics[width=.15\textwidth]{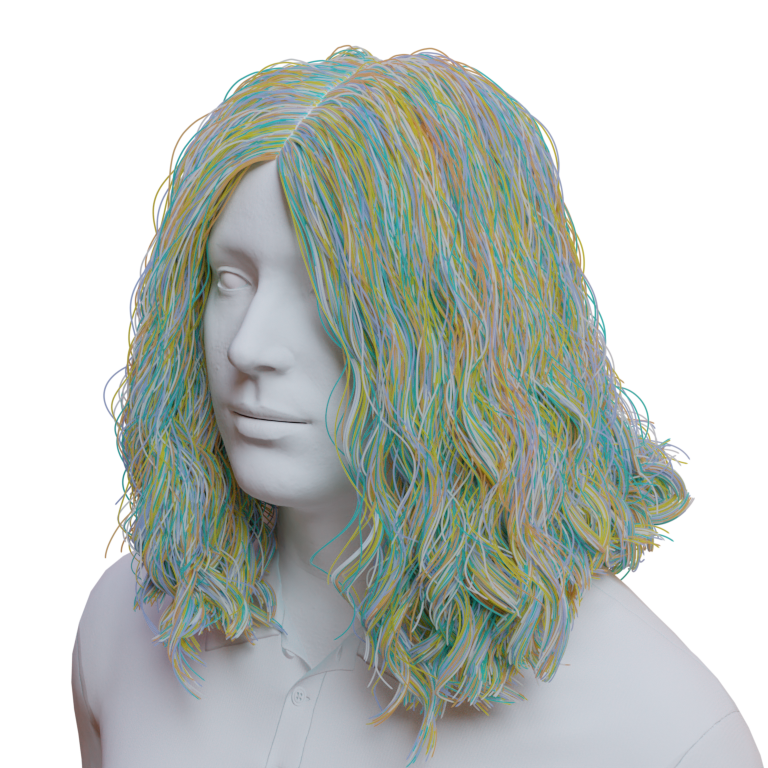}};
    \node[font=\footnotesize\selectfont, align=center] (full-strands_text) at (full-strands |- rgb_input_text) {3D hair strands};

     \begin{scope}
         \newcommand\CorssAttnDistAbove{1.3cm}
         \draw[line width=0.2mm,->]
          let
            \p1=(diffusion-enc1.north)
          in
            (\x1,\CorssAttnDistAbove) -- (diffusion-enc1.north);
        \draw[line width=0.2mm,->]
          let
            \p1=(diffusion-enc2.north)
          in
            (\x1,\CorssAttnDistAbove) -- (diffusion-enc2.north);
        \draw[line width=0.2mm,->]
          let
            \p1=(diffusion-middle.north)
          in
            (\x1,\CorssAttnDistAbove) -- (diffusion-middle.north);
        \draw[line width=0.2mm,->]
          let
            \p1=(diffusion-dec1.north)
          in
            (\x1,\CorssAttnDistAbove) -- (diffusion-dec1.north);
        \draw[line width=0.2mm,->]
          let
            \p1=(diffusion-dec2.north)
          in
            (\x1,\CorssAttnDistAbove) -- (diffusion-dec2.north);
        \draw[line width=0.2mm]
          let
            \p1=(diffusion-enc1.north),
            \p2=(diffusion-dec2.north)
          in
            (\x1,\CorssAttnDistAbove) -- (\x2,\CorssAttnDistAbove);

        \draw[line width=0.2mm]
          let
            \p1=($(dino-latent.north)+(0.0, 0.0)$),
          in
           (\x1,\CorssAttnDistAbove) -- ($(dino-latent.north) +(0.0, 0.0)$);

         \draw[line width=0.2mm]
          let
            \p1=(dino-latent.north),
            \p2=(diffusion-enc1.north)
          in
            (\x1,\CorssAttnDistAbove) -- (\x2,\CorssAttnDistAbove) node [pos=0.8, above, text=black, font=\tiny] {Cross-attention};
     \end{scope}


     \draw [->,orange] ([yshift=10.0,xshift=9.0]scalp-clean.north) to 
     [out=60,in=160] 
     node [pos=0.5, above, text=black, font=\tiny] {UV-mapping}
     ([yshift=30.0,xshift=25.0]scalp.west) ;

     \node[] (scalp-pixel) at ($(scalp.center) +(0.1, 0.8)$) {};
     \draw [->,orange] (scalp-pixel) to [out=30,in=110] (strandec-strand_latent.north);

     \draw [->,orange] ([yshift=0.0,xshift=-9.0]single-strand.north east) to [out=60,in=170] ([yshift=30.0,xshift=17.0]full-strands.west) node [above left, text=black, font=\tiny] {$\times 100$K};

     \draw [<->, loosely dashed, ultra thin, arrows={angle 90-angle 90}] (-2.0, 2.0) -- node[midway,fill=white] {\small RGB conditioning} (2.0, 2.0);
     \draw [<->, loosely dashed, ultra thin, arrows={angle 90-angle 90}] (2.5, 2.0) -- node[midway,fill=white] {\small Scalp Diffusion} (8.5, 2.0);
     \draw [<->, loosely dashed, ultra thin, arrows={angle 90-angle 90}] (9.0, 2.0) -- node[midway,fill=white] {\small Strand Decoding} (15.0, 2.0);

	\end{tikzpicture}%
    \vspace{-0.2cm}
	\caption{\textbf{Method.} Given a single RGB image, we use a pretrained DINOv2 model to extract  local and global features, which are used to guide a scalp diffusion model. The scalp diffusion model denoises a density map and a scalp texture containing latent codes for strand geometry. Finally, we probabilistically sample texels from the scalp texture and decode the latent code $\latent$ into strands of 256 points. Decoding in parallel \num{100}K strands yields the final hairstyle.}
    \vspace{-0.1cm}
	\label{fig:overview}
\end{figure*}

We then leverage our dataset to train a novel hair diffusion framework, called \textbf{DiffLocks}, that is optionally conditioned on single RGB image in order to create highly detailed 3D strands;
see Fig.~\ref{fig:teaser}.
Specifically, we use Hourglass Diffusion Transformers (HDiT)~\cite{hdit} as a diffusion architecture and exploit features from a pretrained DINOv2~\cite{dinov2} model as a conditioning input to guide and control the generated results; see Fig.~\ref{fig:overview}.
The DINOv2 features are richer than the oriented filters used in many previous methods. 
Note that we directly regress the latent code for individual hair strands (instead of guide strands) enabling the transformer to learn detailed spatial relationships between strands across the scalp.
With sufficient training data, there is no longer a need to first embed the hairstyle in a low-D scalp representation.
We also model a density map that defines the probability of a strand being generated at each location on the scalp.
All of these changes lead to increased realism, particularly for curly and balding hairstyles.
Finally, our strand-based hair can be directly used in real-time game engines like Unreal Engine (\cref{fig:hair_in_unreal}) without additional heuristics to densify the hair or add  fine-grained detail.

A key takeaway from our work is that, with sufficient training data, many of the key components of previous methods are no longer needed.
For example, in contrast to the $32 \times 32$ scalp map used by HAAR, our diffusion model predicts a scalp texture of size $256 \times 256$ (each pixel containing a latent code for a single strand) from which DiffLocks directly decodes 3D strands. 
No post-processing is needed, such as hair upsampling or the addition of random noise (cf.~\cite{zhou2023groomgen}). 
The dataset already contains realistic hairstyles with flyaway strands, so the diffusion model learns to create them.

In summary, our main contributions are:
	\begin{itemize}
		\item 
            a novel 3D hair dataset containing $40$K diverse and realistic 3D hairstyles with corresponding RGB images. We make the dataset public in order to support future research in hair modeling and synthesis,
        \item an improved scalp representation that facilitates the reconstruction of hair styles with different hair density (bald spots, male pattern baldness, etc.),
		\item a novel conditional scalp diffusion framework for robustly generating 3D hair directly from a single RGB image, bypassing a 2D orientation map. This framework can effectively create a variety of challenging hairstyles, while achieving the state-of-the-art performance and enable, for the first time, reconstruction of afro-like and bald hairstyles. The model will be available for research.

	\end{itemize}


\section{Related Works}

\noindent\textbf{3D Hair Reconstruction.}  The pioneering hair reconstruction work of Paris et al.~\cite{paris2004capture} set the stage for numerous multi-view-based hair reconstruction methods. 
Since then, various methods \cite{nam2019strand,kuang2022deepmvshair,luo2012multi,luo2013structure,zhou2024groomcap} reconstruct the 3D geometric features of hair constrained by a 2D orientation map. In many cases this requires specialized hardware or multi-camera rigs. 
In contrast, monocular video based methods \cite{sklyarova2023neural_haircut,wu2024monohair,luo2024gaussianhair,takimoto2024dr} sacrifice some accuracy to avoid expensive and specialized equipment.
While achieving impressive results, such methods still require carefully captured data.  
To relax the capture conditions, single-view-based methods \cite{chai2016autohair,yang2019dynamic,zhang2019hair,saito20183d,wu2022neuralhdhair,zhang2019hair} employ deep neural networks to infer volumetric fields by exploiting data priors learned from 3D synthetic datasets. 
Such methods enable rapid 3D hair reconstruction but typically generate low fidelity hair and can handle only a limited range of hairstyles due to the significant lack of diversity in existing 3D hair datasets. 
Similar to these methods, our approach also reconstructs hair from a single image. However, our work focuses on fast and robust reconstruction of diverse hairstyles, including challenging types such as balding and tightly curled (afro-like) styles; such hairstyles have received relatively little attention.

\smallskip\noindent\textbf{3D Hair Generation.} In addition to hair reconstruction, hair generation is another challenge, which, if solved, could provide synthetic data for learning high-quality hair reconstruction. 
Ren et al.~\cite{ren2021hair} propose a heuristic example-based hair generation method that is significantly limited by the reference hair models. Saito et al.~\cite{saito20183d} encode a 3D hair dataset with a Variational autoencoder (VAE), enabling them to generate plausible yet overly smooth hairstyles. To achieve high-quality and diverse hairstyles, Perm \cite{he2024perm} leverages existing public 3D datasets, disentangling the global shape and local details of hair strands. 
Despite progress, the scale and diversity of synthetic 3D hair datasets is still far from what is needed. GroomGen \cite{zhou2023groomgen} and HAAR \cite{sklyarova2023haar} incorporate parametrized variations based on high-quality hairstyles crafted by artists, further enhancing the diversity of 3D hair datasets.
They also train generative models to produce 3D hairstyles from noisy inputs or text prompts. 
Recently, Curly-Cue~\cite{HWU2024} focuses on the principles behind curly, afro-style, hair and provides tools for animators to create realistic hairstyles.
In our work, we propose a low-cost synthetic 3D hair generation pipeline and use it to construct a new dataset that is more diverse than previous ones,
containing 40K 3D synthetic hairstyles. 
Additionally, we use this dataset to train a generative hair model that is conditioned on a single input image.

\smallskip\noindent\textbf{Image-To-3D via Diffusion Models.}
Early work on neural reconstruction of 3D objects from images~\cite{neurallift, nerdi, realfusion} distills information from pretrained 2D diffusion models, such as  Stable Diffusion~\cite{stablediffusion}, using techniques like Score Distillation Sampling (SDS)~\cite{dreamfusion}. 
However, these pretrained models lack the multi-view information essential for accurate 3D reconstruction.  
To address this, later approaches~\cite{zero123, zero123++, one2345} train custom 2D diffusion models to generate multi-view images from a given input, enabling more accurate 3D object reconstruction by integrating this multi-view data~\cite{efficientdreamer, wonder3d}.
While effective, this approach can be time-intensive due to the sampling and reconstruction processes, which has led to an alternative line of research focused on training diffusion models to generate 3D representations directly, such as meshes~\cite{meshdiffusion} or triplane representations~\cite{chan2022efficient}. However, these methods often demand significant computational resources for 3D processing. 
In contrast, our method produces a 2D representation that contains 3D hair information, enabling direct decoding into 3D hair without the need for extensive multi-view generation or resource-intensive training of 3D diffusion models.

\section{Method}

In this section, we present our pipeline for 3D hair reconstruction from a single image, consisting of three key stages: 1) A data generation pipeline using Blender's geometry nodes to create a large, diverse, image-3D paired synthetic hair dataset (\cref{data generation}); 2) Parameterization of hair strands and hairstyles into a canonical scalp space to obtain scalp textures (\cref{parameterization}); And 3) training a conditional diffusion model, DiffLocks, using the generated synthetic images and the scalp textures as shown in \cref{fig:overview}.

\subsection{Data Generation}\label{data generation}

\input{fig/randomize_single_base_hairstyle}

A key bottleneck in the field of 3D hair reconstruction is the limited availability of public datasets, with only 343 hairstyles for USC-HairSalon \cite{hu2015single} and 10 for CT2Hair \cite{shen2023ct2hair}. Additionally, these datasets lack corresponding 2D images, which limits their usability.
Thus, here we propose a novel data generation pipeline, which uses Blender geometry nodes to add variation to a small set of manually defined guide strands as shown in \cref{fig:randomize_single_base}. Next, we will describe the main steps of the pipeline.

\smallskip\noindent\textbf{Scalp creation.} The first step in creating the new dataset is defining a scalp mesh onto which to build the hairstyles. Due to its wide applicability in the human-reconstruction field, we use the SMPL-X model~\cite{smplx} to define our scalp. We create a SMPL-X neutral body with mean body shape from which we manually select the vertices corresponding to the scalp and extract a mesh of \num{513} vertices as seen in~\cref{fig:overview}. The scalp mesh is then UV-unwrapped into a single chart using an As-Rigid-As-Possible (ARAP) parametrization~\cite{sorkine2007rigid} in order to minimize distortions. This operation is done only once and need not be repeated for each subject since they all share the same SMPL-X body topology.
 
\smallskip\noindent\textbf{Base hairstyle creation.} The scalp mesh is imported into Blender and a series of hairstyles is manually created in the form of coarsely defined guide strands (\cref{fig:randomize_single_base}). The base hairstyles define a starting point for the randomization process and therefore do not need to contain fine details, rather defining just the overall shape and direction of the hair. We manually create \num{75} base hairstyles, each represented by $\approx 50$ strands. We note that this is not a labour intensive process, as each base hairstyle takes only a few minutes to create. 

\smallskip\noindent\textbf{Randomization pipeline.} The guide strands are passed through a series of transformations defined as geometry nodes in Blender that gradually add more detail to the hair. The guide strands are first interpolated to \num{100}K strands, 
after which clumping, curling and noise are applied. These operations are repeated multiple times with various intensities. Special attention must be given to prevent the hair from penetrating the scalp or intersecting with the body for which we apply shrinkwrap modifiers at various stages in the pipeline. In total the pipeline consists of \num{58} geometry nodes that transform the hair together with \num{349} auxiliary nodes that help feed parameters into the main nodes.
We also define a hair BSDF using the Chiang model~\cite{chiang2015practical} for the strands material. 
Lastly we apply physics simulation to the strands using Blender's cloth simulation and randomize wind direction and gravity strength. 
A total of \num{110} parameters govern the whole Blender generation pipeline (geometry+material) and we select them at random with minimum and maximum ranges chosen empirically.

\smallskip\noindent\textbf{Scene setup.} In order to render realistic images, it is crucial to match the lighting of real scenes. For this, we use \num{255} HDRIs from Poly Haven\footnote{\href {https://polyhaven.com/hdris}{https://polyhaven.com/hdris}} and randomize their exposure and azimuthal angle. 
Additionally, we randomly select albedo, normal and roughness textures for the bodies from a library of artist-created SMPL-X textures. 
Furthermore, we also randomly change the face expression coefficients within the range $[-1,1]$. We keep the jaw closed since we observed it can cause issues with strands physics.
To train our network for robustness to different camera parameters, we also randomize camera position (maintaining the look-at direction towards the head) and focal length in the range \SIrange{40}{150}{\mm}. For a study on network robustness to camera parameters please refer to the \textbf{SupMat}.

\smallskip\noindent\textbf{Full dataset.} The full dataset contains \num{40}K samples each containing 3D strands with $\approx100$K strands and a corresponding RGB image rendered through path tracing at $768 \times 768$ resolution. We additionally save a density map that defines the probability of a strand being generated at each texel of the scalp as shown in~\cref{fig:scalp_texture_vis}. The density map is especially useful for modelling hairstyles with balding patterns.

\subsection{Hairstyle Parameterization}\label{parameterization}
\paragraph{Strand Parameterization.} Formally, we represent a 3D strand as a set of $\length$ 3D points $\strand = \left\{\pos_1,\pos_2,\cdots,\pos_{\length} \right\}$, where $\length=256$. 
A full hairstyle consists of a set of $\approx 100$K strands $\Hair=\left\{\strand_j \right\}_0^{100\textrm{K}}$.
Similar to Neural Strands~\cite{rosu2022neural}, we train a strand VAE to compress individual strands into a compact latent code $\latent \in \Real^{64}$.
The encoder of the VAE denoted by $\encoder()$ is modeled as a series of 1D convolutions and the decoder denoted by $\generator()$ is a modulated SIREN \cite{mehta2021modulated} that maps the latent code $\latent$ back to a series of 3D points $\strand$.

In addition to the position and direction loss proposed by Neural Strands~\cite{rosu2022neural}, we incorporate a curvature loss $\Loss_\textrm{cur}$. This is motivated by the observation that, for highly coiled hair, it is more perceptually impactful to match the curvature of the strand rather than matching the position of every point along the strand (i.e., as long as the strand \textit{looks} curvy, it is acceptable to commit errors in per-point position). For an evaluation of this observation please refer to the \textbf{SupMat}. We define the local curvature as the instantaneous change in direction: $\curv_i = \direction_{i+1}-\direction_{i}$ where the direction is defined as the change in position: $\direction_i=\pos_{i+1}-\pos_{i}$. Therefore, the new curvature loss is defined as the L1 distance between the ground-truth and the prediction: $\Loss_\textrm{cur} = \sum_{0}^\length \abs{\curv_i - \tilde{\curv_i}}$, where $\hat{\curv_i}$ is the curvature of the predicted strand.

The final training loss of our strands VAE is a weighted combination of a data term consisting of positional, directional and curvature losses together with a KL regularization term:
\begin{align}
\begin{split}\label{eq:strand}
    \Loss_\textrm{data} &= \sum_{0}^\length \abs{\pos_i - \tilde{\pos_i}} +\lambda_{1}\abs{\direction_i - \tilde{\direction_i}}  + \lambda_{2}\abs{\curv_i - \tilde{\curv_i}}
\end{split}\\
\begin{split}\label{eq:strand_vae}
    \Loss_\textrm{VAE} &= \Loss_\textrm{data} + \lambda_\textrm{KL}\Loss_\textrm{KL}.
\end{split}
\end{align}

For more details on the hyperparameters please refer to the \textbf{SupMat}.

\paragraph{Neural Scalp Texture.}

\begin{figure}
\captionsetup[subfigure]{justification=centering}
     \centering
     \begin{subfigure}[t]{0.11\textwidth}
         \centering
         \includegraphics[width=\textwidth]{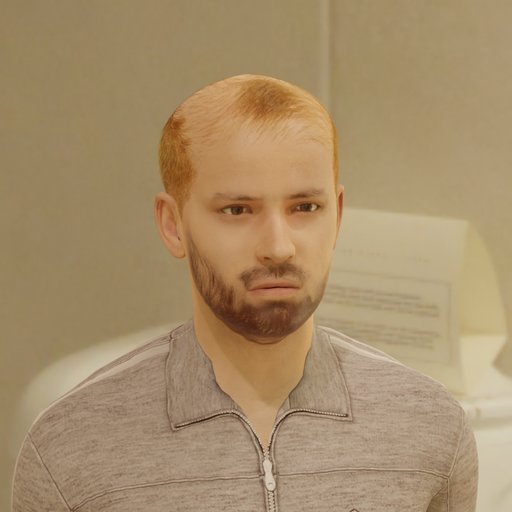}
         \caption{Synthetic\\Image}
         \label{fig:scalp_texture_vis_a}
     \end{subfigure}
     \hspace{-0.1em}
     \begin{subfigure}[t]{0.11\textwidth}
         \centering
         \includegraphics[width=\textwidth]{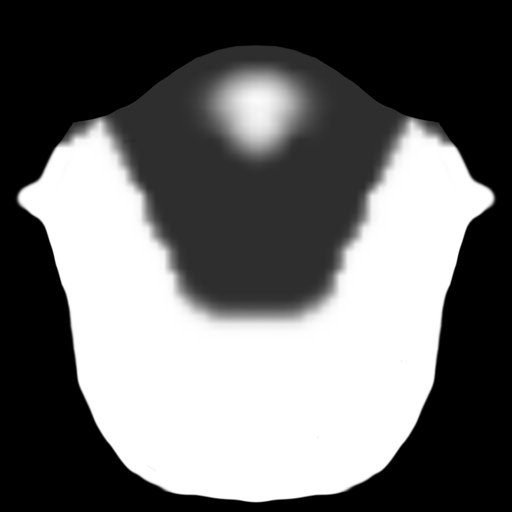}
         \caption{Scalp density}
         \label{fig:scalp_texture_vis_b}
     \end{subfigure}
     \hspace{-0.1em}
     \begin{subfigure}[t]{0.11\textwidth}
         \centering
         \includegraphics[width=\textwidth]{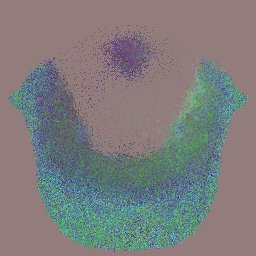}
         \caption{Raw scalp texture}
         \label{fig:scalp_texture_vis_c}
     \end{subfigure}
     \hspace{-0.1em}
     \begin{subfigure}[t]{0.11\textwidth}
         \centering
         \includegraphics[width=\textwidth]{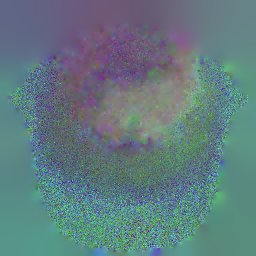}
         \caption{Interpolated texture}
         \label{fig:scalp_texture_vis_d}
     \end{subfigure}
     \vspace{-0.2cm}
    \caption{\textbf{Scalp interpolation}. Embedding each strand onto a the scalp texture can leave gaps where no strand is present. In order to enable probabilistic sampling of strands, we interpolate the sparse strand embeddings to achieve a smooth scalp texture that can be sampled at any position according to the density.}
    \label{fig:scalp_texture_vis}
\end{figure}

Each 3D hairstyle generated from the Blender pipeline is parametrized onto the scalp surface using a regular 2D texture. Specifically, each strand $\strand$ is encoded to a latent code $\latent$ by the strand encoder $\latent=\encoder(\strand)$ and directly assigned to the texel corresponding to the root of the strand.
If two strands are so close together that they are assigned the same texel on the scalp, one of them is chosen at random.
This produces a raw scalp texture $\scalptexture_{raw} \in \R^{256\times256\times64}$, as shown in~\cref{fig:scalp_texture_vis_c} where each texel contains a latent strand code.
Since not all regions of the scalp have an equal hair density, we introduce a density map $\density \in \R^{256\times256\times1}$ with values in the range $[0,1]$ that defines the probability of a strand being present at each texel. For more discussion on the properties of the density map please refer to the \textbf{SupMat}.

The aforementioned creation of the scalp texture can create sparse regions with no latent codes since not every texel will correspond to a strand. This sparsity is particularly evident for hairstyles with balding patterns and can cause issues when decoding the texture back to 3D since sampling a texel with no assigned latent code would yield undefined strand geometry.
To avoid this issue, we  interpolate the sparsely assigned latent codes by using a push-pull algorithm~\cite{kraus2009pull} that ensures that every texel on the scalp has a valid strand latent code as shown in~\cref{fig:scalp_texture_vis_d}.

With the interpolated scalp texture $\scalptexture$, we can sample at UV-positions according to the density $\density$, and produce a full hairstyle $\Hair$ by decoding each latent code:
\begin{equation}
    \Hair = \generator(\mathbf{\emph{Sample}}(\scalptexture,\density)),
\end{equation}
where $\emph{Sample}(\cdot)$ performs probabilistic sampling in UV space such that regions with high density are sampled more often, allowing us to represent hairstyles with varying density of hair. To create a full hairstyle we sample \num{100}K positions on the scalp and extract strand latent codes $\latent$ using bilinear interpolation.

\subsection{Conditional Scalp Diffusion}\label{diffusion}

\paragraph{Diffusion framework.} 
Similar to previous methods \cite{zhou2023groomgen,sklyarova2023haar} we train a generative model of scalp textures $\scalptexture$ to learn the prior distribution over hairstyles. Furthermore, we condition the model on image data, enabling high-quality 3D hair generation from a single image. 
Specifically, we train a denoiser model $\denoiser$ using a Hourglass Diffusion Transformer~\cite{hdit} architecture, which is particularly efficient for pixel-space diffusion tasks.
We denote the clean training sample for the diffusion as the concatenation of scalp texture and density map: $\mathbf{y} = [ \scalptexture,\density ]$ and the noised input $\mathbf{x} = \mathbf{y} + \mathbf{n}$ as a combination of the clean data $\mathbf{y}$ and noise $\mathbf{n} \sim\ \mathcal{N}(0,\sigma^2 \mathbf{I})$ with noise level $\sigma$.
Following EDM~\cite{karras2022elucidating}, we parametrize the denoiser to allow it to predict $\mathbf{y}$ or $\mathbf{n}$, or something in between:
%
\begin{equation}
    \begin{aligned}
        \denoiser(\mathbf{x};\sigma) = \cskip(\sigma) \mathbf{x} + \cout(\sigma) \denoiserInner \bigl( \cin(\sigma) \mathbf{x}; \cnoise(\sigma) \bigr)
    \end{aligned}
\end{equation}
where $\denoiserInner$ is the network to train, $\cskip$ is as a skip connection modulator, $\cin$ and $\cout$  scale the input and output magnitude, and $\cnoise$ is the time condition of the network. 
Please refer to the EDM paper~\cite{karras2022elucidating} for more details.

The training loss for each spatial element and each noise level is defined as:
\begin{equation}
    \begin{aligned}
        \Loss(\denoiser,\mathbf{y},\mathbf{n},\sigma) = \norm{ \denoiser(\mathbf{y}+\mathbf{n};\sigma)  - \mathbf{y} }_2^2 .
    \end{aligned}
\end{equation}

\paragraph{Loss scaling.} 
The denoising network outputs a clean scalp texture $\scalptexture$ and a density map $\density$. However, we observe that not all channels of the scalp texture are equally significant, some encode relevant information such as length or curliness, while others are noisy and have little impact on the decoded strand. We propose to down-weight the loss of the channels of the scalp textures based on their perceptual change to the strand shape. Since we have no direct metric to evaluate perceptual change, we reuse the data term from the loss $\Loss_\textrm{data}$ used to train the StrandVAE.
More specifically, we first define the mean strand predicted by the VAE as $\strand_0 = \generator(\latent_0)$ by decoding from a zero latent vector $\latent_0=[0]^{64}$. Following that, we individually perturb each dimension of the latent code, decode a strand, and calculate the distance to the mean strand as measured by the loss. This can be seen as a metric that quantifies perceptual changes as a result of individual changes to the latent code dimensions.  Therefore, the weight associated with each dimension of the latent code is:
\begin{equation}
    \begin{aligned}
        w_i = \Loss_\textrm{strand}( S_0,  \generator(\latent_0+ \epsilon \cdot \mathbf{j}_i)),
    \end{aligned}
\end{equation}
where $\mathbf{j}_i$ is vector of zeros with a $1$ at the i-th position and $\epsilon$ is the perturbation strength which we empirically set to \num{0.8}.
We obtain a final weight vector by normalizing:
\begin{equation}
    \begin{aligned}
        \mathbf{w} = \frac{w_i}{\sum_0^{64} w_i}.
    \end{aligned}
\end{equation}
For an evaluation and discussion of the proposed weighting scheme, please see \textbf{SupMat}.

In addition, we also incorporate a weighting term in order to equalize the error contribution across noise levels by taking the approach of EDM2~\cite{karras2024analyzing}, which treats denoising as a form of multi-task learning and lets the network predict a log-variance of the uncertainty $u(\noise)$ for a certain noise level. 
Combining the per-channel weighting with the per-noise level weighting we obtain our final formulation for the diffusion loss as an expectation over noise levels and spatial dimensions:
\begin{equation}
    \begin{aligned}
        \Loss(\denoiser,u) = \E_{\mathbf{y}, \mathbf{n},\sigma} \left[ \frac{\textbf{w}}{e^{u(\sigma)}} \Loss(\denoiser,\mathbf{y}, \mathbf{n}, \sigma) + u(\noise)  \right] .
    \end{aligned}
\end{equation}

\paragraph{RGB conditioning.} 
Previous approaches have primarily relied on 2D orientation maps~\cite{paris2004capture,zheng2023hairstep} extracted using Gabor filters in order to align the generated strands with the image. However, orientation maps cannot infer the hair growth direction and tend to be noisy for heavily occluded hairstyles. In our work, we forego the need to compute orientation maps and instead condition our diffusion network using robust DINOv2~\cite{dinov2} features extracted from the image. Specifically, we use the pre-trained DINOv2-large model to encode the RGB image into a feature map $\featuremap \in \R^{55\times55\times1024}$ and a global CLS token $\cls \in \R^{1024}$. Additionally, our rendered images are sufficiently realistic such that the network trained only on synthetic data, can generalize to in-the-wild images.

Since the scalp texture and the RGB image are not spatially aligned, we add cross-attention layers before each downsampling layer of the denoiser and let the network learn which regions of the feature map $\featuremap$ it should attend to. The CLS token $\cls$ is used to globally condition the network similar to the time condition $\sigma$ as described in HDiT~\cite{hdit}.

\begin{figure}[t]

	\centering
	\begin{tikzpicture}[ remember picture, >={Stealth[inset=1pt,length=8pt,angle'=30,round]} ]

     \def\xshift{1.9}
     \def\yshift{-2.5}
     \def\w{0.13}

    \node[] (origin) at (0,0) {};

    \node[inner sep=0pt, anchor=south] (base0) at (\xshift*0, 0) {\adjincludegraphics[width=\w\textwidth, trim={{0.1\width} {0.1\height} {0.1\width} 0}, clip]{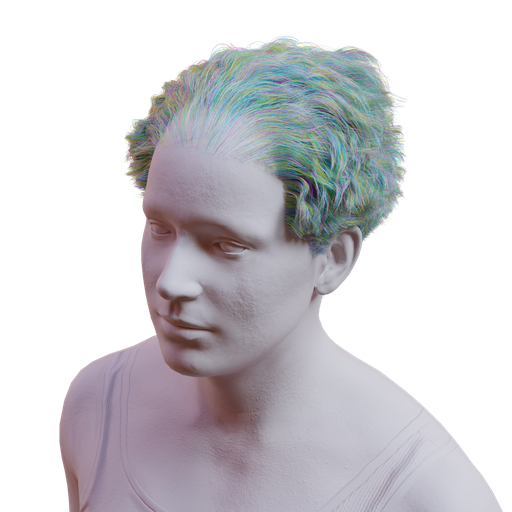}};
    \node[inner sep=0pt, anchor=south] (base0_1) at (\xshift*1, 0) {\adjincludegraphics[width=\w\textwidth, trim={{0.1\width} {0.1\height} {0.1\width} 0}, clip]{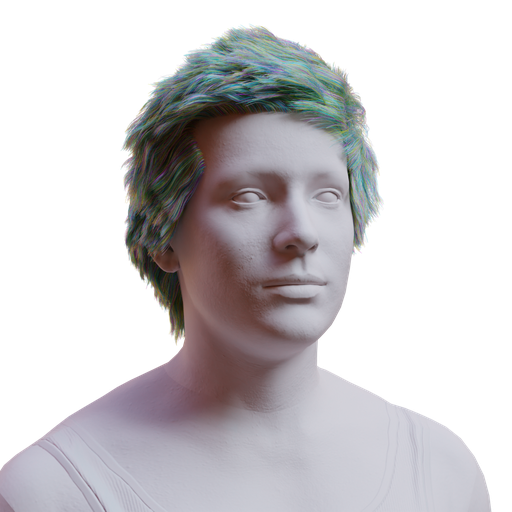}};
    \node[inner sep=0pt, anchor=south] (base0_2) at (\xshift*2, 0) {\adjincludegraphics[width=\w\textwidth, trim={{0.1\width} {0.1\height} {0.1\width} 0}, clip]{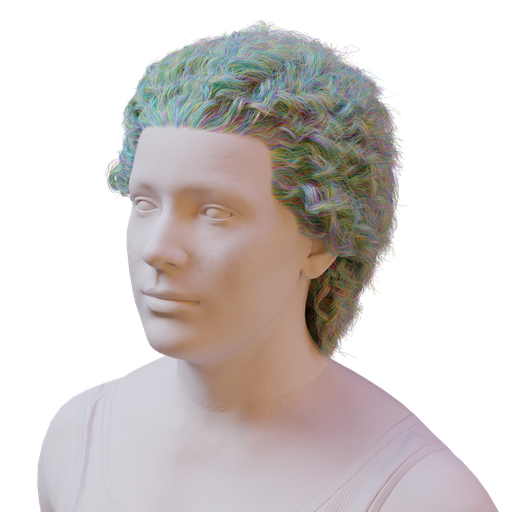}};
    \node[inner sep=0pt, anchor=south] (base0_3) at (\xshift*3, 0) {\adjincludegraphics[width=\w\textwidth, trim={{0.1\width} {0.1\height} {0.1\width} 0}, clip]{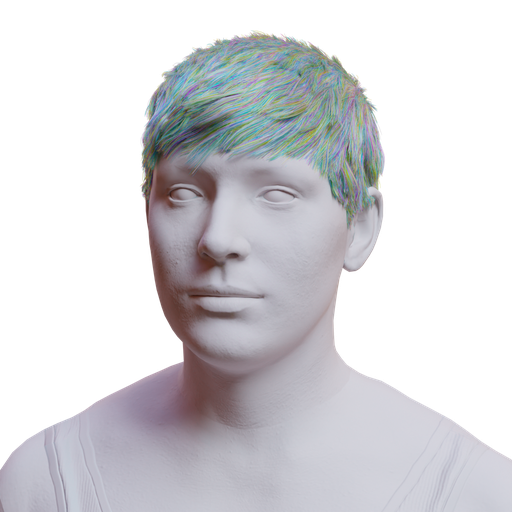}};

    \node[inner sep=0pt, anchor=south] (base0_6) at (\xshift*0, \yshift) {\adjincludegraphics[width=\w\textwidth, trim={{0.1\width} {0.1\height} {0.1\width} 0}, clip]{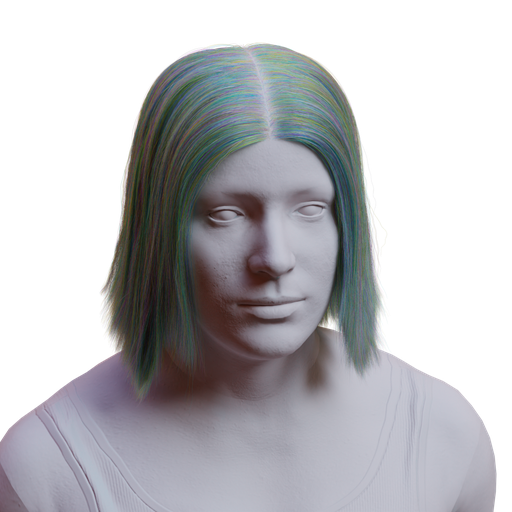}};
    \node[inner sep=0pt, anchor=south] (base0_7) at (\xshift*1, \yshift) {\adjincludegraphics[width=.11\textwidth, trim={{0.1\width} 0 {0.1\width} 0}, clip]{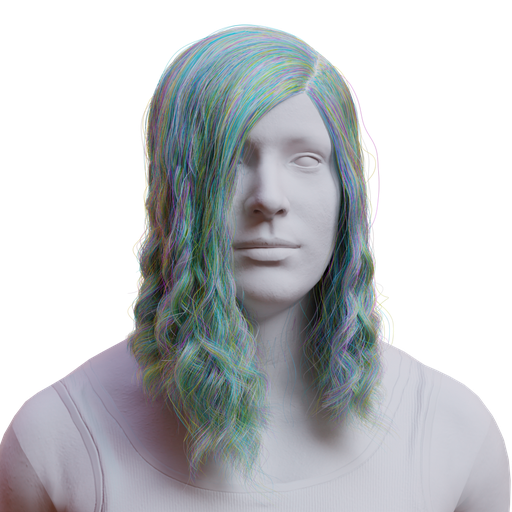}};
    \node[inner sep=0pt, anchor=south] (base0_8) at (\xshift*2, \yshift) {\adjincludegraphics[width=\w\textwidth, trim={{0.1\width} {0.1\height} {0.1\width} 0}, clip]{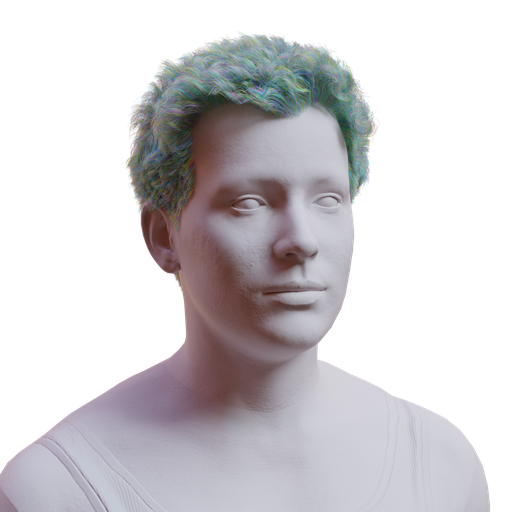}};
    \node[inner sep=0pt, anchor=south] (base0_9) at (\xshift*3, \yshift) {\adjincludegraphics[width=\w\textwidth, trim={{0.1\width} {0.1\height} {0.1\width} 0}, clip]{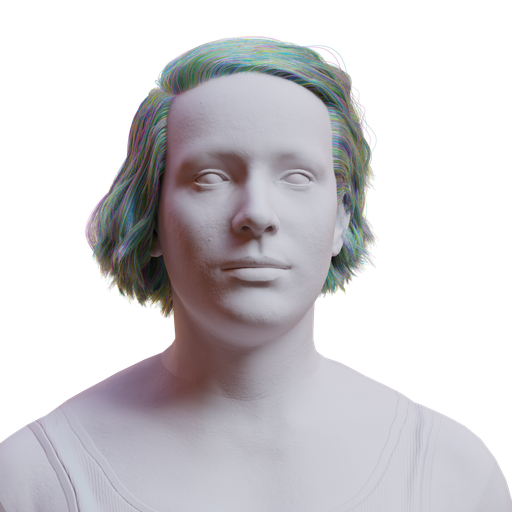}};


	\end{tikzpicture}%
    \vspace{-0.2cm}
	\caption{\textbf{Unconditional generation.} We train our diffusion model with Classifier-free Guidance~\cite{ho2022classifier} so that the same model can be evaluated both conditionally and unconditionally. Here we show unconditionally generated results where we observe a range of high-quality and diverse hairstyles.   }
	\label{fig:unconditional_generation}
\end{figure}

Furthermore, in order to use the same network both conditionally and unconditionally, we train the denoiser using Classifier-Free Guidance \cite{ho2022classifier} by randomly dropping the conditioning signal with $10\%$ probability (\cref{fig:unconditional_generation}). 

For experiments on conditioning using DINOv2 features vs orientation maps, please refer to \textbf{SupMat}.

\input{fig/hair_in_unreal}
\begin{figure*}[t]

	\centering
	\begin{tikzpicture}[ remember picture, >={Stealth[inset=1pt,length=8pt,angle'=30,round]} ]

     \def\xshift{2.7}
     \def\yshift{-3.0}
     \def\w{0.17}
     \def\firstshift{-0.2}

     \def\cropBottom{0.1cm}

    \node[] (origin) at (0,0) {};

    \definetrim{freeman_crop}{0.0 0.2 0.0 0.0}

    \node[inner sep=0pt] (freeman_rgb) at (\firstshift+\xshift*0, 0) {\adjincludegraphics[width=\w\textwidth, trim={{0.1\width} {0.2\height} {0.1\width} 0}, clip]{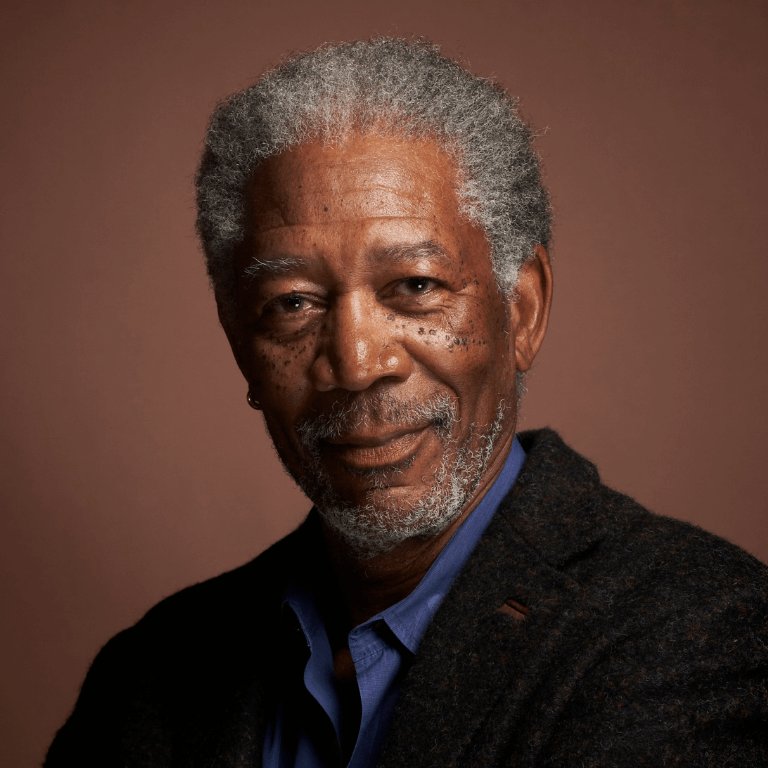}};
    \node[inner sep=0pt] (freeman_ours) at (\xshift*1, 0) {\adjincludegraphics[width=\w\textwidth, trim={{0.1\width} {0.2\height} {0.1\width} 0}, clip]{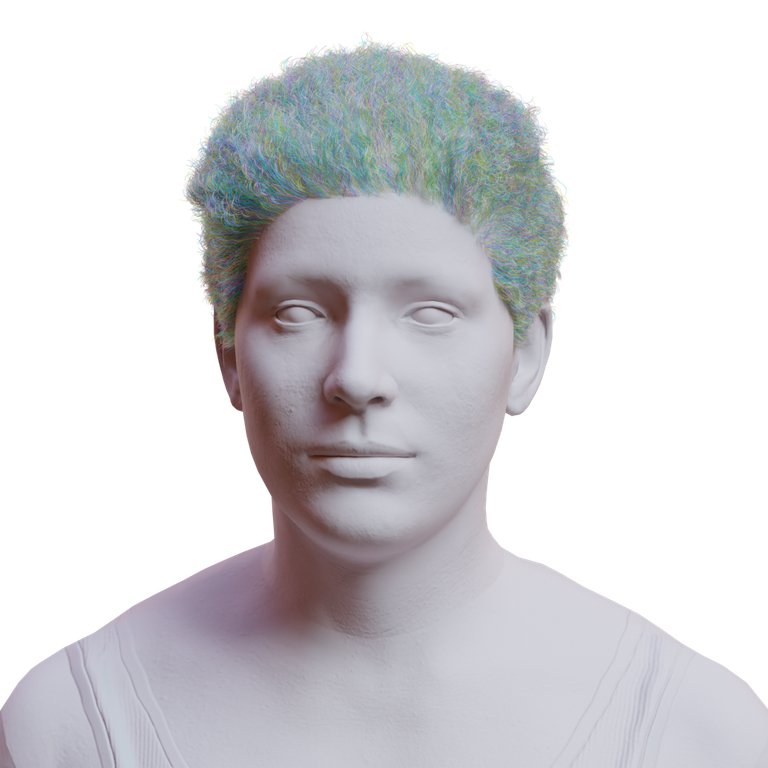}};
    \node[inner sep=0pt] (freeman_hairstep) at (\xshift*2, 0) {\adjincludegraphics[width=\w\textwidth, trim={{0.1\width} {0.2\height} {0.1\width} 0}, clip]{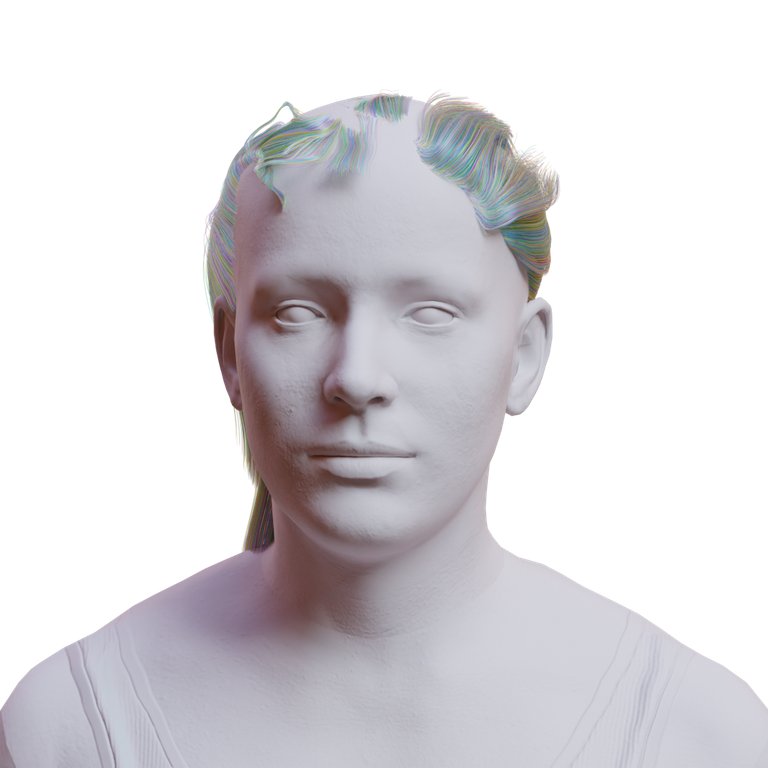}};
    \node[inner sep=0pt] (freeman_neural_hd) at (\xshift*3, 0) {\adjincludegraphics[width=\w\textwidth, trim={{0.1\width} {0.2\height} {0.1\width} 0}, clip]{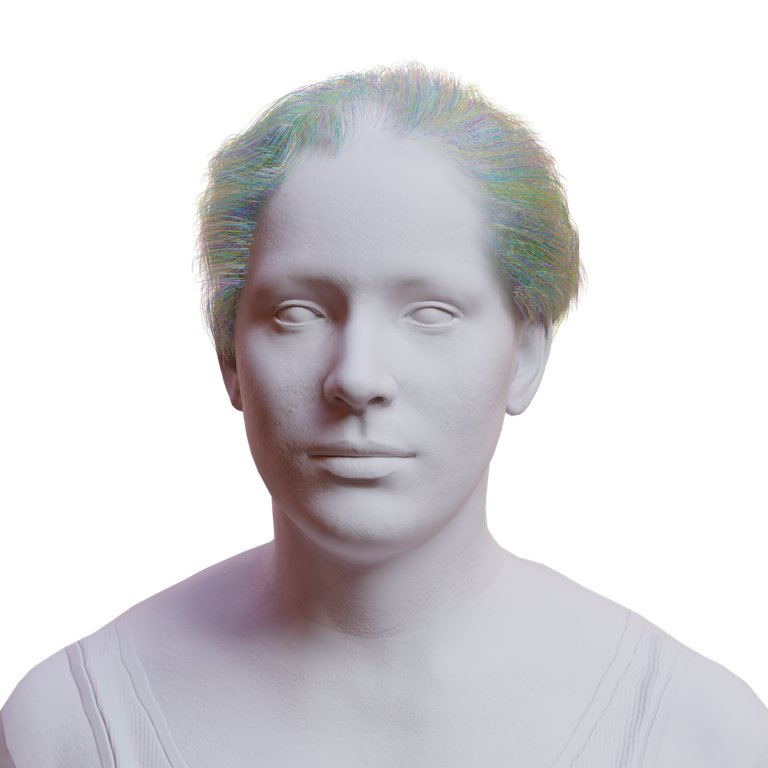}};
    \node[inner sep=0pt] (freeman_haar) at (\xshift*4, 0) {\adjincludegraphics[width=\w\textwidth, trim={{0.1\width} {0.2\height} {0.1\width} 0}, clip]{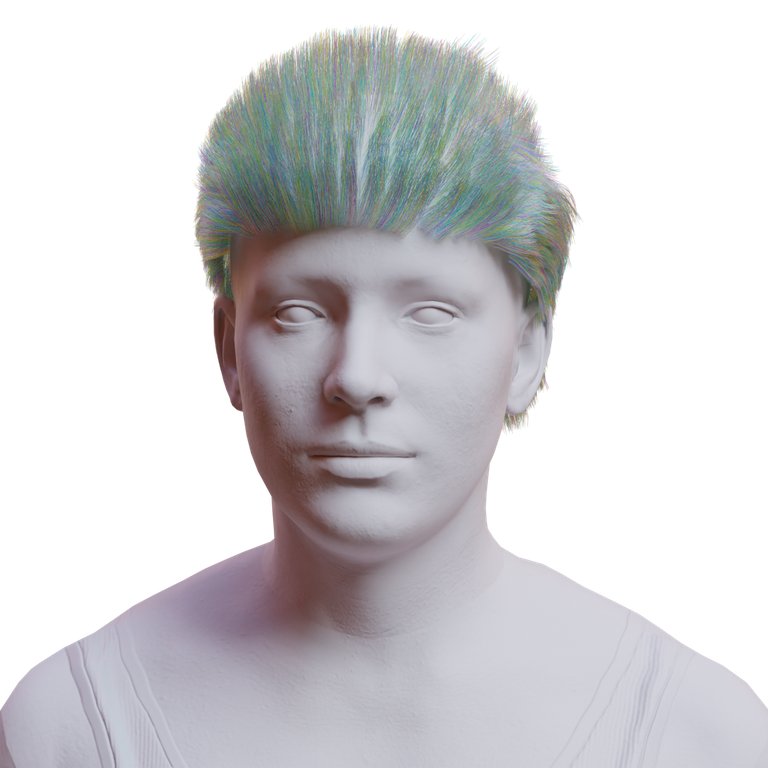}};

    \node[inner sep=0pt] (bale_rgb) at (\firstshift+\xshift*0, \yshift) {\adjincludegraphics[width=\w\textwidth, trim={{0.1\width} {0.2\height} {0.1\width} 0}, clip]{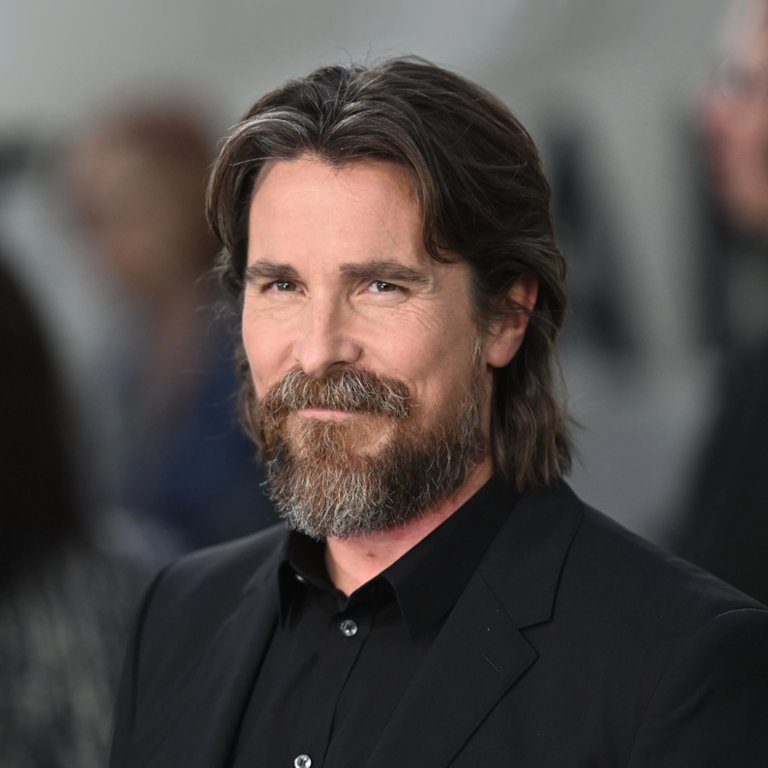}};
    \node[inner sep=0pt] (bale_ours) at (\xshift*1, \yshift) {\adjincludegraphics[width=\w\textwidth, trim={{0.1\width} {0.15\height} {0.1\width} {0.05\height}}, clip]{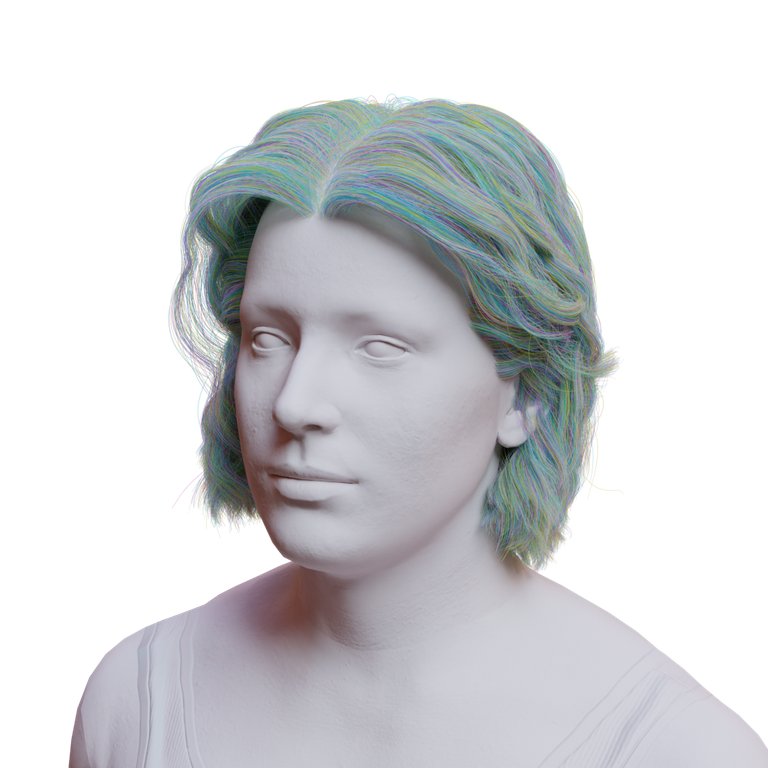}};
    \node[inner sep=0pt] (bale_hairstep) at (\xshift*2, \yshift) {\adjincludegraphics[width=\w\textwidth, trim={{0.1\width} {0.15\height} {0.1\width} {0.05\height}}, clip]{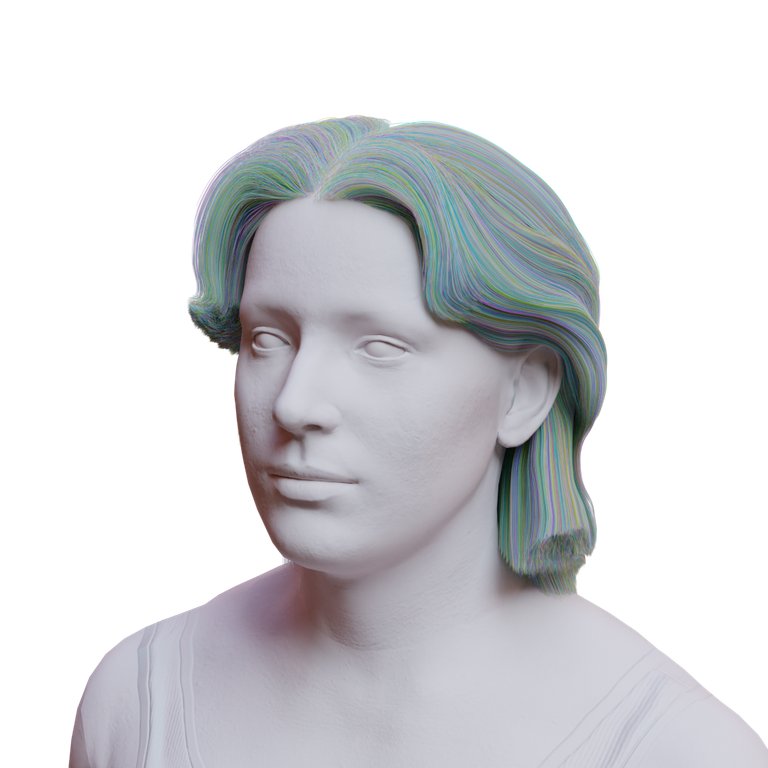}};
    \node[inner sep=0pt] (bale_neural_hd) at (\xshift*3, \yshift) {\adjincludegraphics[width=\w\textwidth, trim={{0.1\width} {0.15\height} {0.1\width} {0.05\height}}, clip]{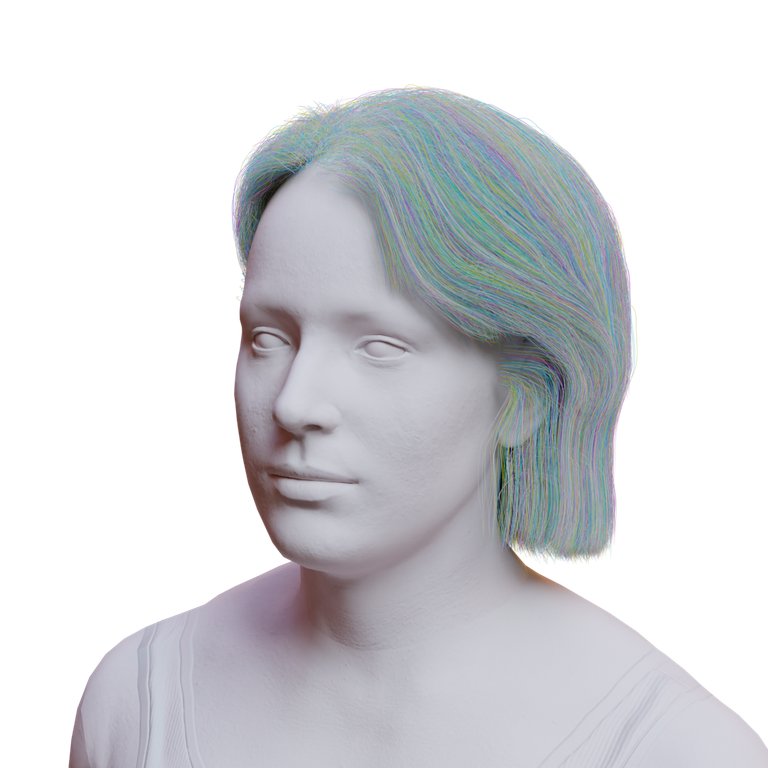}};
    \node[inner sep=0pt] (bale_haar) at (\xshift*4, \yshift) {\adjincludegraphics[width=\w\textwidth, trim={{0.1\width} {0.15\height} {0.1\width} {0.05\height}}, clip]{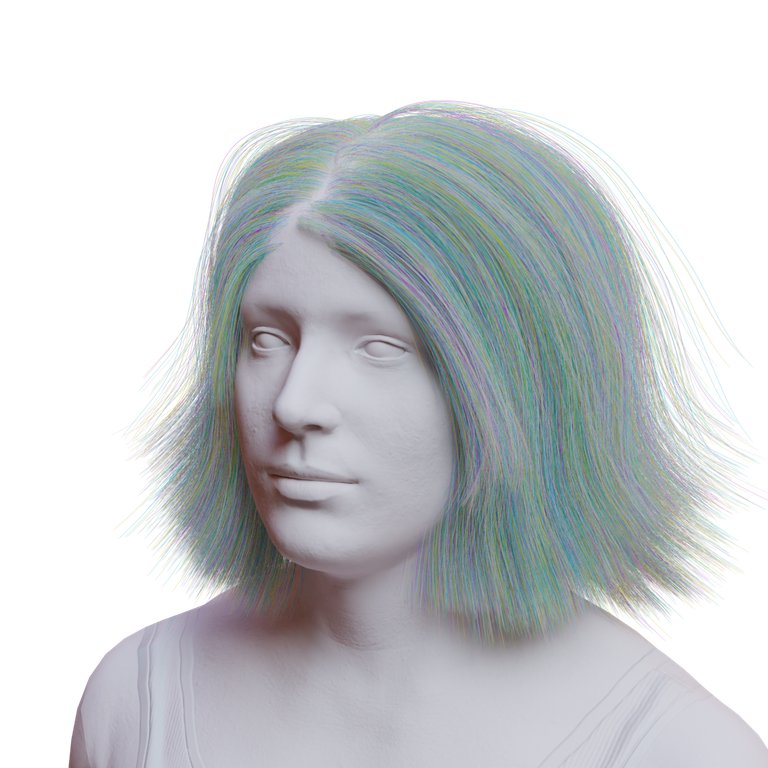}};


    \node[inner sep=0pt] (harris_rgb) at (\firstshift+\xshift*0, \yshift*2) {\adjincludegraphics[width=\w\textwidth, trim={{0.1\width} {0.2\height} {0.1\width} 0}, clip]{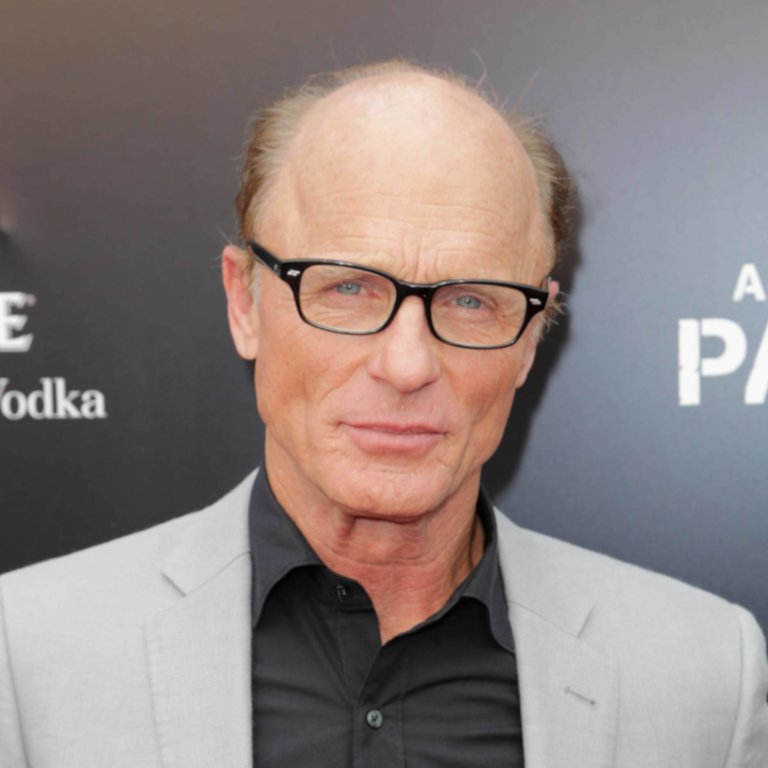}};
    \node[inner sep=0pt] (harris_ours) at (\xshift*1, \yshift*2) {\adjincludegraphics[width=\w\textwidth, trim={{0.1\width} {0.17\height} {0.1\width} {0.03\height}}, clip]{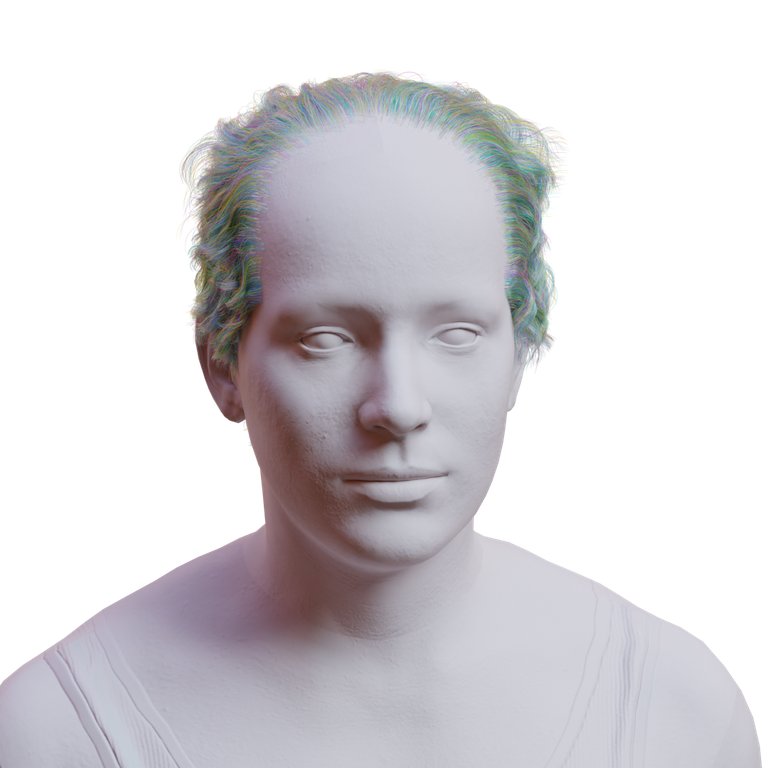}};
    \node[inner sep=0pt] (harris_hairstep) at (\xshift*2, \yshift*2) {\adjincludegraphics[width=\w\textwidth, trim={{0.1\width} {0.17\height} {0.1\width} {0.03\height}}, clip]{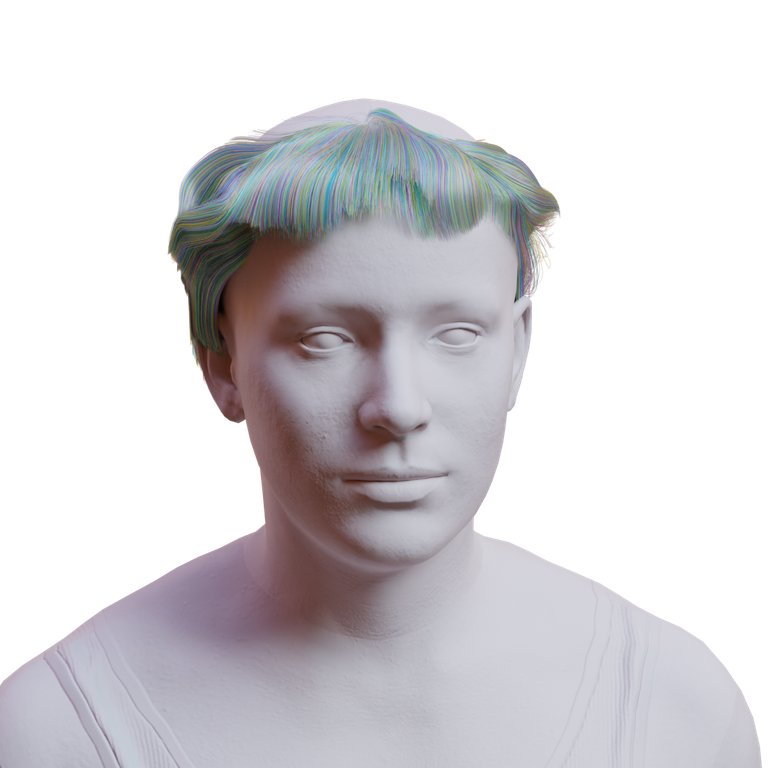}};
    \node[inner sep=0pt] (harris_neural_hd) at (\xshift*3, \yshift*2) {\adjincludegraphics[width=\w\textwidth, trim={{0.1\width} {0.17\height} {0.1\width} {0.03\height}}, clip]{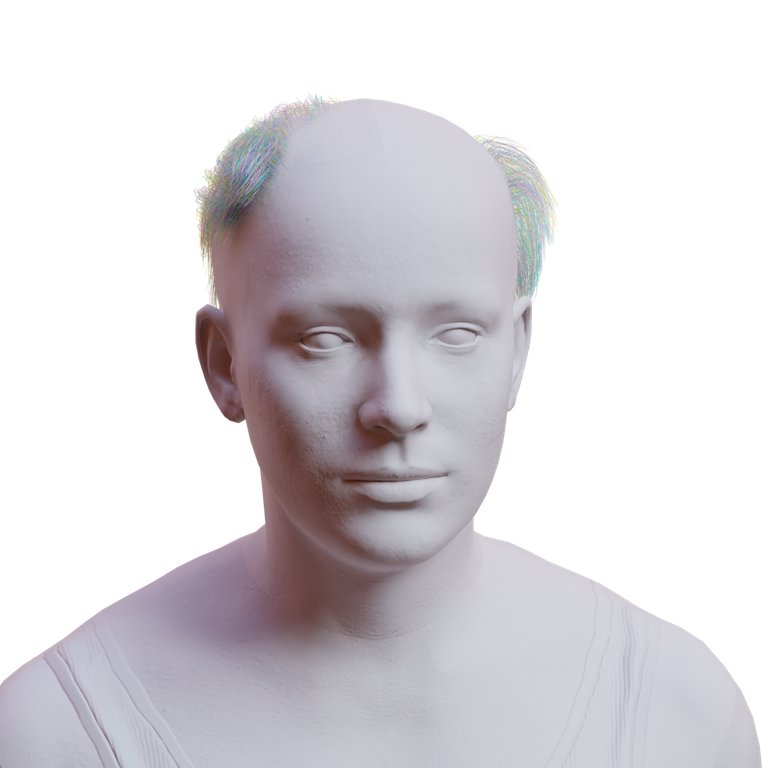}};
    \node[inner sep=0pt] (harris_haar) at (\xshift*4, \yshift*2) {\adjincludegraphics[width=\w\textwidth, trim={{0.1\width} {0.17\height} {0.1\width} {0.03\height}}, clip]{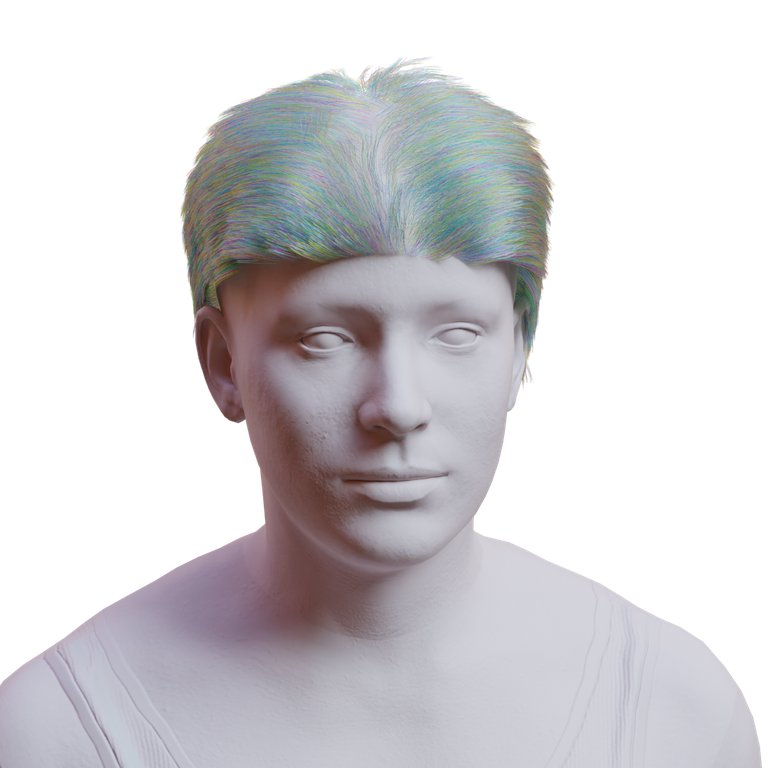}};

    \node[inner sep=0pt] (hathaway_rgb) at (\firstshift+\xshift*0, \yshift*3) {\adjincludegraphics[width=\w\textwidth, trim={{0.05\width} {0.1\height} {0.05\width} 0}, clip]{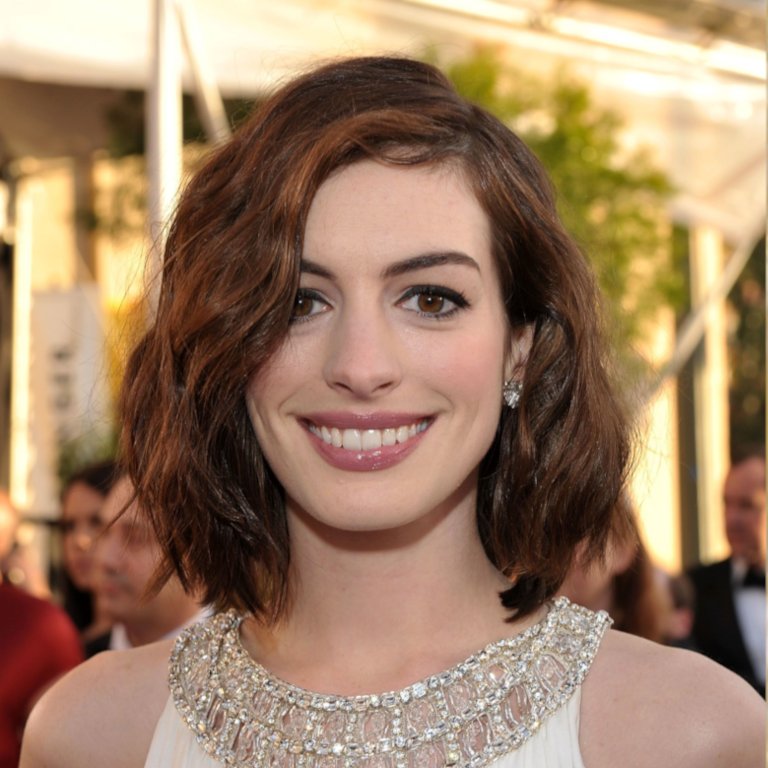}};
    \node[inner sep=0pt] (hatahway_ours) at (\xshift*1, \yshift*3) {\adjincludegraphics[width=\w\textwidth, trim={{0.05\width} {0.1\height} {0.05\width} 0}, clip]{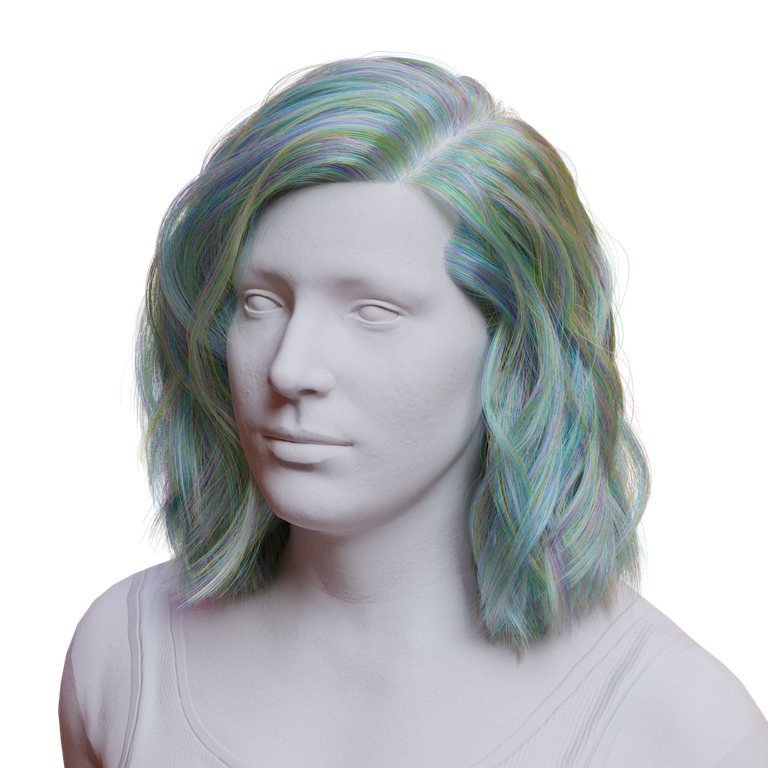}};
    \node[inner sep=0pt] (hathaway_hairstep) at (\xshift*2, \yshift*3) {\adjincludegraphics[width=\w\textwidth, trim={{0.05\width} {0.1\height} {0.05\width} 0}, clip]{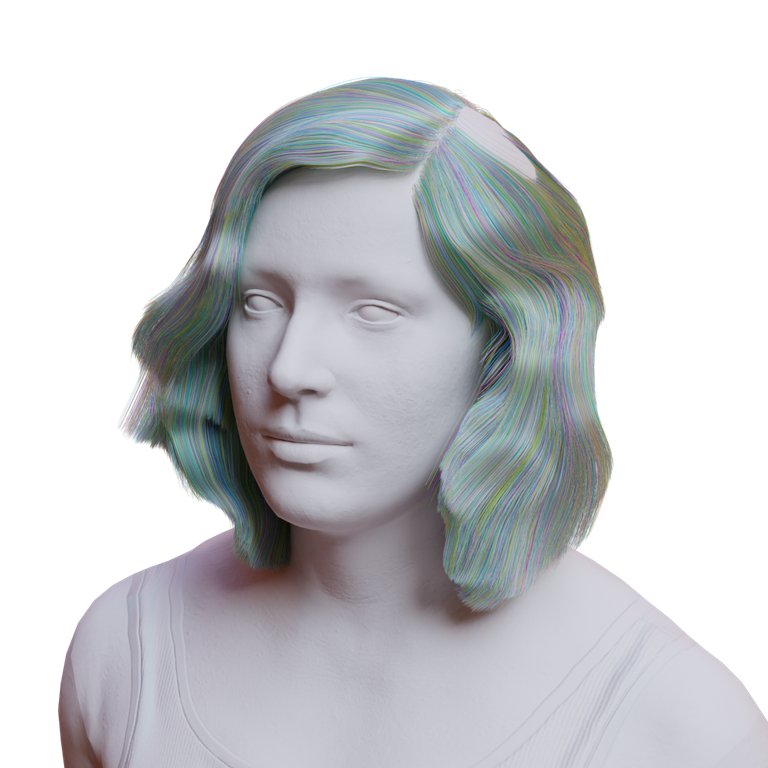}};
    \node[inner sep=0pt] (hatahway_neural_hd) at (\xshift*3, \yshift*3) {\adjincludegraphics[width=\w\textwidth, trim={{0.05\width} {0.1\height} {0.05\width} 0}, clip]{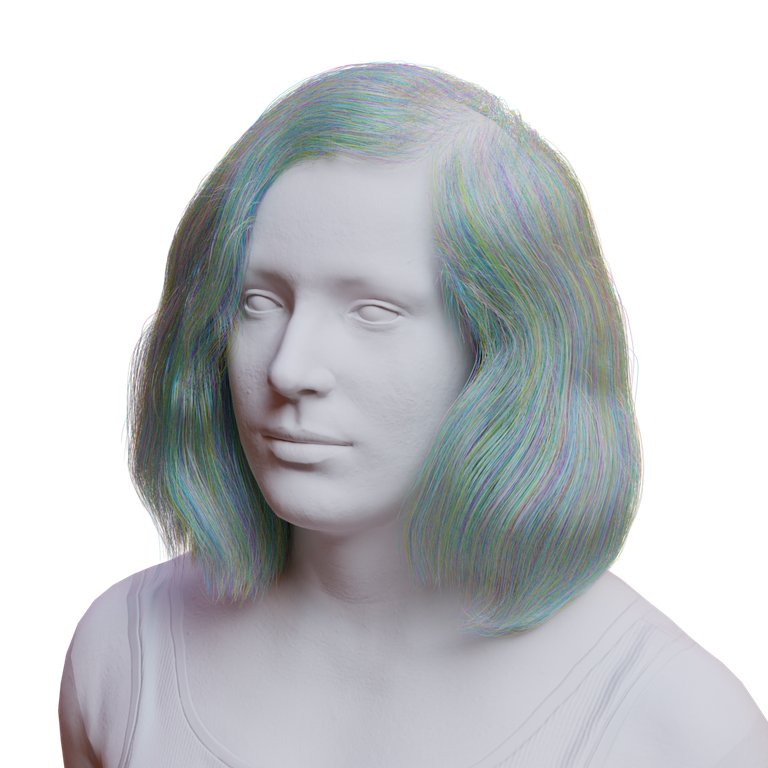}};
    \node[inner sep=0pt] (hatahway_haar) at (\xshift*4, \yshift*3) {\adjincludegraphics[width=\w\textwidth, trim={{0.05\width} {0.1\height} {0.05\width} 0}, clip]{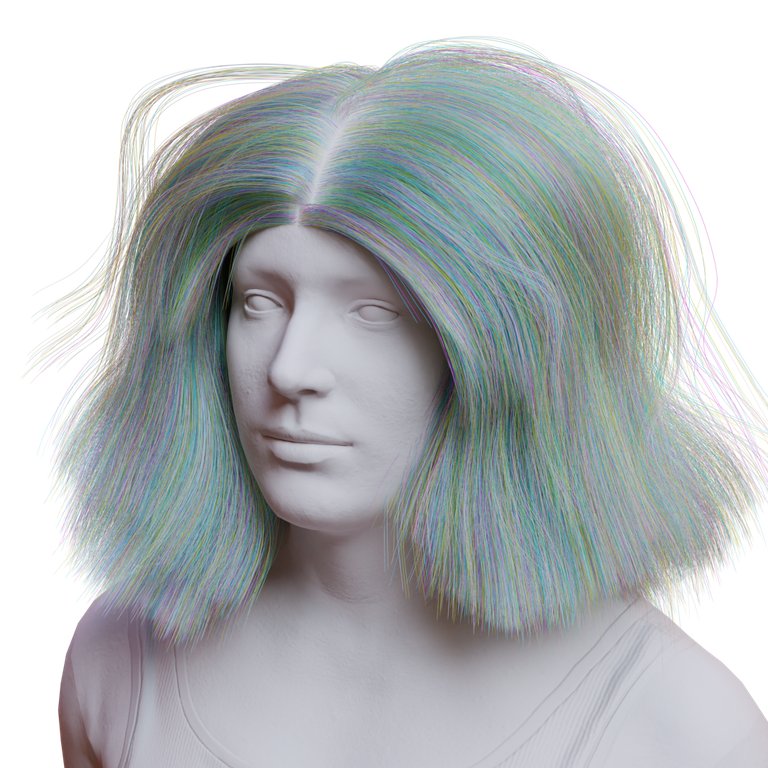}};

    \node[inner sep=0pt] (buzzcut_rgb) at (\firstshift+\xshift*0, \yshift*4) {\adjincludegraphics[width=\w\textwidth, trim={{0.1\width} {0.2\height} {0.1\width} 0}, clip]{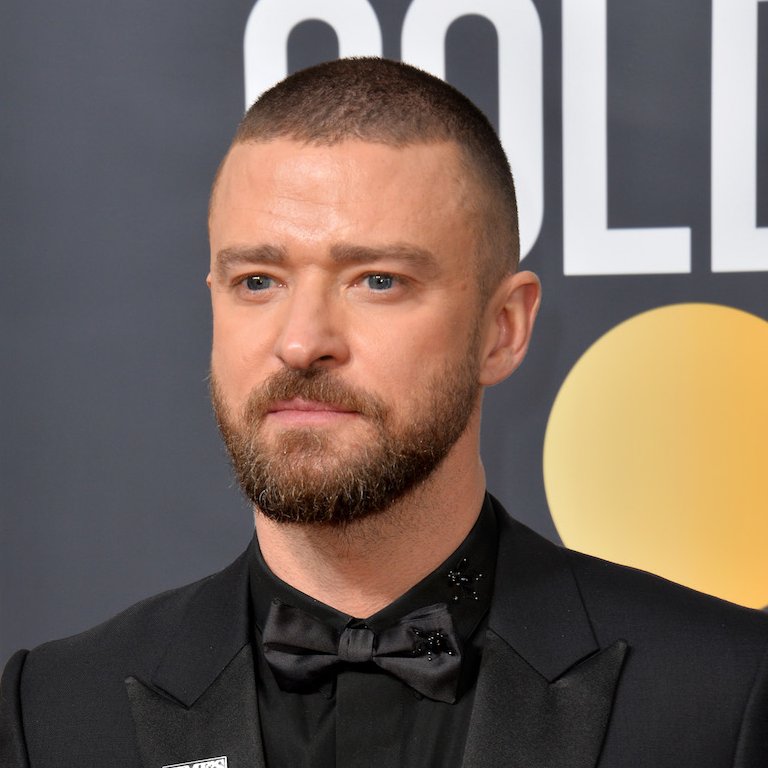}};
    \node[inner sep=0pt] (buzzcut_ours) at (\xshift*1, \yshift*4) {\adjincludegraphics[width=\w\textwidth, trim={{0.1\width} {0.15\height} {0.1\width} {0.05\height}}, clip]{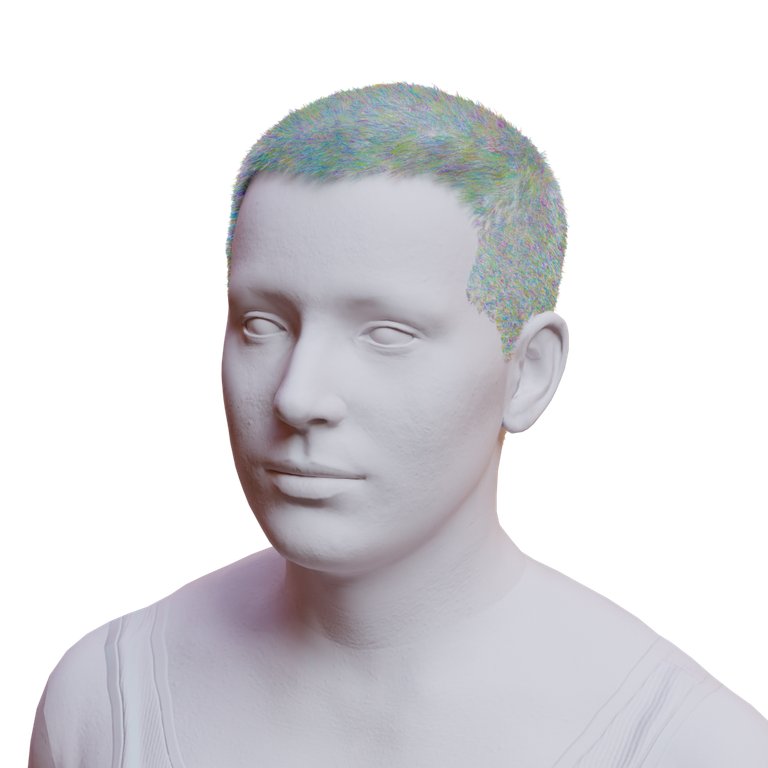}};
    \node[inner sep=0pt] (buzzcut_hairstep) at (\xshift*2, \yshift*4) {\adjincludegraphics[width=\w\textwidth, trim={{0.1\width} {0.15\height} {0.1\width} {0.05\height}}, clip]{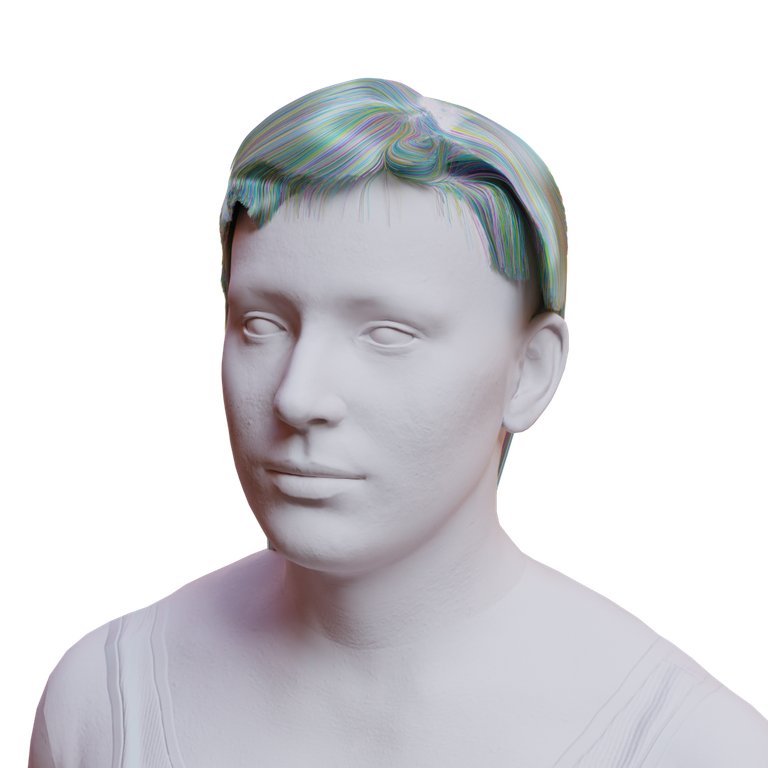}};
    \node[inner sep=0pt] (buzzcut_neural_hd) at (\xshift*3, \yshift*4) {\adjincludegraphics[width=\w\textwidth, trim={{0.1\width} {0.15\height} {0.1\width} {0.05\height}}, clip]{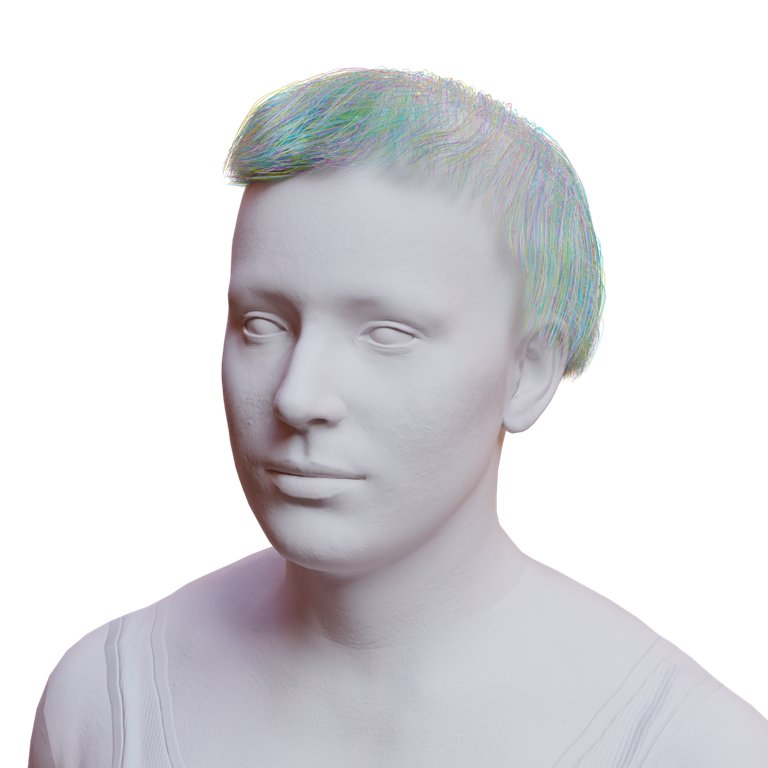}};
    \node[inner sep=0pt] (buzzcut_haar) at (\xshift*4, \yshift*4) {\adjincludegraphics[width=\w\textwidth, trim={{0.1\width} {0.15\height} {0.1\width} {0.05\height}}, clip]{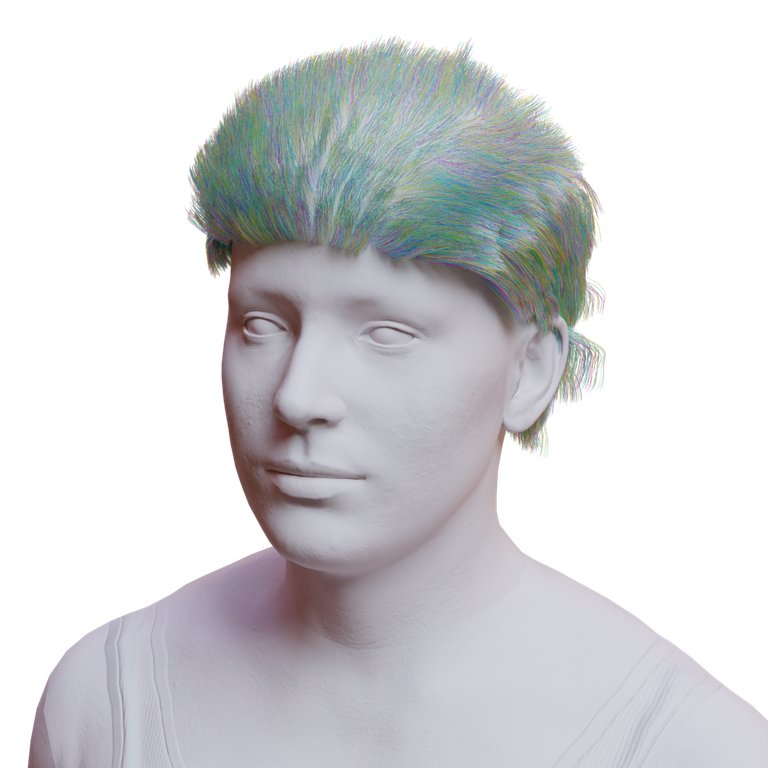}};



    \node[font=\footnotesize\selectfont, align=left, below = -0.0cm of buzzcut_rgb] {RGB input};
    \node[font=\footnotesize\selectfont, align=left, below = -0.0cm of buzzcut_ours] {Ours};
    \node[font=\footnotesize\selectfont, align=left, below = -0.0cm of buzzcut_hairstep] {HairStep};
    \node[font=\footnotesize\selectfont, align=left, below = -0.0cm of buzzcut_neural_hd] {NeuralHDHair};
    \node[font=\footnotesize\selectfont, align=center, below = -0.0cm of buzzcut_haar] {HAAR\\(img $\rightarrow$ text $\rightarrow$ 3D)};

	\end{tikzpicture}
	\vspace{-0.15in}
	\caption{\textbf{Qualitative comparison.} We compare our method with HairStep~\cite{zheng2023hairstep}, NeuralHDHair~\cite{wu2022neuralhdhair} and HAAR~\cite{sklyarova2023haar} on real images. Our method can robustly reconstruct a large variety of hairstyles with significantly more detail than previous approaches. We note that HAAR was designed to reconstruct hair given text captions, so we use LLaVA~\cite{liu2024improved} to obtain text description of hair, followed by applying their text-to-3D method.
      }
    \vspace{-0.05in}
	\label{fig:qualitative_results}
\end{figure*}
\begin{figure}[t]

	\centering
	\begin{tikzpicture}[ remember picture, >={Stealth[inset=1pt,length=8pt,angle'=30,round]} ]

     \def\xshift{1.7}
     \def\yshift{-0.25}

    \node[] (origin) at (0,0) {};

    \node[inner sep=0pt, anchor=south] (rgb) at (-0.6+\xshift*0, 0) {\adjincludegraphics[width=.085\textwidth, trim={{0.1\width} {0.1\height} {0.1\width} 0}, clip]{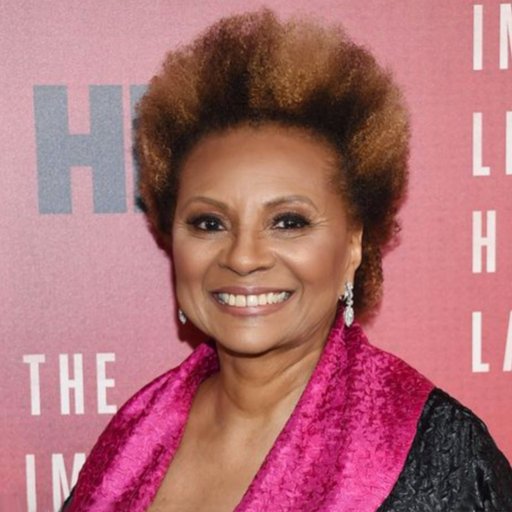}};

    \node[inner sep=0pt, anchor=south] (vae) at (2.45+\xshift*0, 0) {\adjincludegraphics[width=.085\textwidth, trim={{0.15\width} {0.1\height} {0.15\width} 0}, clip]{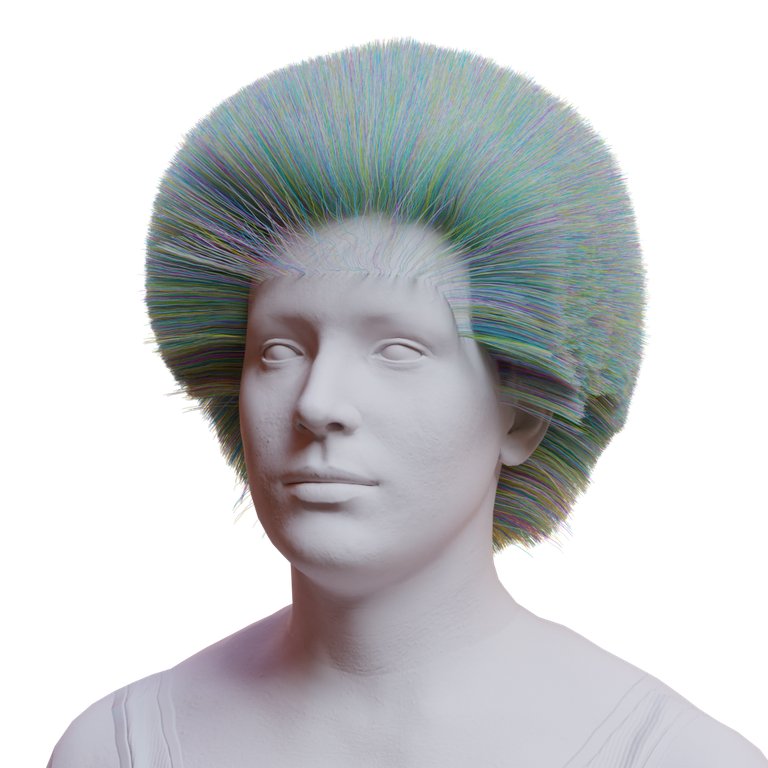}};
    
     \node[inner sep=0pt, anchor=south] (nolocalcond) at (2.45+\xshift*1, 0) {\adjincludegraphics[width=.085\textwidth, trim={{0.15\width} {0.1\height} {0.15\width} 0}, clip]{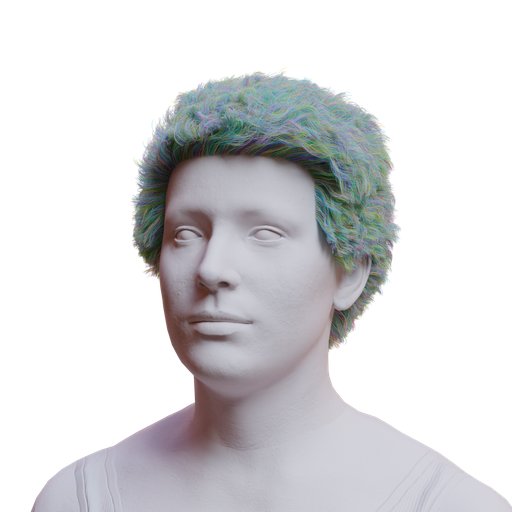}};
    
     \node[inner sep=0pt, anchor=south] (full) at (2.45+\xshift*2, 0) {\adjincludegraphics[width=.085\textwidth, trim={{0.15\width} {0.1\height} {0.15\width} 0}, clip]{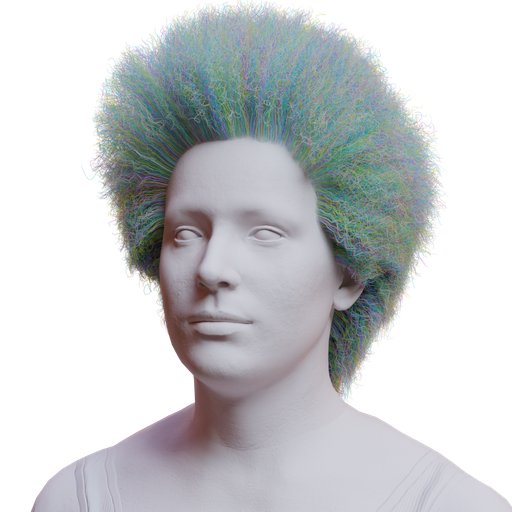}};


     \draw [-, ultra thin, dashed ] ($(rgb.north east)+(0.05,0.0)$) -- ($(rgb.south east)+(0.05, -0.7)$);

     \node[font=\footnotesize\selectfont, align=left, anchor=west] (model) at (0.17, -0.2) {Model:};
     \node[font=\footnotesize\selectfont, align=left, anchor=west] (cond) at (0.17, -0.6) {Condition:};
    
    \node[font=\footnotesize\selectfont, align=center, below = -0.0cm of rgb] {RGB Input};
    



    \node[font=\footnotesize\selectfont] at (vae |- model) {VAE};
    \node[font=\footnotesize\selectfont] at (vae |- cond) {Global+Local};

    \node[font=\footnotesize\selectfont] at (nolocalcond |- model) {Diffusion};
    \node[font=\footnotesize\selectfont] at (nolocalcond |- cond) {Global};

    \node[font=\footnotesize\selectfont] at (full |- model) {Diffusion};
    \node[font=\footnotesize\selectfont] at (full |- cond) {Global+Local};

	\end{tikzpicture}%
    \vspace{-0.15cm}
	\caption{\textbf{Ablation of conditioning signal and model architecture.} Removing the local conditioning in the form of feature map $\featuremap$ and relying only on the global CLS token severely affects the ability to reconstruct hair. Also generating scalp textures using a VAE yields overly smooth reconstructions. The combination of diffusion and local+global conditioning captures the most detail. }
	\label{fig:ablation_no_local_vae}
\end{figure}

\section{Experiments}\label{experiments}

We evaluate \sysname~using both synthetic data and in-the-wild images. 
We also perform ablation studies to assess the importance of the different components of our method (\cref{ablation}). For implementation details and additional experiments, please refer to \textbf{SupMat}.

We compare our method with Hairstep \cite{zheng2023hairstep} and NeuralHDHair \cite{wu2022neuralhdhair} as they are both state-of-the-art methods for reconstructing 3D hair from a single image. We also compare with the conditional 3D hair generation method of HAAR \cite{sklyarova2023haar} although it is mainly designed for text-to-hair.

\subsection{Qualitative}
The qualitative comparison with Hairstep, NeuralHDHair and HAAR is shown in \cref{fig:qualitative_results}. NeuralHDHair and HairStep tend to create hair that matches the overall shape from the image but that lacks fine-grained detail and curls, reconstructing strands that run parallel to each other. Furthermore, they cannot correctly reconstruct balding and afro-like hairstyles. In contrast, our method produces high-quality and realistic hair strands for a wide variety of images.
Despite being a method designed for text-to-hair, we also compare with HAAR since they have released a pipeline for image-to-hairstyle by leveraging LLaVA to obtain text descriptions. We observe that their method often cannot match the hairstyle in the image.

\subsection{Quantitative}

Quantitative evaluation is performed using the synthetic dataset of~\citet{yuksel2009hair} from which we use two medium-length hairstyles: straight and curly.
To render RGB images of the dataset we use Unreal Engine instead of Blender in order to remove possible bias our method might have towards the path-traced images.
The accuracy and precision results are reported in~\cref{tab:quantitative_cem}.

Additionally, we note that current synthetic datasets do not contain enough variety of hairstyles and propose a new synthetic dataset containing 10 representative hairstyles created by our Blender pipeline. We also evaluate our method and the baselines on this dataset and report the results in~\cref{tab:quantitative} although we recognize that 
our method has an advantage because it is trained on data created by the same Blender pipeline. 

Across the two datasets we observe that our method can reconstruct the backside of the hair more faithfully, reinforcing the importance of varied data from which to learn a powerful prior.
For the backside reconstruction, robustness to viewpoint, and more visualizations see \textbf{SupMat}.

\subsection{Runtime Performance}
We evaluate the runtime performance in Tab.~\ref{tab:quantitative_compare_time} by first dividing the hair reconstruction pipeline into three parts, 1) extraction of RGB features (DINOv2 features or in the case of the baseline methods: orientation maps, hair segmentation, etc.), 2) generation of the hair model, which in our case is the scalp texture while baseline methods create a 3D orientation field, 3) strand generation, which is the final step to either decode or grow the strands.
All results are obtained on an NVIDIA H100 GPU. 
We observe that our method is overall the fastest as we do not need pre- or post-processing steps like head fitting, hair segmentation or hair growth and refinement. Note that our generated hair can be directly imported into Unreal Engine and rendered in real-time (60FPS) as shown in~\cref{fig:hair_in_unreal}; see \textbf{Supplemental Video.}

\subsection{Ablation Study} \label{ablation}

We perform an ablation study on the architecture and the importance of the conditioning signal in~\cref{fig:ablation_no_local_vae}.
Changing the scalp generation model to a VAE model similar to GroomGen~\cite{zhou2023groomgen} causes the hair to be overly smooth and lack local detail.
Our proposed diffusion model conditioned with global and local features in the form of $\featuremap$ and $\cls$ manages to capture the most detail. However, the local feature map is crucial and removing it causes the generated hair to deviate significantly from the input image.

\begin{figure}
    \centering
    \resizebox{\linewidth}{!}{
    \begin{tabular}{l rrr | rrr | rrr}
        \setlength{\tabcolsep}{0pt}
        & \multicolumn{9}{c}{\textbf{Thresholds: mm} $/$ \textbf{degrees}} \\
        \textbf{Method} & $2 / 20$ & $3 / 30$ & $4 / 40$ & $2 / 20$ & $3 / 30$ & $4 / 40$ & $2 / 20$ & $3 / 30$ & $4 / 40$ \\
        \cline{2-10}
        & \multicolumn{3}{c}{\textbf{Precision}($\uparrow$)} & \multicolumn{3}{c}{\textbf{Recall}($\uparrow$)} & \multicolumn{3}{c}{\textbf{F-score}($\uparrow$)} \\
        \hline
        NeuralHDHair~\cite{wu2022neuralhdhair}	&	\textbf{47.91}	& 65.07	&	73.76	&	31.59	&	51.10	&	64.63	&	38.00	&	57.12	&	68.73	\\
 HairStep\cite{zheng2023hairstep}	&	40.88	&	55.99	&	64.07	&	32.16	&	44.96	&	53.15	&	35.50	&	49.06	&	57.20	\\
 Ours	&	43.60	&	\textbf{68.30}	&	\textbf{81.79}	&	\textbf{34.97}	&	\textbf{58.27}	&	\textbf{71.79}	&	\textbf{38.35}	&	\textbf{62.41}	&	\textbf{76.20}	\\

    \end{tabular}
    }
    \vspace{-0.3cm}

    \captionof{table}{Quantitative comparison with \cite{wu2022neuralhdhair,zheng2023hairstep} on~\cite{yuksel2009hair}. It includes straight and curly hairstyles, please refer to the \textbf{SupMat} for more visualizations.}
    \label{tab:quantitative_cem}
     \vspace{-0.4cm}
\end{figure}

\begin{figure}
    \centering
    \resizebox{\linewidth}{!}{
    \begin{tabular}{l rrr | rrr | rrr}
        \setlength{\tabcolsep}{0pt}
        & \multicolumn{9}{c}{\textbf{Thresholds: mm} $/$ \textbf{degrees}} \\
        \textbf{Method} & $2 / 20$ & $3 / 30$ & $4 / 40$ & $2 / 20$ & $3 / 30$ & $4 / 40$ & $2 / 20$ & $3 / 30$ & $4 / 40$ \\
        \cline{2-10}
        & \multicolumn{3}{c}{\textbf{Precision}($\uparrow$)} & \multicolumn{3}{c}{\textbf{Recall}($\uparrow$)} & \multicolumn{3}{c}{\textbf{F-score}($\uparrow$)} \\
        \hline
        NeuralHDHair~\cite{wu2022neuralhdhair}	&	18.65	& 36.21	&	49.46	&	19.26	&	37.33	&	51.60	&	18.55	&	35.76	&	49.30	\\
 HairStep\cite{zheng2023hairstep}	&	20.22	&	35.54	&	45.73	&	20.80	&	38.53	&	52.82	&	19.42	&	34.38	&	45.48	\\
 Ours	&	\textbf{63.06}	&	\textbf{84.16}	&	\textbf{92.78}	&	\textbf{57.91}	&	\textbf{79.71}	&	\textbf{89.81}	&	\textbf{60.13}	&	\textbf{81.68}	&	\textbf{91.17}	\\

    \end{tabular}
    }
    \vspace{-0.3cm}
    \captionof{table}{Quantitative comparison with \cite{wu2022neuralhdhair,zheng2023hairstep} on the DiffLocks evaluation set. 
    Our results show substantial improvement over the baseline, particularly when reconstructing the backside of the hair due to our model's ability to learn a powerful prior over hairstyles.}
    
    \label{tab:quantitative}
     \vspace{-0.2cm}
\end{figure}

\begin{table}[!t]
\centering
\scriptsize
    \begin{tabular}{r   | c c c }
     & \shortstack{HairStep} & \shortstack{NeuralHDHair} & \shortstack{Ours} \\
     \hline
     RGB features ($\downarrow$)   & 3.17  & 4.86 & \textbf{0.10}\\
     hair model  ($\downarrow$)   & \textbf{0.45}  & 4.99 & 1.83 \\
     strand creation  ($\downarrow$)   & 49.78  & 8.47 & \textbf{0.40} \\
     total (s) ($\downarrow$)   & 53.40 &  18.32 &  \textbf{2.33}\\
     \end{tabular}
     \vspace{-0.1in}
         \caption{ \text{Timing.}
    Comparison of the time in seconds needed for each part of the hair creation pipeline.}
    \label{tab:quantitative_compare_time}
\end{table}

\section{Limitations and future work}\label{limitations}
Despite the large amount of data, there are still gaps in the synthetic dataset; e.g.~it lacks hairstyles with braids, accessories, ponytails, and buns. 
However, we believe that further improvements to the pipeline will allow such details to be created. 
Furthermore, the generation pipeline can also be easily extended to beards and eyebrows.

We also note that DiffLocks prioritizes local hair details over perfect image alignment. However, for many tasks, generation or reconstruction is just the first step. When hair is animated with physics, the precise initial strand alignment is not critical, as the dynamic nature of hair makes the initial configuration transient.

\section{Conclusion}\label{conclusion} 
We proposed a novel approach for generating 3D hair strands from a single image by first creating the largest dataset to date of synthetic hair paired with RGB images, and then by proposing an image-conditioned diffusion model based on scalp textures. Our method can create highly intricate 3D strands, surpassing previous methods and without requiring heuristic-based methods to increase the realism.

\noindent\textbf{Acknowledgments and Disclosures.}
We give thanks to Nathan Bajandas for all the Unreal images and videos.
While MJB is a co-founder and Chief Scientist at Meshcapade, his research in this project was performed solely at, and funded solely by, the Max Planck Society.

{
    \small
    \bibliographystyle{ieeenat_fullname}
    \bibliography{main}
}
\maketitlesupplementary


\section{Implementation details.}


\subsection{Strand VAE}
We train the strandVAE using a batch size of \num{200} strands, each consisting of \num{256} points. 
For the loss we set the directional weight term to $\lambda_1=2e-3$, the curvature weight term to $\lambda_2=7.8e-2$ and the KL term to $\lambda_\textrm{KL}=6e-4$.
We train the StrandVAE for a total of 3M iterations using AdamW at a learning rate of $3e-3$ with cosine decay schedule.

\subsection{Diffusion model}
We follow similar optimizer parameters as HDiT~\cite{hdit} and train the diffusion model for $\approx 400$K iterations at an effective batch size of \num{128} using a constant learning rate of $5e-4$.

\section{Dataset.}
The synthetic hair dataset was created by launching 12 Blender processes in parallel, each creating a chunk of the $40$K samples. The process is particularly CPU-intensive since the Blender geometry nodes don't tend to take advantage of GPU acceleration. As a result, the generation process took approximately 1 week on a cluster with 4 H100 GPUs and a Intel Xeon Platinum 8480+ CPU (56 Cores).

\section{Discussions}
\subsection{Density Map}
\begin{figure}[h]

	\centering
	\begin{tikzpicture}[  ]

     \def\xshift{1.95}
     \def\yshift{-0.25}



    \node[inner sep=0pt, anchor=south west] (base0) at (-1.2+\xshift*0, 0) {\includegraphics[width=.12\textwidth]{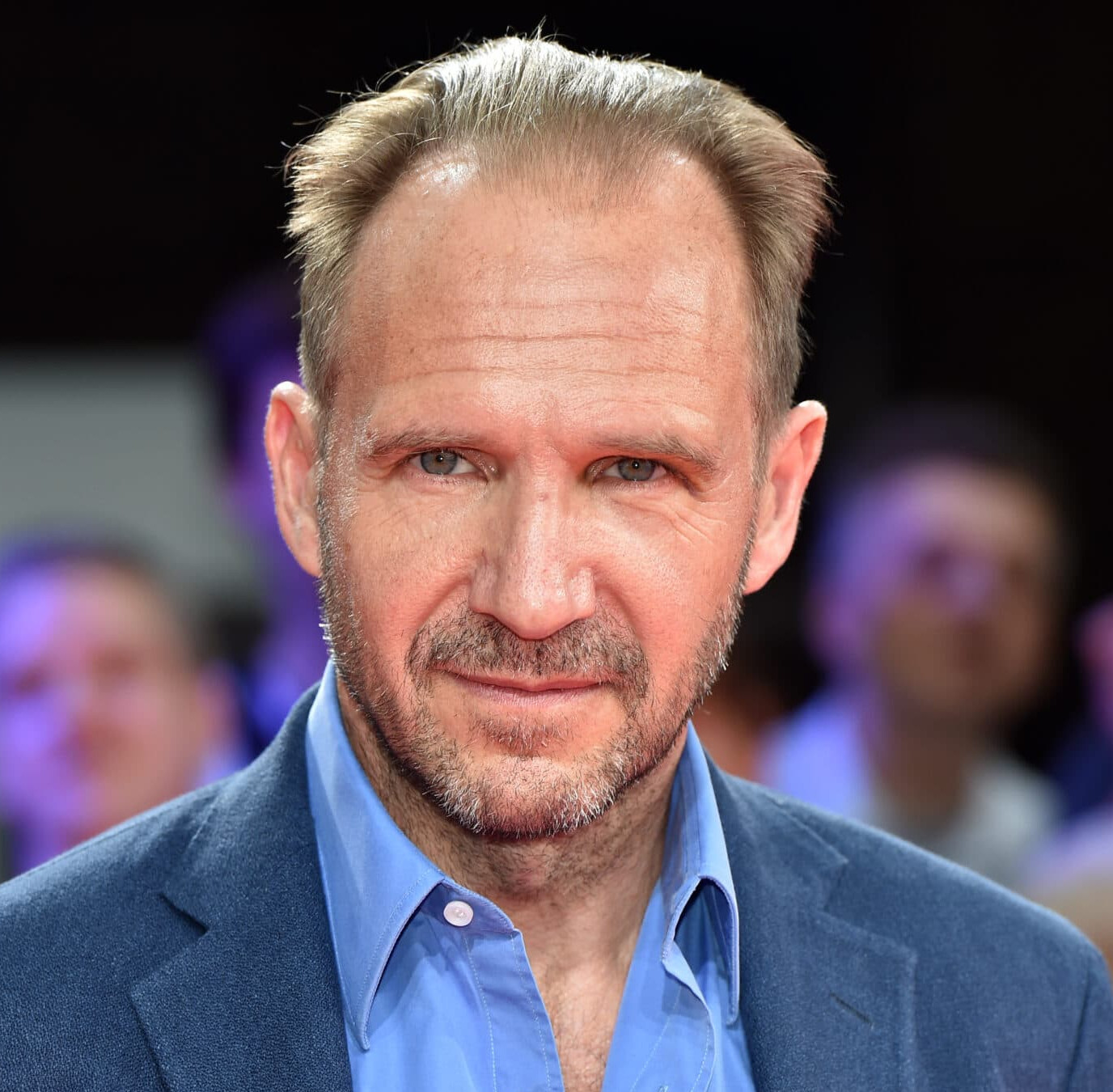}};
    \node[inner sep=0pt, anchor=south] (def) at (-0.1+\xshift*1, 0) {\includegraphics[width=.12\textwidth]{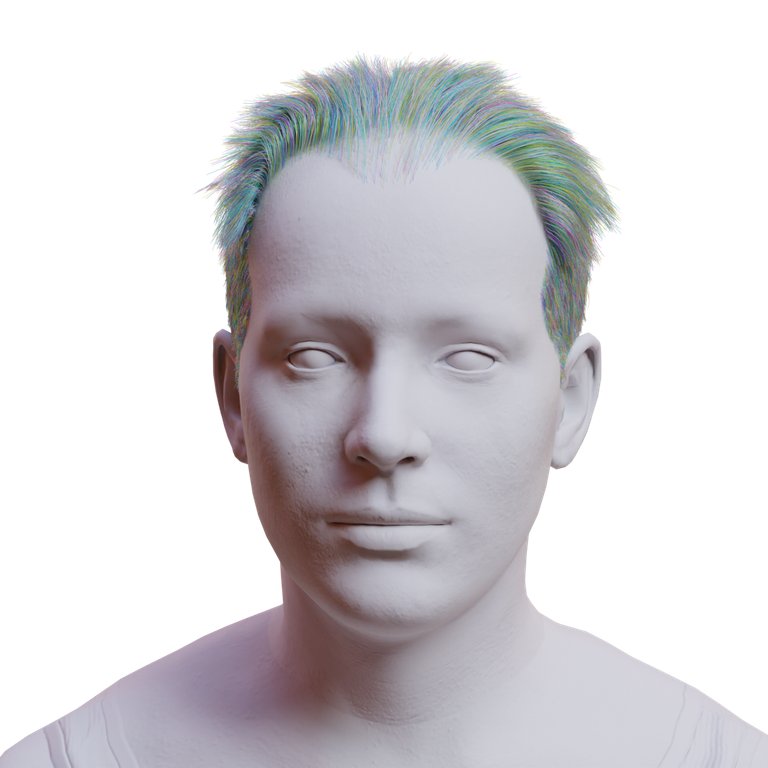}};
    \node[inner sep=0pt, anchor=south] (def2) at (\xshift*2, 0) {\includegraphics[width=.12\textwidth]{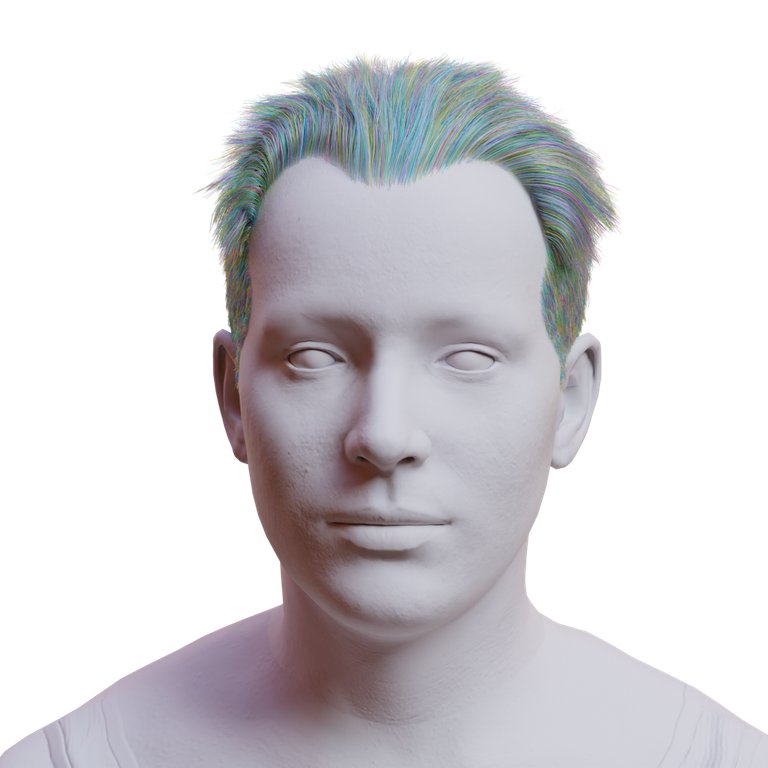}};
    \node[inner sep=0pt, anchor=south] (def3) at (\xshift*3, 0) {\includegraphics[width=.12\textwidth]{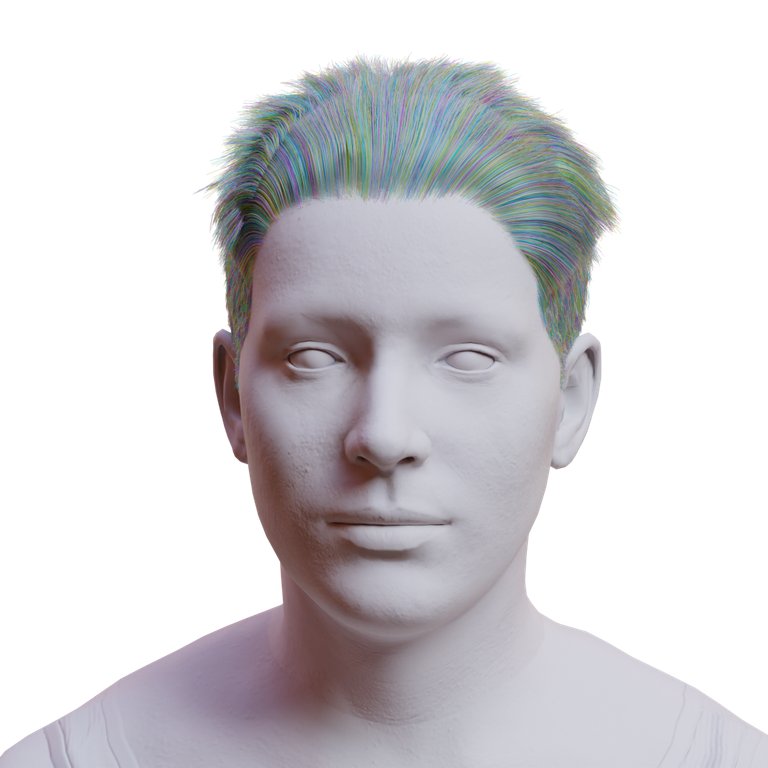}};
    \node[font=\footnotesize\selectfont, align=center, below = -0.0cm of base0] {RGB Input};
    \node[font=\footnotesize\selectfont, align=center, below = -0.0cm of def] {Generated hair};
    \node[font=\footnotesize\selectfont, align=center, below = -0.0cm of def2] {Increased density};
    \node[font=\footnotesize\selectfont, align=center, below = -0.0cm of def3] {Changed map};



	\end{tikzpicture}
	\vspace{-0.1in}
	\caption{The density map allows the density of hair or hairline to be easily controlled by directly modifying its values.}
	\label{fig:density_map}
\end{figure}

Compared to GroomGen’s binary map, we treat the proposed density map as an alternative hair representation with different properties. GroomGen's binary mask marks exact scalp pixels where strands grow, while our focus is on local hair density rather than precise root positions. This makes the density map smoother 
and easier to edit—since strand latent codes exist for every pixel, we can modify hair density by simply sampling more strands~(\cref{fig:density_map}).
\vspace{-0.05in}

\section{Additional Evaluation and Results}

\input{fig/robustness_to_camera_views}

\begin{figure}
\captionsetup[subfigure]{labelformat=empty, justification=centering}
     \centering
     \begin{subfigure}[t]{0.2\textwidth}
         \centering
         \includegraphics[width=\textwidth]{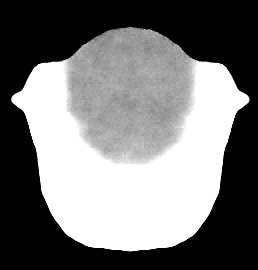}
         \caption{w/o channel weighting}
     \end{subfigure}
     \hspace{-0.1em}
     \begin{subfigure}[t]{0.2\textwidth}
         \centering
         \includegraphics[width=\textwidth]{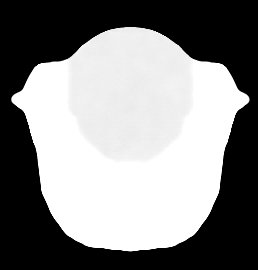}
         \caption{w/ channel weighting}
     \end{subfigure}
    \caption{\textbf{Ablation channel weighting}. 
    Without channel weighting, the diffusion model tends to generate noisy density maps. Applying our proposed weighing, the network focuses on the important information from the scalp textures and allowes it to create smoother density maps. 
    }
    \label{fig:weighting_ablation}
\end{figure}

\subsection{Additional studies}
\paragraph{Camera pose robustness.} To further demonstrate the robustness of our method, we evaluate it on images captured from varying camera poses from the H3DS dataset~\cite{ramon2021h3d}. As shown in \cref{fig:robustness_camera_views}, our approach can reconstruct consistent hairstyles despite large changes in camera position and distance to the subject. 

\begin{figure}
    \centering
    \includegraphics[width=0.35\textwidth]{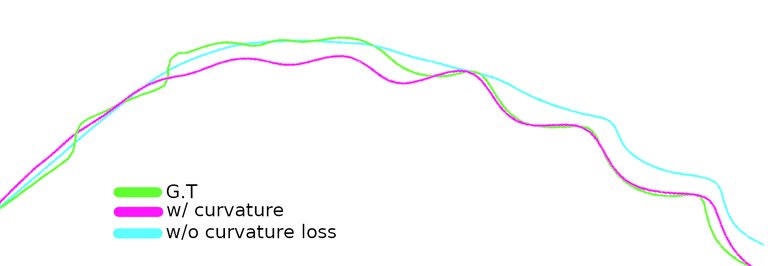}
    \caption{
    \textbf{StrandVAE ablation}. We observe that without the curvatuer loss, the predicted strand (blue) tends to be smoother than the ground-truth(green). However, with the curvature loss(purple), the curly pattern is more accurately recovered, improving the visual quality of the whole hairstyle.
    }
    \label{fig:strand_vae_ablation}
\end{figure}

\paragraph{Ablation on curvature loss.} To evaluate the effectiveness of curvature loss while training the strandVAE, we perform an ablation study as shown in~\cref{fig:strand_vae_ablation} in which we encode the ground-truth strand and decode it with models trained with and without the curvature los. We observe that the curvature loss is crucial in encouraging the network to reconstruct curly and wavy hair.

\paragraph{Ablation on weighting scheme.} We also ablate our proposed per-channel weighting scheme in~\cref{fig:weighting_ablation}. It is clear that without the weighting scheme, the output of the diffusion model exhibits more noise which is especially evident when viewing the density map. Our proposed weighting scheme allows the network to focus on the important information stored in the latent code of the strands and ignore noisy dimensions.

\subsection{Generalization ability}
\begin{figure}[h]

	\centering
	\begin{tikzpicture}[  ]

     \def\xshift{1.55}
     \def\yshift{-0.25}


    \node[inner sep=0pt, anchor=south west] (base0) at (-1.2+\xshift*0, 0) {\includegraphics[width=.10\textwidth]{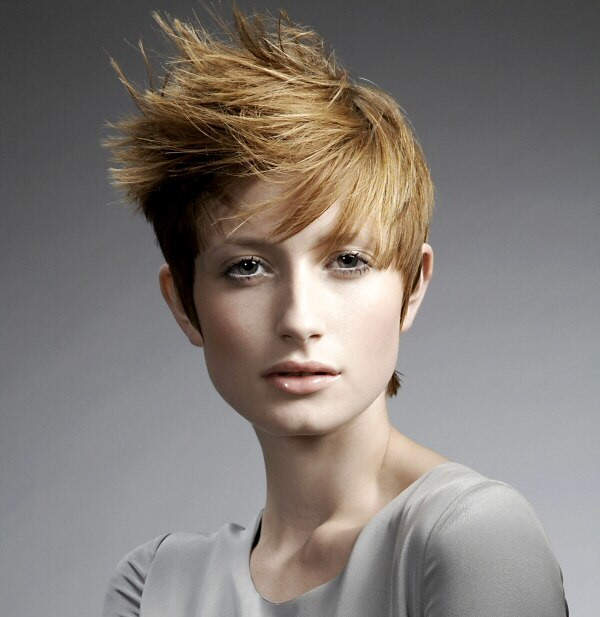}};
    \node[inner sep=0pt, anchor=south] (def) at (-0.1+\xshift*1, 0) {\includegraphics[width=.10\textwidth]{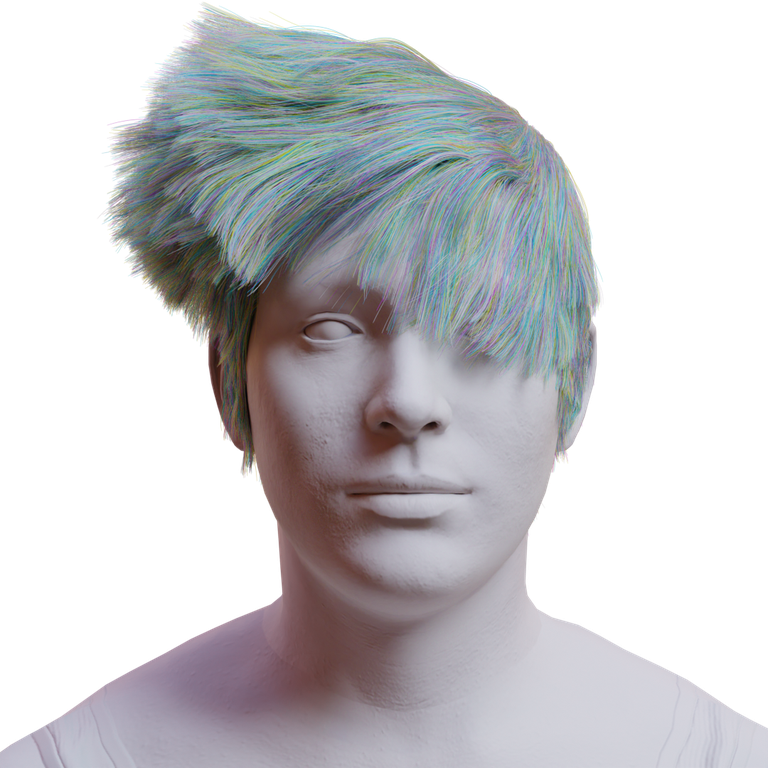}};
    
    \node[inner sep=0pt, anchor=south] (t1) at (0.1+\xshift*2, 0) {\includegraphics[width=.10\textwidth]{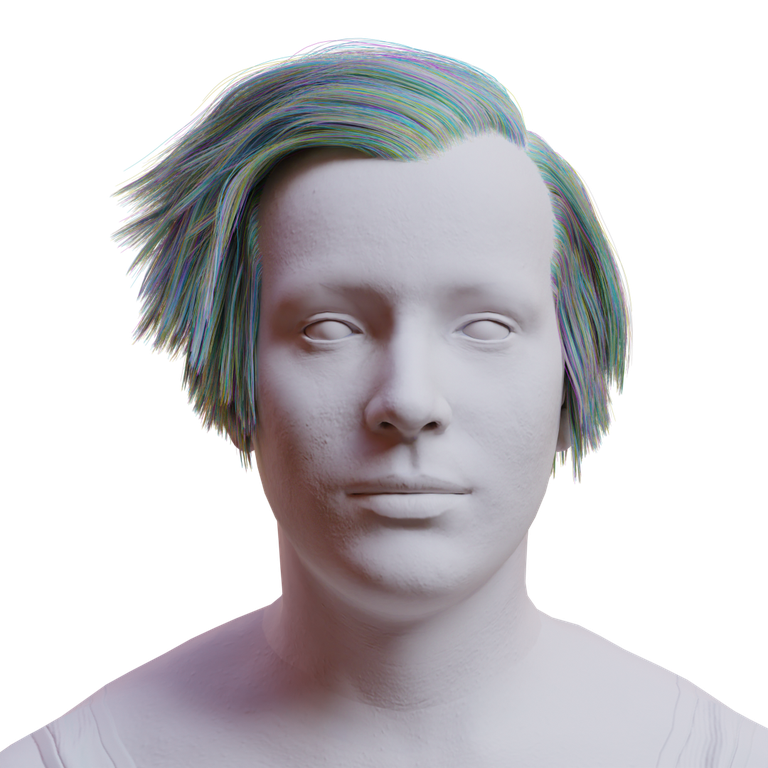}};
     \node[inner sep=0pt, anchor=south] (t2) at (\xshift*3, 0) {\includegraphics[width=.10\textwidth]{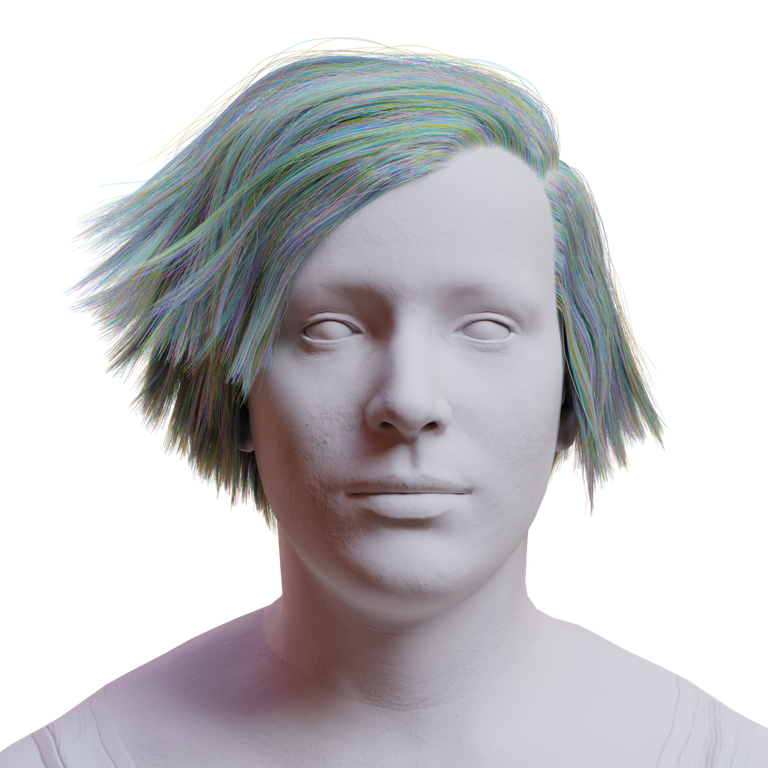}};
    \node[inner sep=0pt, anchor=south] (t3) at (-0.1+\xshift*4, 0) {\includegraphics[width=.10\textwidth]{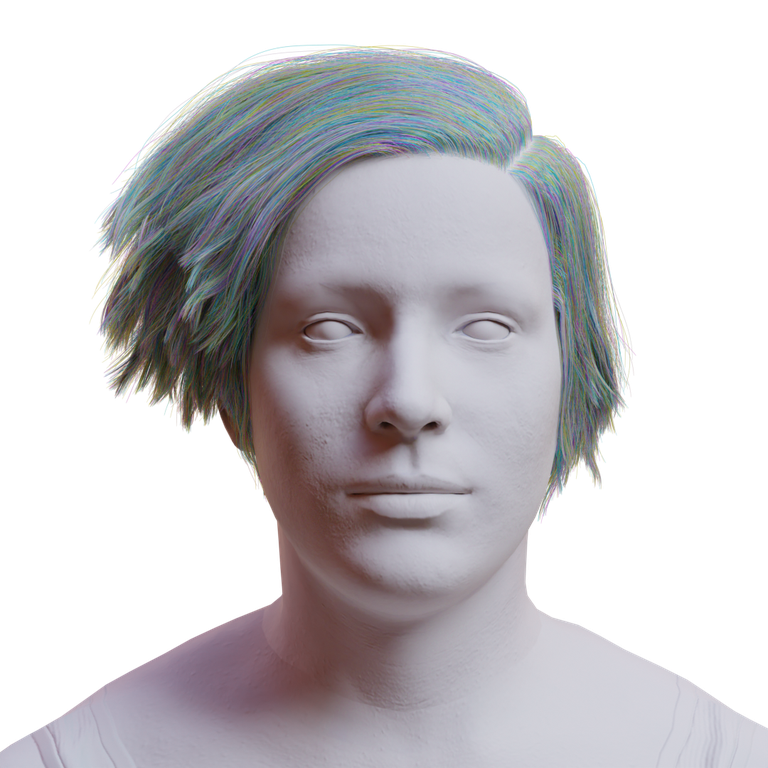}};
     
    \node[font=\footnotesize\selectfont, align=center, below = -0.0cm of base0] {RGB Input};
    \node[font=\footnotesize\selectfont, align=center, below = -0.0cm of def] {Generated hair};
    
    \node[font=\footnotesize\selectfont, align=center, below = -0.0cm of t1] {Top-1};
    \node[font=\footnotesize\selectfont, align=center, below = -0.0cm of t2] {Top-2};
    \node[font=\footnotesize\selectfont, align=center, below = -0.0cm of t3] {Top-3};



	\end{tikzpicture}
	\vspace{-0.1in}
	\caption{DiffLocks generalizes beyond the training set: Generated hair and the closest top\-3 hairstyles from the training set.}
	\label{fig:generalization}
\end{figure}

We test reconstructing a hairstyle outside the training set and find the network generalizes well~(\cref{fig:generalization}). The significant gap between the generated hair and the top-3 samples from the training set confirms that our method generates new hairstyles rather than merely retrieving from training data.
\unskip

\subsection{DINOv2 vs orientation map}
\begin{figure}[h]

	\centering
	\begin{tikzpicture}[  ]

     \def\xshift{1.8}
     \def\yshift{0.0}



    \node[inner sep=0pt, anchor=south west] (base0) at (-1.2+\xshift*0, 0) {\includegraphics[trim={6cm 5cm 6cm 1cm},clip,width=.085\textwidth]{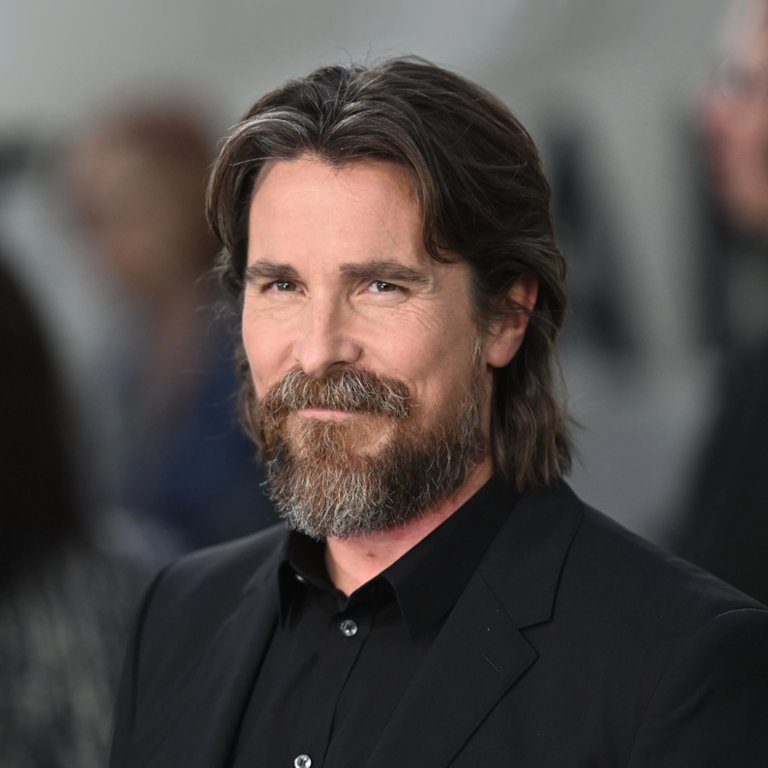}};

    \node[inner sep=0pt, anchor=south] (h1) at (-0.1+\xshift*1, \yshift) {\includegraphics[trim={6cm 5cm 6cm 3cm},clip,width=.095\textwidth]{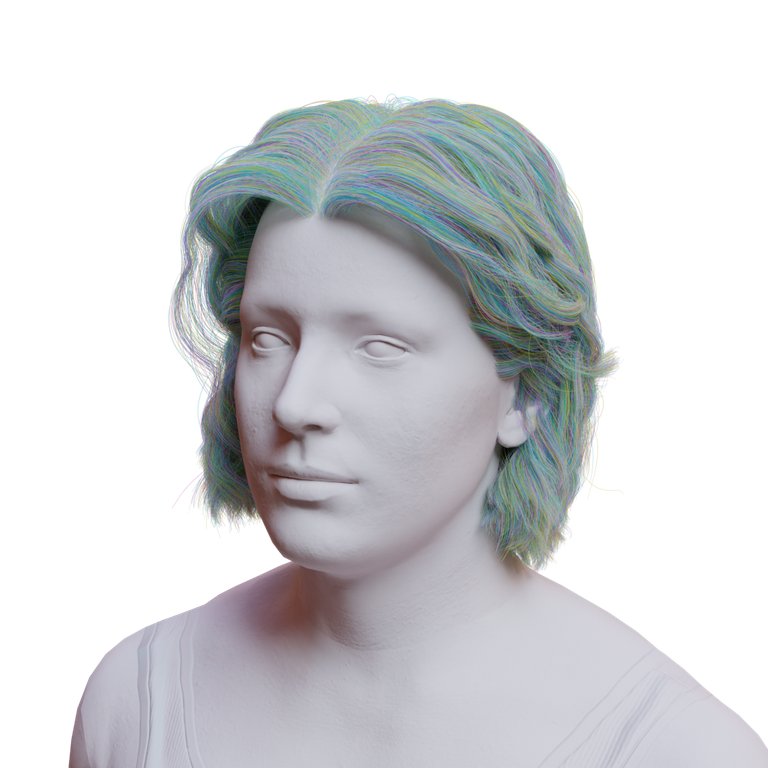}};
    \node[inner sep=0pt, anchor=south] (h2) at (-0.1+\xshift*2, \yshift) {\includegraphics[trim={6cm 5cm 6cm 3cm},clip,width=.095\textwidth]{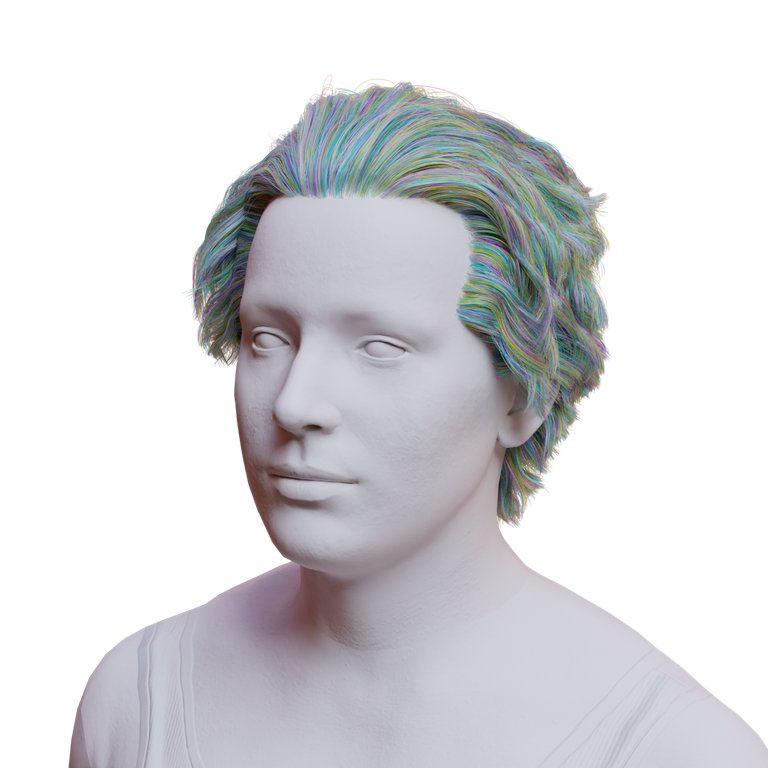}};

    \node[font=\footnotesize\selectfont, align=center, below = -0.0cm of h1] {DINOv2};
    \node[font=\footnotesize\selectfont, align=center, below = -0.0cm of h2] {Orientation map};




	\end{tikzpicture}
	\vspace{-0.15in}
	\caption{DINOV2 features are a more robust and richer conditioning signal than orientation maps.}
	\label{fig:dino_vs_ori}
\end{figure}

We ran an experiment where we replace DINO features with an orientation map together with hair segmentation mask. We find that DINO features are more robust especially for short or dark hair~(\cref{fig:dino_vs_ori}) where the orientation map can be noisy.
\vspace{-0.05in}
\unskip

\subsection{Extended comparison with baselines} 

We provide extended quantitative comparison and qualitative comparison with HairStep~\cite{zheng2023hairstep} and NeuralHDHair~\cite{wu2022neuralhdhair} on the DiffLocks evaluation dataset and Yuksel dataset~\cite{yuksel2009hair}.
We note that the DiffLocks evaluation set is created using our proposed Blender pipeline and contains samples that were not used during training.

\begin{figure*}[t]

	\centering
	\begin{tikzpicture}[ remember picture, >={Stealth[inset=1pt,length=8pt,angle'=30,round]} ]

     \def\xshift{2.7}
     \def\yshift{-3.0}
     \def\w{0.17}
     \def\firstshift{-1.0}
     \def\YshiftBlock{-0.4}

     \def\cropBottom{0.1cm}

    \node[] (origin) at (0,0) {};

    \node[inner sep=0pt] (straight) at (\firstshift+\xshift*0, 0) {\adjincludegraphics[width=\w\textwidth]{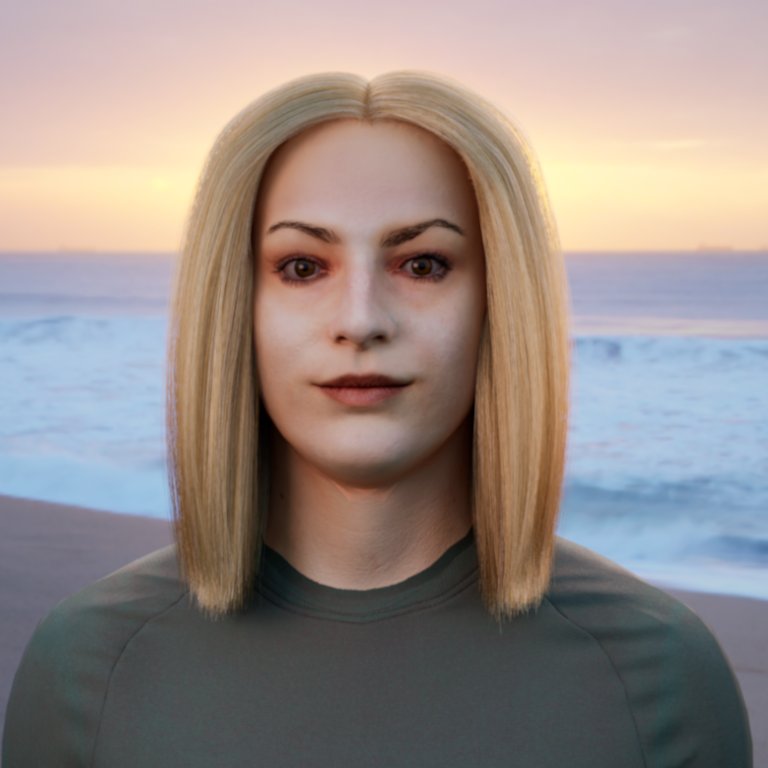}};
    \node[inner sep=0pt] (ours_straight_front) at (\xshift*1, 0) {\adjincludegraphics[width=\w\textwidth]{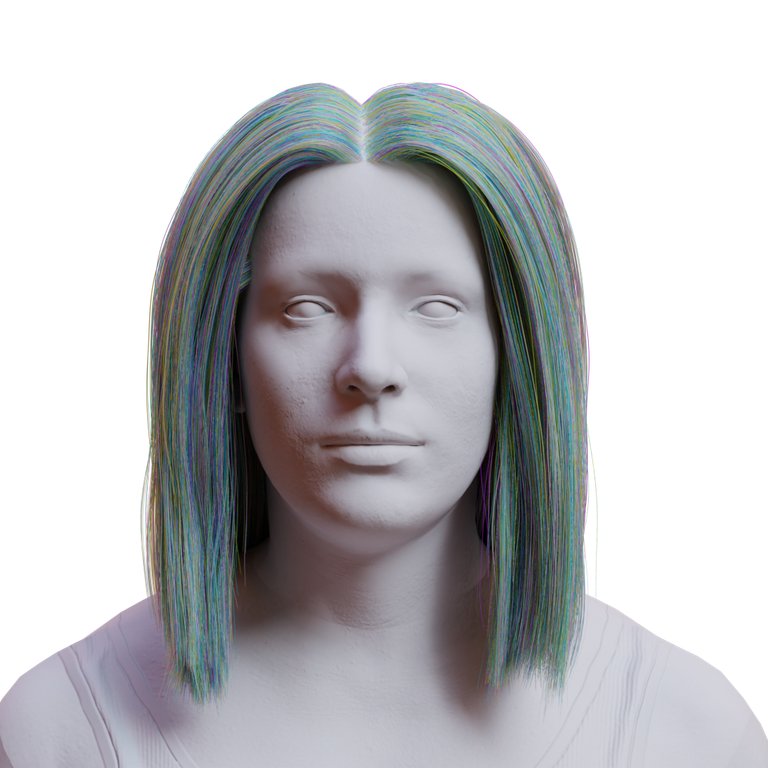}};
    \node[inner sep=0pt] (freeman_rgb) at (\xshift*2, 0) {\adjincludegraphics[width=\w\textwidth]{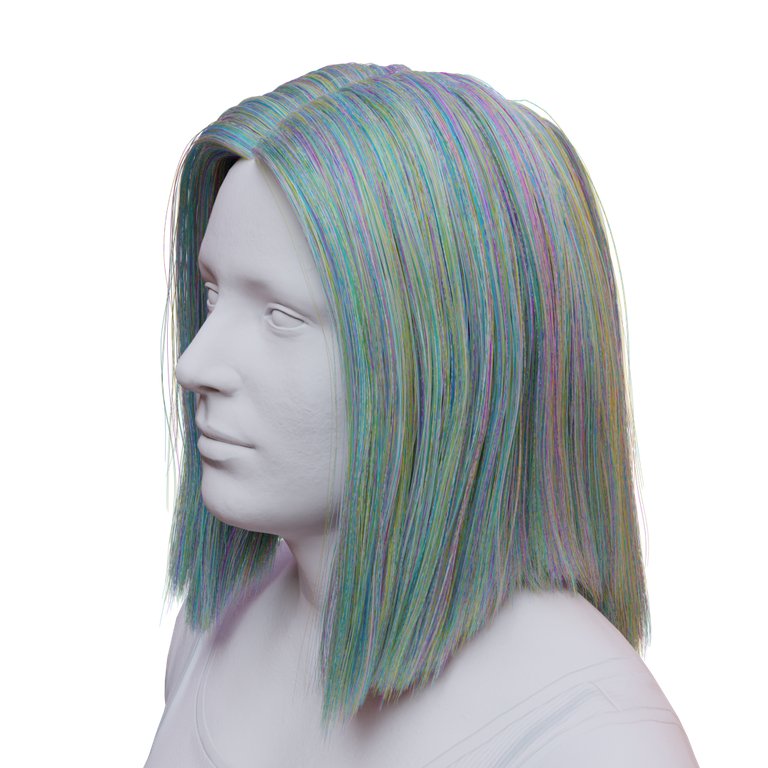}};
    \node[inner sep=0pt] (freeman_rgb) at (\xshift*3, 0) {\adjincludegraphics[width=\w\textwidth]{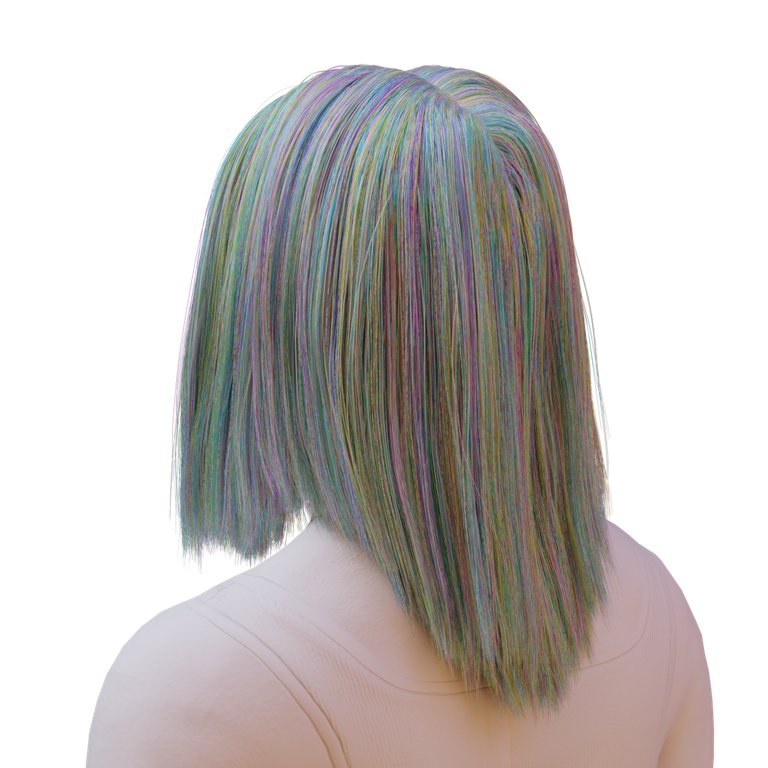}};
    \node[inner sep=0pt] (hairstep_straight_front) at (\xshift*1, \yshift*1) {\adjincludegraphics[width=\w\textwidth]{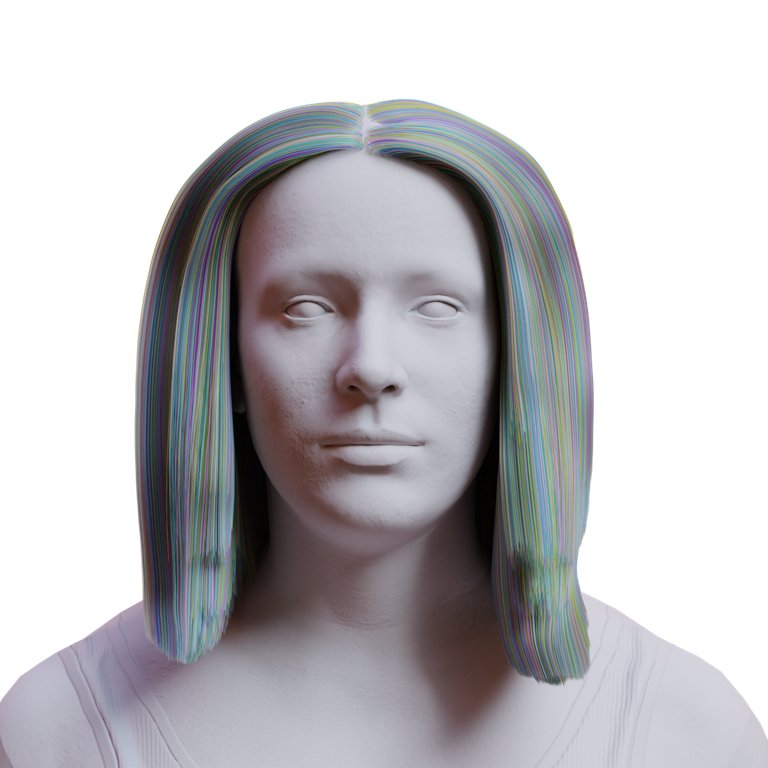}};
    \node[inner sep=0pt] (freeman_rgb) at (\xshift*2, \yshift*1) {\adjincludegraphics[width=\w\textwidth]{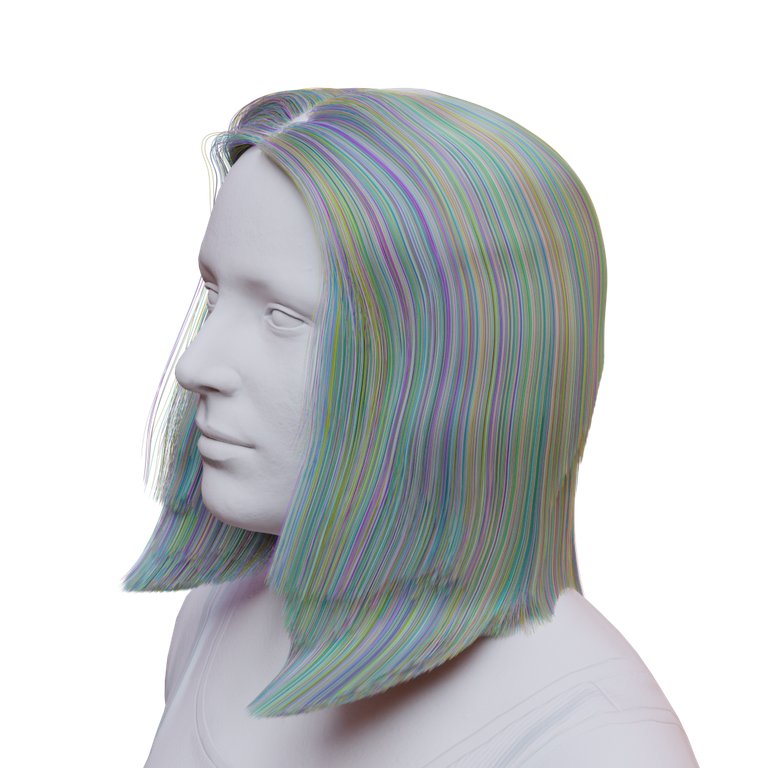}};
    \node[inner sep=0pt] (freeman_rgb) at (\xshift*3, \yshift*1) {\adjincludegraphics[width=\w\textwidth]{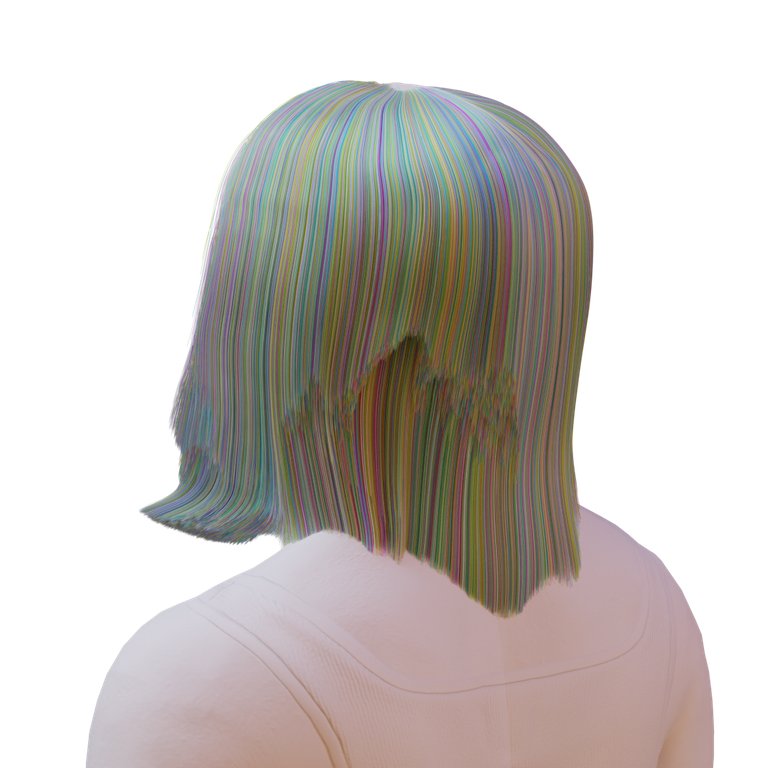}};
    \node[inner sep=0pt] (neuralhdhair_straight_front) at (\xshift*1, \yshift*2) {\adjincludegraphics[width=\w\textwidth]{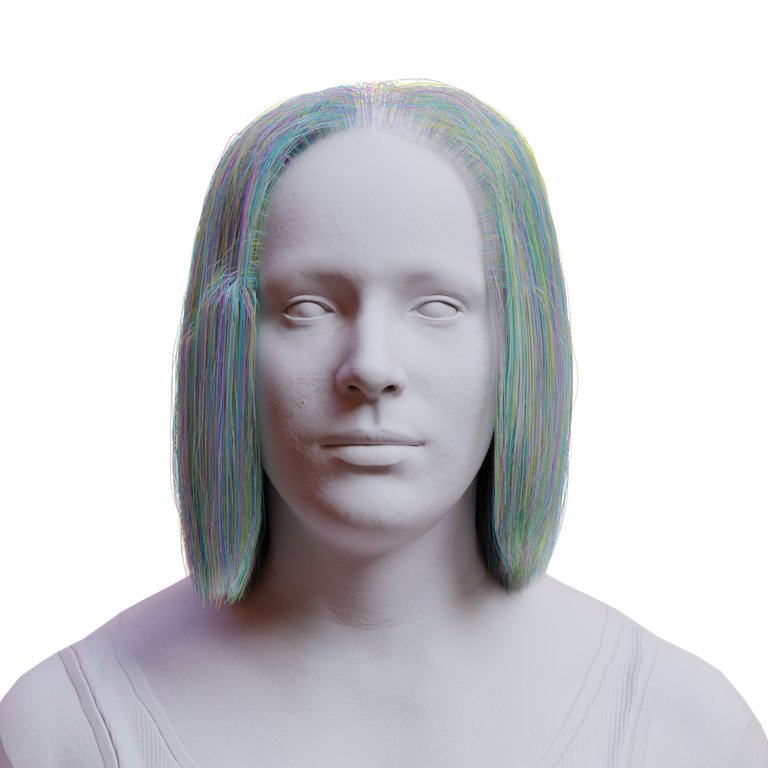}};
    \node[inner sep=0pt] (freeman_rgb) at (\xshift*2, \yshift*2) {\adjincludegraphics[width=\w\textwidth]{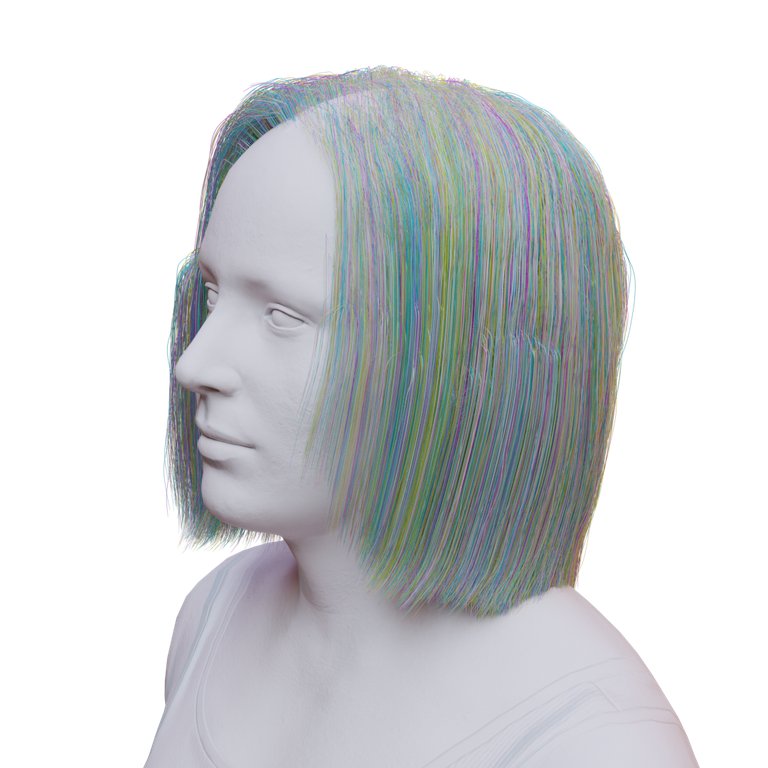}};
    \node[inner sep=0pt] (freeman_rgb) at (\xshift*3, \yshift*2) {\adjincludegraphics[width=\w\textwidth]{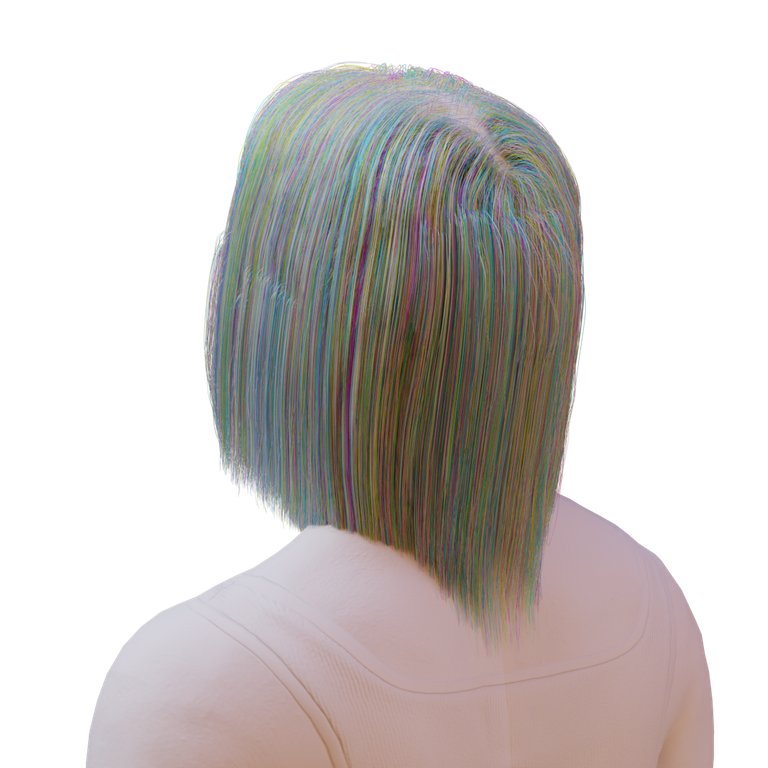}};

     \node[left=-0.0cm of ours_straight_front] (ours) {\rotatebox{90}{Ours}};
     \node[left=-0.0cm of hairstep_straight_front] (ours) {\rotatebox{90}{HairStep}};
      \node[left=-0.0cm of neuralhdhair_straight_front] (ours) {\rotatebox{90}{NeuralHDHair}};
    \node[below=-0.0cm of straight] (ours) {Straight};

    \node[inner sep=0pt] (curly) at (\firstshift+\xshift*0, \yshift*3+\YshiftBlock) {\adjincludegraphics[width=\w\textwidth]{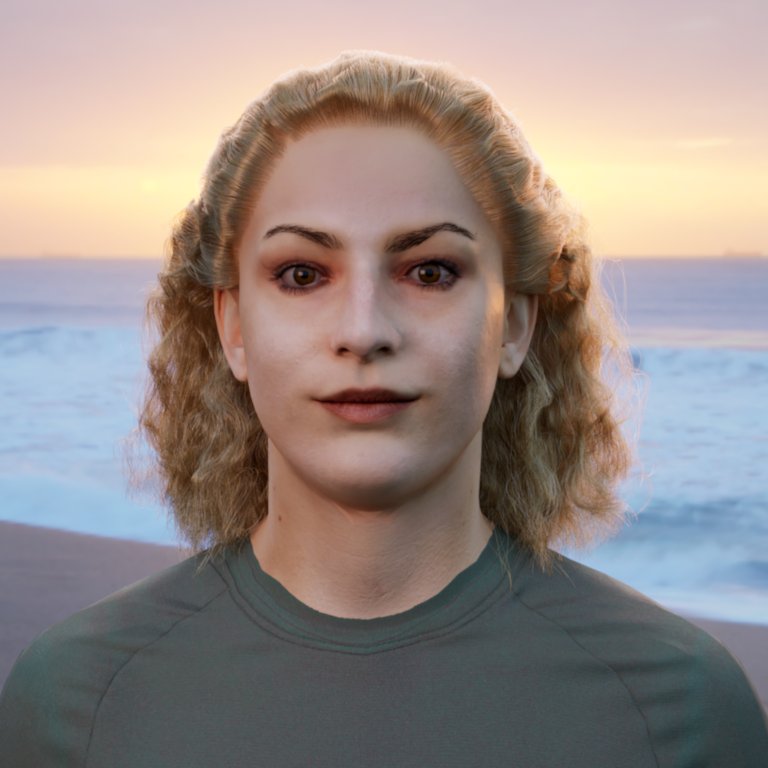}};
    \node[inner sep=0pt] (ours_curly_front) at (\xshift*1, \yshift*3+\YshiftBlock) {\adjincludegraphics[width=\w\textwidth]{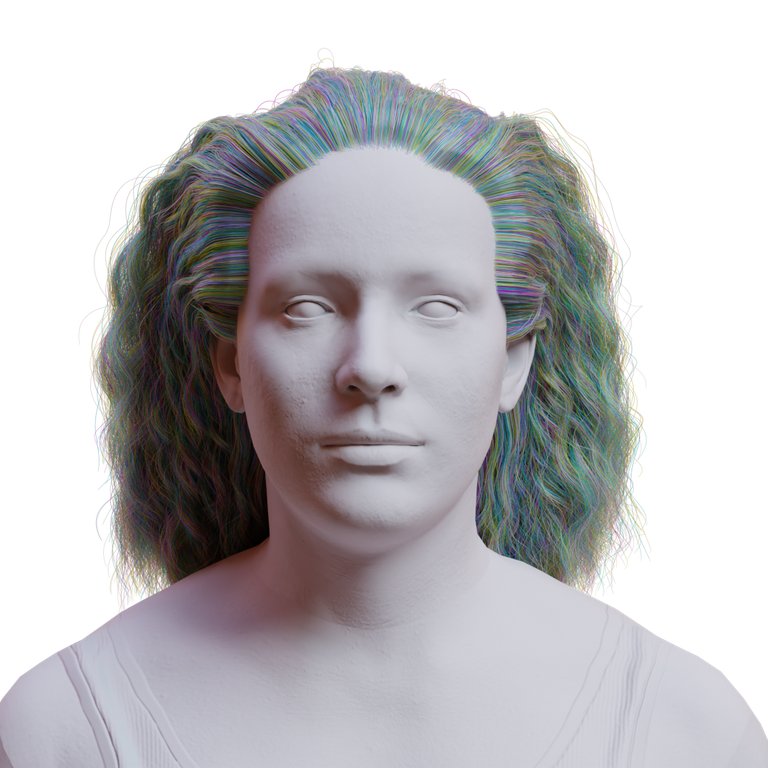}};
    \node[inner sep=0pt] (freeman_rgb) at (\xshift*2, \yshift*3+\YshiftBlock) {\adjincludegraphics[width=\w\textwidth]{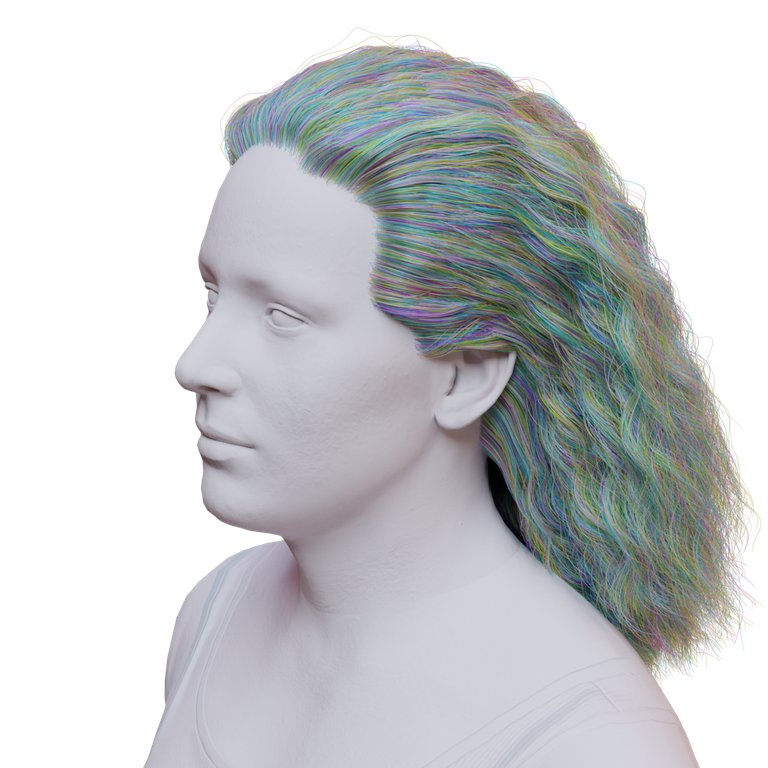}};
    \node[inner sep=0pt] (freeman_rgb) at (\xshift*3, \yshift*3+\YshiftBlock) {\adjincludegraphics[width=\w\textwidth]{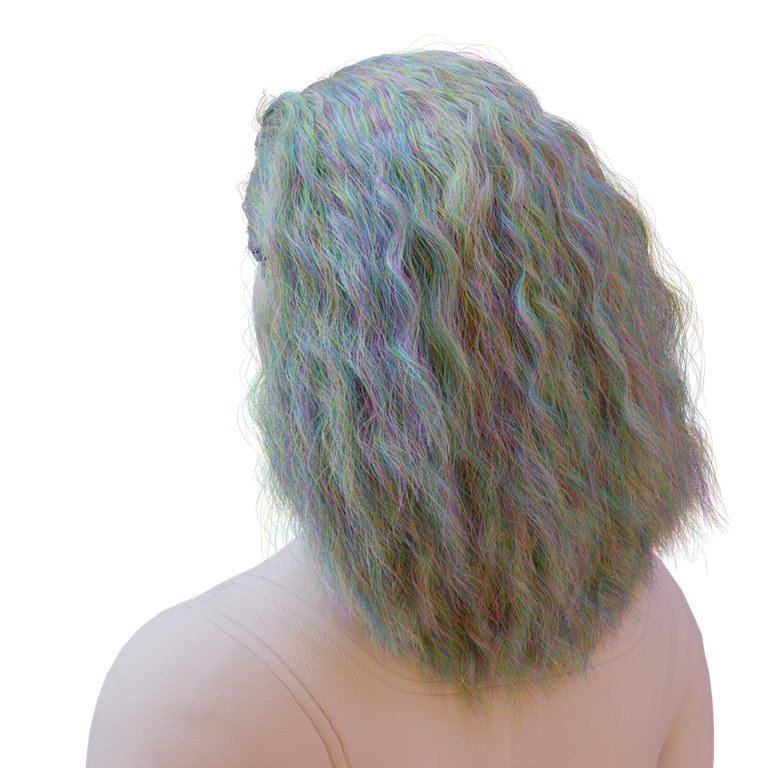}};
    \node[inner sep=0pt] (hairstep_curly_front) at (\xshift*1, \yshift*4+\YshiftBlock) {\adjincludegraphics[width=\w\textwidth]{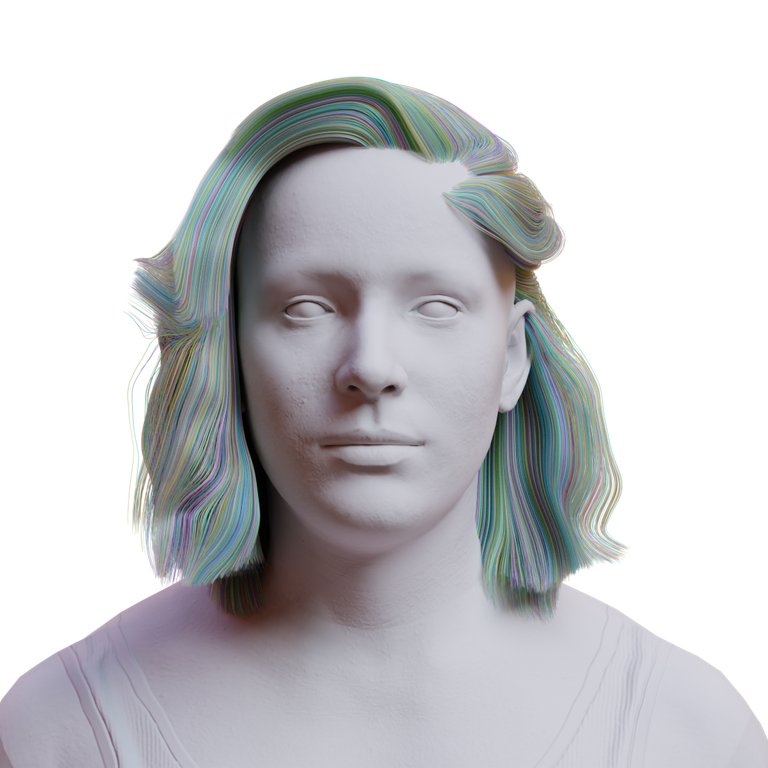}};
    \node[inner sep=0pt] (freeman_rgb) at (\xshift*2, \yshift*4+\YshiftBlock) {\adjincludegraphics[width=\w\textwidth]{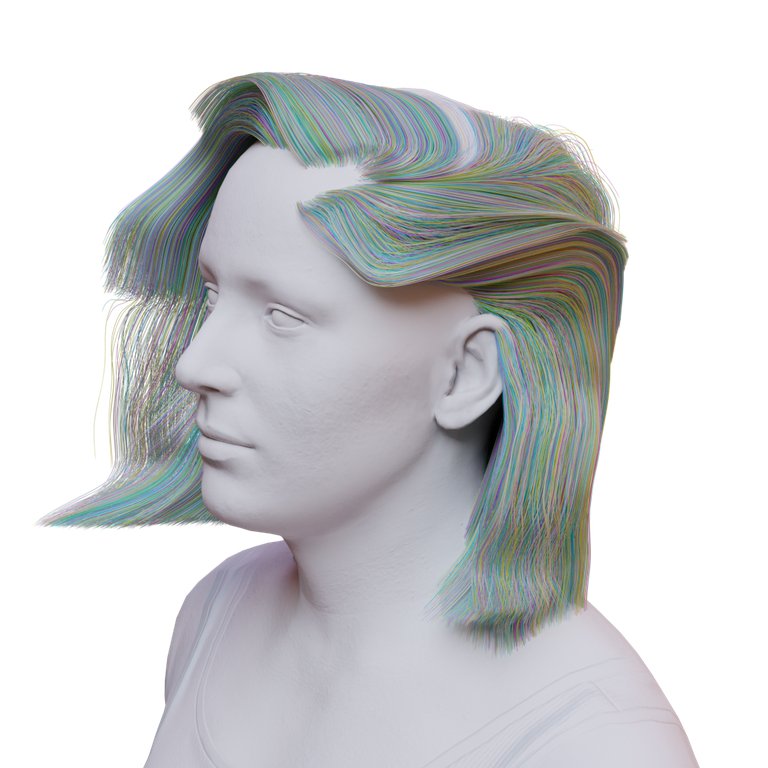}};
    \node[inner sep=0pt] (freeman_rgb) at (\xshift*3, \yshift*4+\YshiftBlock) {\adjincludegraphics[width=\w\textwidth]{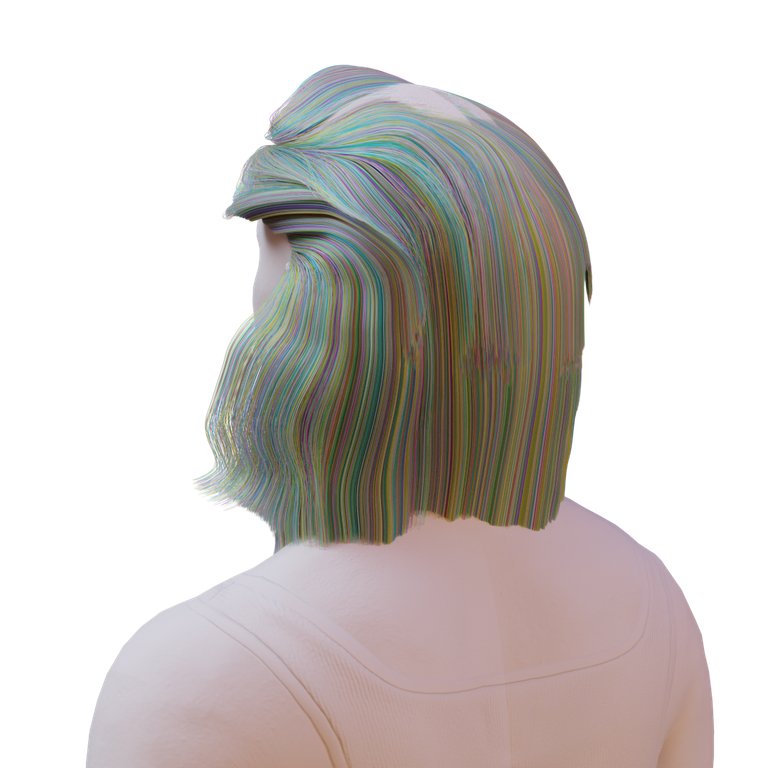}};
    \node[inner sep=0pt] (neuralhdhair_curly_front) at (\xshift*1, \yshift*5+\YshiftBlock) {\adjincludegraphics[width=\w\textwidth]{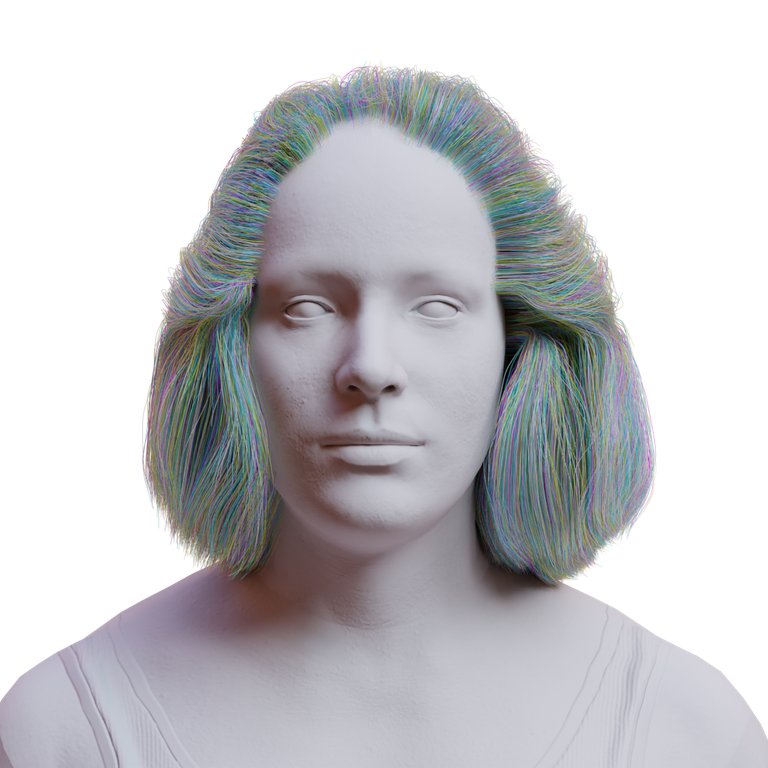}};
    \node[inner sep=0pt] (freeman_rgb) at (\xshift*2, \yshift*5+\YshiftBlock) {\adjincludegraphics[width=\w\textwidth]{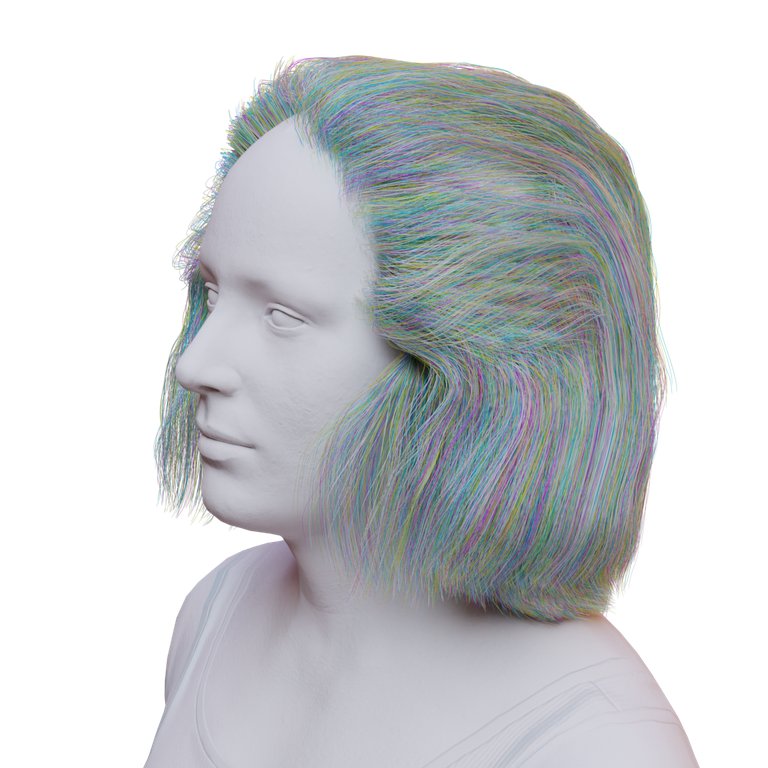}};
    \node[inner sep=0pt] (freeman_rgb) at (\xshift*3, \yshift*5+\YshiftBlock) {\adjincludegraphics[width=\w\textwidth]{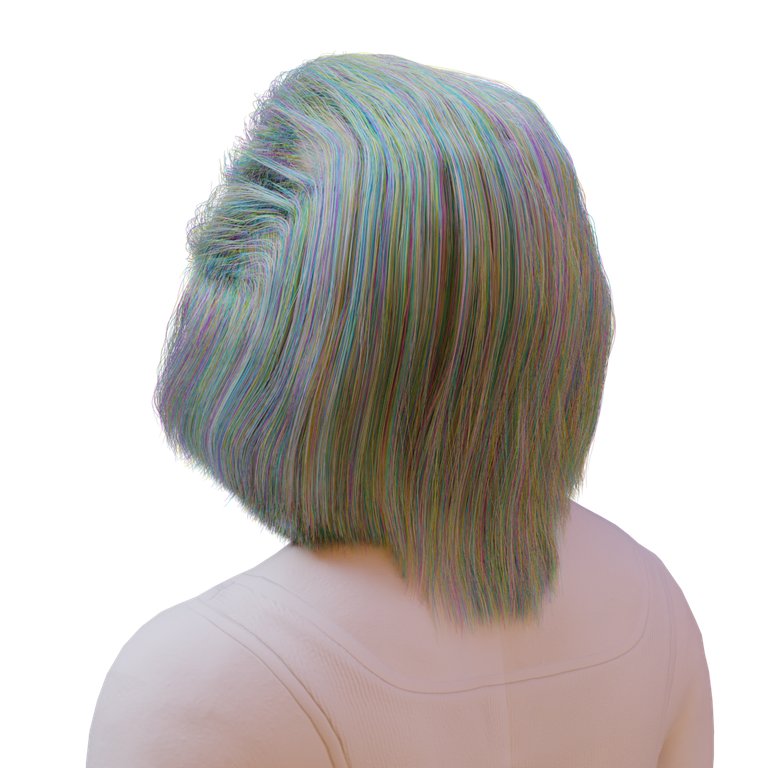}};

     \node[left=-0.0cm of ours_curly_front] (ours) {\rotatebox{90}{Ours}};
     \node[left=-0.0cm of hairstep_curly_front] (ours) {\rotatebox{90}{HairStep}};
      \node[left=-0.0cm of neuralhdhair_curly_front] (ours) {\rotatebox{90}{NeuralHDHair}};
     \node[below=-0.0cm of curly] (ours) {Curly};

	\end{tikzpicture}
	\caption{Results on the synthetic dataset of~\cite{yuksel2009hair}}
    \vspace{-0.05in}
	\label{fig:yuksel_results}
\end{figure*}

\paragraph{Qualitative.} The qualitative comparison with HairStep and NeuralHDHair is shown in~\cref{fig:yuksel_results}. Their method performs well on straight hairstyles but struggle with curly and wavy ones, especially when their predicted 2D orientation map~\cite{paris2004capture} represent an incorrect parting or growth direction.
The reliance on intermediate representations like orientations maps is a long-standing limitation of hair reconstruction. In contract, our method effectively reconstructs both straight and curly hairstyles, remaining unaffected by 2D orientation ambiguities, as it directly utilizes RGB images as input. 


An additional limitation of previous methods becomes evident when viewing the backside of the head where the baselines methods tend to create balding areas or overly smooth strands.
In contrast, our approach, powered by robust data priors, generates more reasonable and realistic results for occluded or invisible regions.

\begin{figure*}[htbp]
    \centering
    \setlength{\tabcolsep}{8pt} 
    \begin{minipage}{\textwidth}

    \begin{minipage}{0.12\linewidth}
        \includegraphics[width=\linewidth]{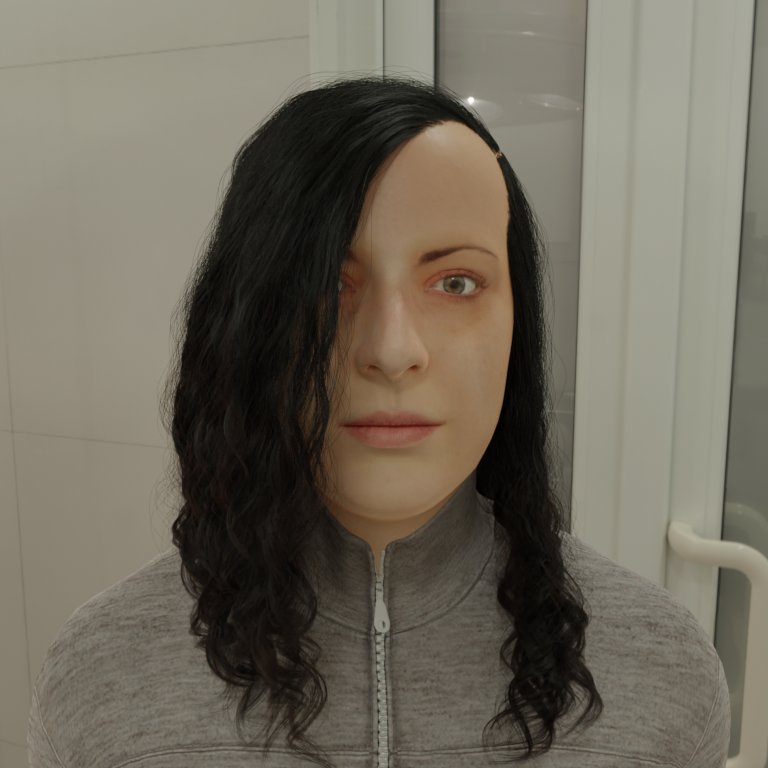} 
    \end{minipage}
    \hfill 
    \begin{minipage}{0.8\textwidth}
       
        \scriptsize 
        \begin{tabular}{l rrr | rrr | rrr}
            \setlength{\tabcolsep}{0pt}
            \renewcommand{\arraystretch}{3}
            & \multicolumn{9}{c}{\textbf{Thresholds: mm} $/$ \textbf{degrees}} \\
            \textbf{Method} & $2/20$ & $3/30$ & $4/40$ & $2/20$ & $3/30$ & $4/40$ & $2/20$ & $3/30$ & $4/40$ \\
            \cline{2-10}
            & \multicolumn{3}{c}{\textbf{Precision} ($\uparrow$)} & \multicolumn{3}{c}{\textbf{Recall} ($\uparrow$)} & \multicolumn{3}{c}{\textbf{F-score} ($\uparrow$)} \\
            \hline
            NeuralHDHair~\cite{wu2022neuralhdhair}	&	27.57	& 47.30	&	57.12	&	22.67	&	44.90	&	63.06	&	24.88	&	46.07	&	59.94	\\
         HairStep\cite{zheng2023hairstep}	&	32.10	&	52.78	&	65.90	&	20.04	&	35.55	&	47.71	&	24.67	&	42.48	&	55.35	\\
         Ours	&	\textbf{49.93}	&	\textbf{73.46}	&	\textbf{84.42}	&	\textbf{52.20}	&	\textbf{76.88}	&	\textbf{88.90}	&	\textbf{51.04}	&	\textbf{75.13}	&	\textbf{86.60}	\\
        \end{tabular}
    \end{minipage}

    \vspace{0mm}

    \begin{minipage}{0.12\linewidth}
        \includegraphics[width=\linewidth]{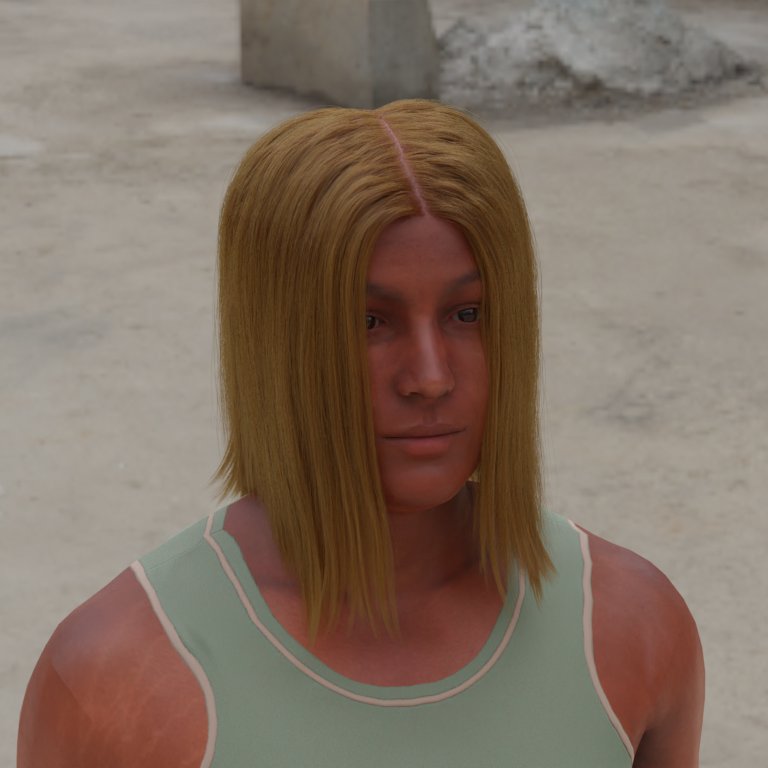} 
    \end{minipage}
    \hfill 
    \begin{minipage}{0.8\textwidth}
        \scriptsize 
        \begin{tabular}{l rrr | rrr | rrr}
            \setlength{\tabcolsep}{0pt}
            \renewcommand{\arraystretch}{3}
            & \multicolumn{9}{c}{\textbf{Thresholds: mm} $/$ \textbf{degrees}} \\
            \textbf{Method} & $2/20$ & $3/30$ & $4/40$ & $2/20$ & $3/30$ & $4/40$ & $2/20$ & $3/30$ & $4/40$ \\
            \cline{2-10}
            & \multicolumn{3}{c}{\textbf{Precision} ($\uparrow$)} & \multicolumn{3}{c}{\textbf{Recall} ($\uparrow$)} & \multicolumn{3}{c}{\textbf{F-score} ($\uparrow$)} \\
            \hline
            NeuralHDHair~\cite{wu2022neuralhdhair} & 38.30 & 54.20 & 61.76 & 54.50 & 80.77 & 89.65 & 44.99 & 64.87 & 73.14 \\
            HairStep~\cite{zheng2023hairstep} & 33.40 & 54.09 & 63.08 & 36.42 & 60.63 & 71.23 & 34.84 & 57.17 & 66.91 \\
            Ours & \textbf{89.68} & \textbf{96.70} & \textbf{98.24} & \textbf{86.47} & \textbf{93.82} & \textbf{96.44} & \textbf{88.05} & \textbf{95.24} & \textbf{97.33} \\
        \end{tabular}
    \end{minipage}

    \vspace{0mm}
    
    \begin{minipage}{0.12\linewidth}
        \includegraphics[width=\linewidth]{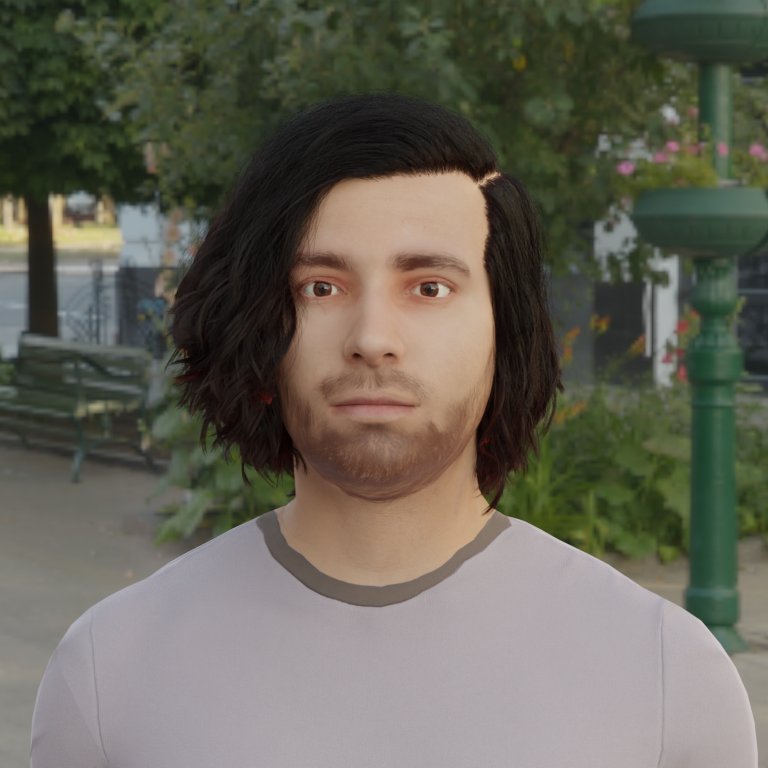} 
    \end{minipage}
    \hfill 
    \begin{minipage}{0.8\textwidth}
        \scriptsize 
        \begin{tabular}{l rrr | rrr | rrr}
            \setlength{\tabcolsep}{0pt}
            \renewcommand{\arraystretch}{3}
            & \multicolumn{9}{c}{\textbf{Thresholds: mm} $/$ \textbf{degrees}} \\
            \textbf{Method} & $2/20$ & $3/30$ & $4/40$ & $2/20$ & $3/30$ & $4/40$ & $2/20$ & $3/30$ & $4/40$ \\
            \cline{2-10}
            & \multicolumn{3}{c}{\textbf{Precision} ($\uparrow$)} & \multicolumn{3}{c}{\textbf{Recall} ($\uparrow$)} & \multicolumn{3}{c}{\textbf{F-score} ($\uparrow$)} \\
            \hline
            NeuralHDHair~\cite{wu2022neuralhdhair} & 31.27 & 62.32 & 80.14 & 32.20 & 60.04 & 77.37 & 31.73 & 61.16 & 78.73 \\
            HairStep~\cite{zheng2023hairstep} & 41.17 & 61.83 & 71.11 & 38.43 & 66.04 & 81.06 & 39.76 & 63.87 & 75.76 \\
            Ours & \textbf{63.71} & \textbf{84.26} & \textbf{92.67} & \textbf{62.97} & \textbf{84.77} & \textbf{93.34} & \textbf{63.34} & \textbf{84.51} & \textbf{93.00} \\
        \end{tabular}
    \end{minipage}

    \vspace{0mm}
    \begin{minipage}{0.12\linewidth}
        \includegraphics[width=\linewidth]{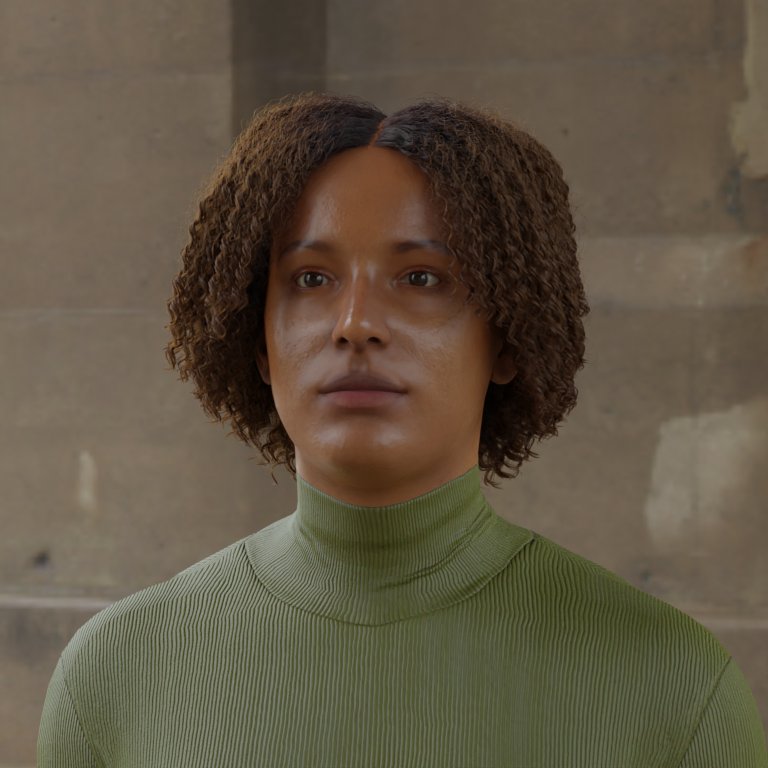} 
    \end{minipage}
    \hfill 
    \begin{minipage}{0.8\textwidth}
        \scriptsize 
        \begin{tabular}{l rrr | rrr | rrr}
            \setlength{\tabcolsep}{0pt}
            \renewcommand{\arraystretch}{3}
            & \multicolumn{9}{c}{\textbf{Thresholds: mm} $/$ \textbf{degrees}} \\
            \textbf{Method} & $2/20$ & $3/30$ & $4/40$ & $2/20$ & $3/30$ & $4/40$ & $2/20$ & $3/30$ & $4/40$ \\
            \cline{2-10}
            & \multicolumn{3}{c}{\textbf{Precision} ($\uparrow$)} & \multicolumn{3}{c}{\textbf{Recall} ($\uparrow$)} & \multicolumn{3}{c}{\textbf{F-score} ($\uparrow$)} \\
            \hline
            NeuralHDHair~\cite{wu2022neuralhdhair} & 16.75 & 43.33 & 62.75 & 9.33 & 22.73 & 37.59 & 11.99 & 29.82 & 47.01 \\
            HairStep~\cite{zheng2023hairstep} & 26.50 & 60.19 & 79.40 & 10.14 & 23.01 & 37.34 & 14.67 & 33.29 & 50.80 \\
            Ours & \textbf{29.53} & \textbf{62.11} & \textbf{81.67} & \textbf{34.39} & \textbf{69.56} & \textbf{87.98} & \textbf{31.77} & \textbf{65.62} & \textbf{84.71} \\
        \end{tabular}
    \end{minipage}

    \vspace{0mm}
    \begin{minipage}{0.12\linewidth}
        \includegraphics[width=\linewidth]{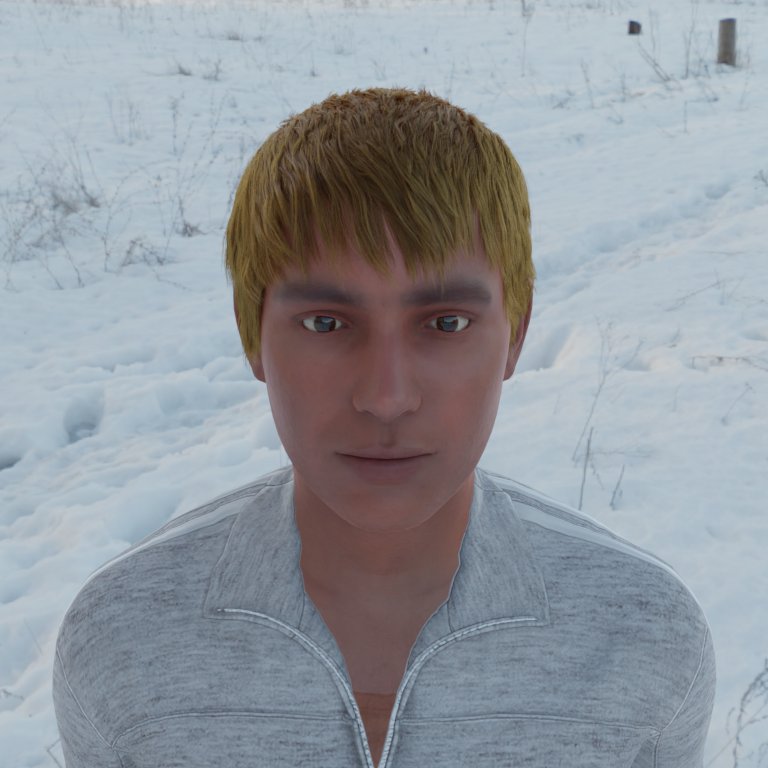} 
    \end{minipage}
    \hfill 
    \begin{minipage}{0.8\textwidth}
        \scriptsize 
        \begin{tabular}{l rrr | rrr | rrr}
            \setlength{\tabcolsep}{0pt}
            \renewcommand{\arraystretch}{3}
            & \multicolumn{9}{c}{\textbf{Thresholds: mm} $/$ \textbf{degrees}} \\
            \textbf{Method} & $2/20$ & $3/30$ & $4/40$ & $2/20$ & $3/30$ & $4/40$ & $2/20$ & $3/30$ & $4/40$ \\
            \cline{2-10}
            & \multicolumn{3}{c}{\textbf{Precision} ($\uparrow$)} & \multicolumn{3}{c}{\textbf{Recall} ($\uparrow$)} & \multicolumn{3}{c}{\textbf{F-score} ($\uparrow$)} \\
            \hline
            NeuralHDHair~\cite{wu2022neuralhdhair} & 16.83 & 31.60 & 43.07 & 20.15 & 42.45 & 62.18 & 18.34 & 36.23 & 50.89 \\
            HairStep~\cite{zheng2023hairstep} & 28.87 & 42.79 & 50.31 & 38.85 & 64.23 & 78.16 & 33.13 & 51.37 & 61.21 \\
            Ours & \textbf{85.73} & \textbf{96.97} & \textbf{99.05} & \textbf{84.34} & \textbf{96.37} & \textbf{98.70} & \textbf{85.03} & \textbf{96.67} & \textbf{98.87  } \\
        \end{tabular}
    \end{minipage}

    \vspace{0mm}
    \begin{minipage}{0.12\linewidth}
        \includegraphics[width=\linewidth]{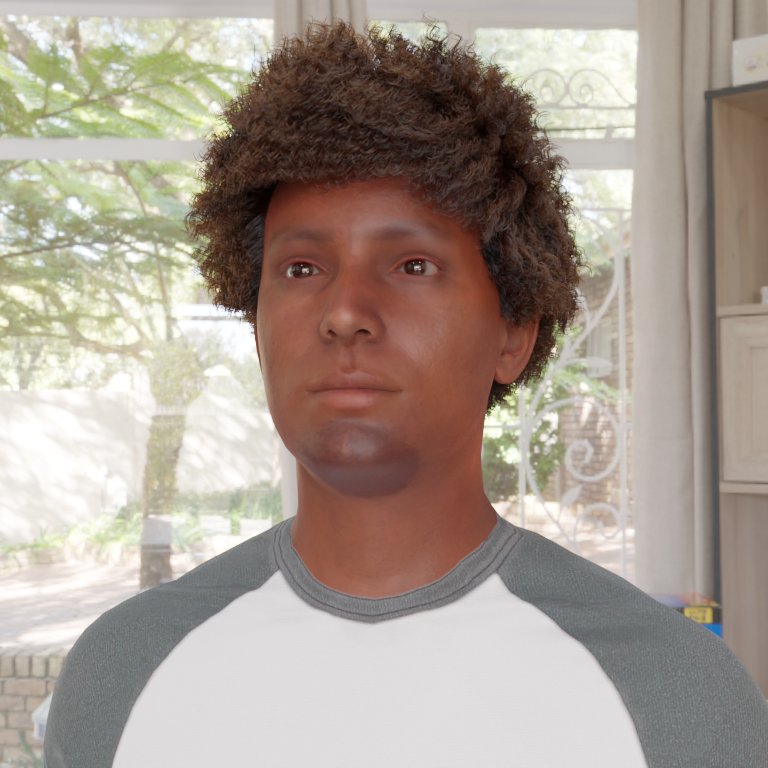} 
    \end{minipage}
    \hfill 
    \begin{minipage}{0.8\textwidth}
        \scriptsize 
        \begin{tabular}{l rrr | rrr | rrr}
            \setlength{\tabcolsep}{0pt}
            \renewcommand{\arraystretch}{3}
            & \multicolumn{9}{c}{\textbf{Thresholds: mm} $/$ \textbf{degrees}} \\
            \textbf{Method} & $2/20$ & $3/30$ & $4/40$ & $2/20$ & $3/30$ & $4/40$ & $2/20$ & $3/30$ & $4/40$ \\
            \cline{2-10}
            & \multicolumn{3}{c}{\textbf{Precision} ($\uparrow$)} & \multicolumn{3}{c}{\textbf{Recall} ($\uparrow$)} & \multicolumn{3}{c}{\textbf{F-score} ($\uparrow$)} \\
            \hline
            NeuralHDHair~\cite{wu2022neuralhdhair} & 10.60 & 31.98 & 52.94 & 5.45 & 17.77 & 32.98 & 7.19 & 22.84 & 40.64 \\
            HairStep~\cite{zheng2023hairstep} & 6.56 & 21.95 & 40.96 & 3.83 & 10.15 & 19.50 & 4.84 & 13.88 & 26.42 \\
            Ours & \textbf{38.33} & \textbf{71.01} & \textbf{88.46} & \textbf{34.37} & \textbf{65.80} & \textbf{83.83} & \textbf{36.24} & \textbf{68.30} & \textbf{86.08} \\
        \end{tabular}
    \end{minipage}

    \vspace{0mm}
    
    \begin{minipage}{0.12\linewidth}
        \includegraphics[width=\linewidth]{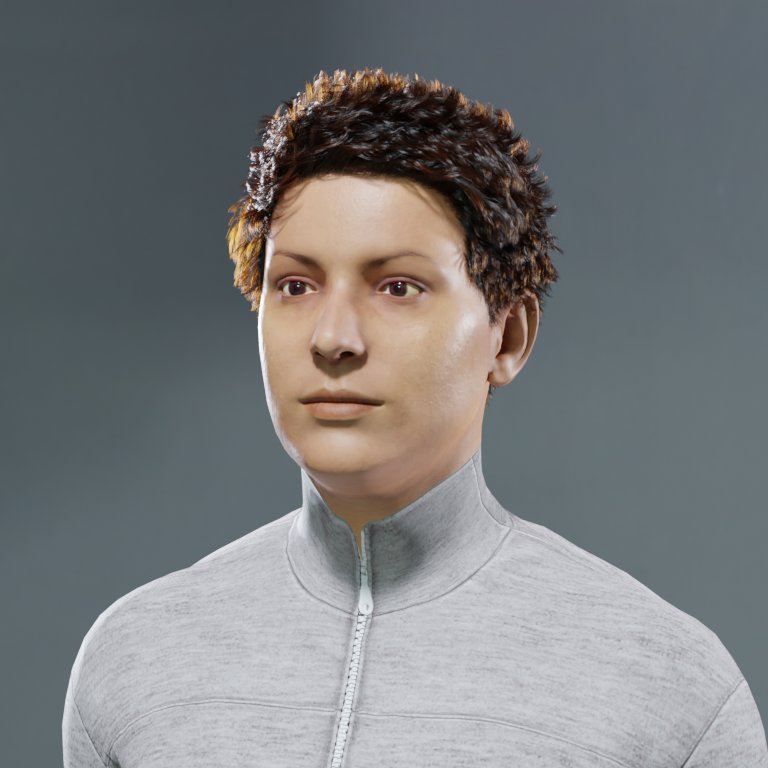} 
    \end{minipage}
    \hfill 
    \begin{minipage}{0.8\textwidth}
        \scriptsize 
        \begin{tabular}{l rrr | rrr | rrr}
            \setlength{\tabcolsep}{0pt}
            \renewcommand{\arraystretch}{3}
            & \multicolumn{9}{c}{\textbf{Thresholds: mm} $/$ \textbf{degrees}} \\
            \textbf{Method} & $2/20$ & $3/30$ & $4/40$ & $2/20$ & $3/30$ & $4/40$ & $2/20$ & $3/30$ & $4/40$ \\
            \cline{2-10}
            & \multicolumn{3}{c}{\textbf{Precision} ($\uparrow$)} & \multicolumn{3}{c}{\textbf{Recall} ($\uparrow$)} & \multicolumn{3}{c}{\textbf{F-score} ($\uparrow$)} \\
            \hline
            NeuralHDHair~\cite{wu2022neuralhdhair} & 9.90 & 24.95 & 40.50 & 10.70 & 24.12 & 38.74 & 10.29 & 24.53 & 39.60 \\
            HairStep~\cite{zheng2023hairstep} & 9.78 & 18.98 & 26.92 & 14.08 & 29.42 & 45.88 & 11.54 & 23.07 & 33.93 \\
            Ours & \textbf{54.17} & \textbf{81.83} & \textbf{92.75} & \textbf{51.87} & \textbf{78.94} & \textbf{90.60} & \textbf{53.00} & \textbf{80.36} & \textbf{91.67} \\
        \end{tabular}
    \end{minipage}

    \vspace{0mm}
    \begin{minipage}{0.12\linewidth}
        \includegraphics[width=\linewidth]{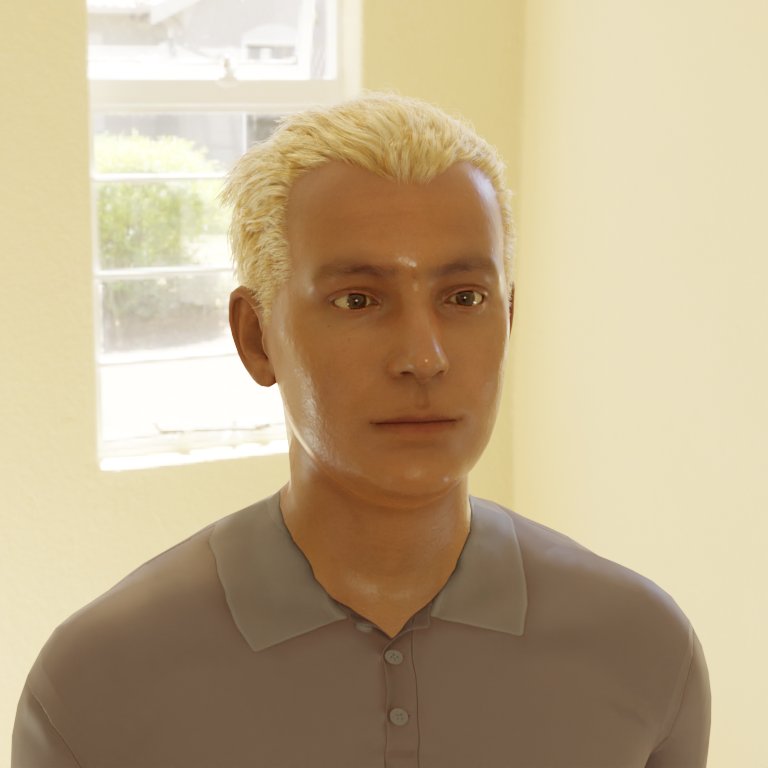} 
    \end{minipage}
    \hfill 
    \begin{minipage}{0.8\textwidth}
        \scriptsize 
        \begin{tabular}{l rrr | rrr | rrr}
            \setlength{\tabcolsep}{0pt}
            \renewcommand{\arraystretch}{3}
            & \multicolumn{9}{c}{\textbf{Thresholds: mm} $/$ \textbf{degrees}} \\
            \textbf{Method} & $2/20$ & $3/30$ & $4/40$ & $2/20$ & $3/30$ & $4/40$ & $2/20$ & $3/30$ & $4/40$ \\
            \cline{2-10}
            & \multicolumn{3}{c}{\textbf{Precision} ($\uparrow$)} & \multicolumn{3}{c}{\textbf{Recall} ($\uparrow$)} & \multicolumn{3}{c}{\textbf{F-score} ($\uparrow$)} \\
            \hline
            NeuralHDHair~\cite{wu2022neuralhdhair} & 17.64 & 31.10 & 41.55 & 21.04 & 43.40 & 59.30 & 19.19 & 36.24 & 48.87 \\
            HairStep~\cite{zheng2023hairstep} & 11.90 & 18.39 & 24.29 & 17.26 & 33.64 & 50.13 & 14.09 & 23.78 & 32.7 \\
            Ours & \textbf{85.82} & \textbf{97.15} & \textbf{85.82} & \textbf{73.10} & \textbf{86.27} & \textbf{90.94} & \textbf{78.95} & \textbf{91.39} & \textbf{94.85} \\
        \end{tabular}
    \end{minipage}

    \vspace{0mm}

    \begin{minipage}{0.12\linewidth}
        \includegraphics[width=\linewidth]{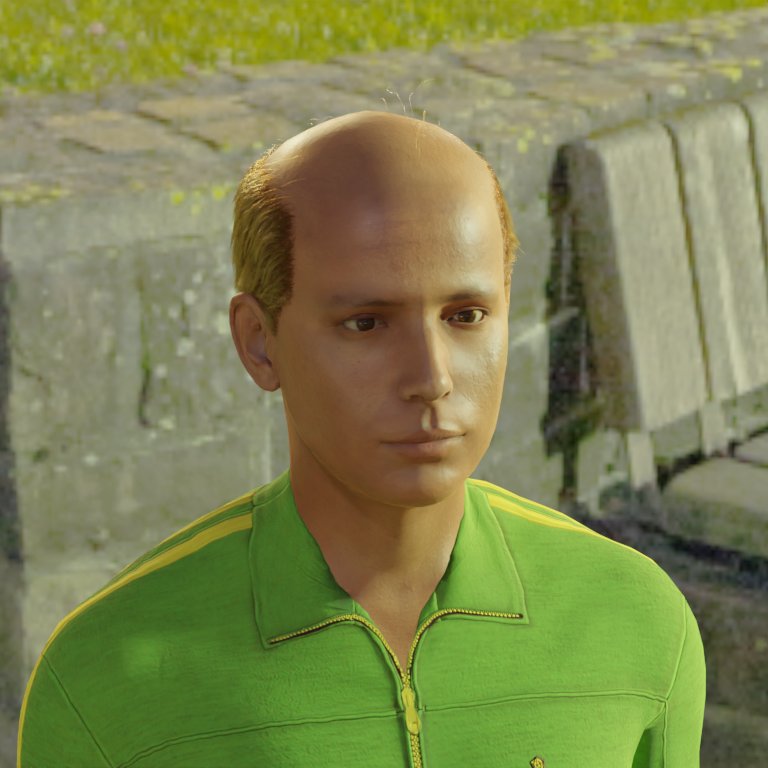} 
    \end{minipage}
    \hfill 
    \begin{minipage}{0.8\textwidth}
        \scriptsize 
        \begin{tabular}{l rrr | rrr | rrr}
            \setlength{\tabcolsep}{0pt}
            \renewcommand{\arraystretch}{3}
            & \multicolumn{9}{c}{\textbf{Thresholds: mm} $/$ \textbf{degrees}} \\
            \textbf{Method} & $2/20$ & $3/30$ & $4/40$ & $2/20$ & $3/30$ & $4/40$ & $2/20$ & $3/30$ & $4/40$ \\
            \cline{2-10}
            & \multicolumn{3}{c}{\textbf{Precision} ($\uparrow$)} & \multicolumn{3}{c}{\textbf{Recall} ($\uparrow$)} & \multicolumn{3}{c}{\textbf{F-score} ($\uparrow$)} \\
            \hline
            NeuralHDHair~\cite{wu2022neuralhdhair} & 11.34 & 21.98 & 34.32 & 8.76 & 19.74 & 27.98 & 9.89 & 20.80 & 30.82 \\
            HairStep~\cite{zheng2023hairstep} & 7.69 & 15.41 & 21.67 & 21.96 & 44.94 & 65.67 & 11.39 & 22.95 & 32.58 \\
            Ours & \textbf{70.03} & \textbf{91.91} & \textbf{97.29} & \textbf{55.16} & \textbf{77.37} & \textbf{87.19} & \textbf{61.71} & \textbf{84.02} & \textbf{91.96} \\
        \end{tabular}
    \end{minipage}

    \vspace{0mm}
    \begin{minipage}{0.12\linewidth}
        \includegraphics[width=\linewidth]{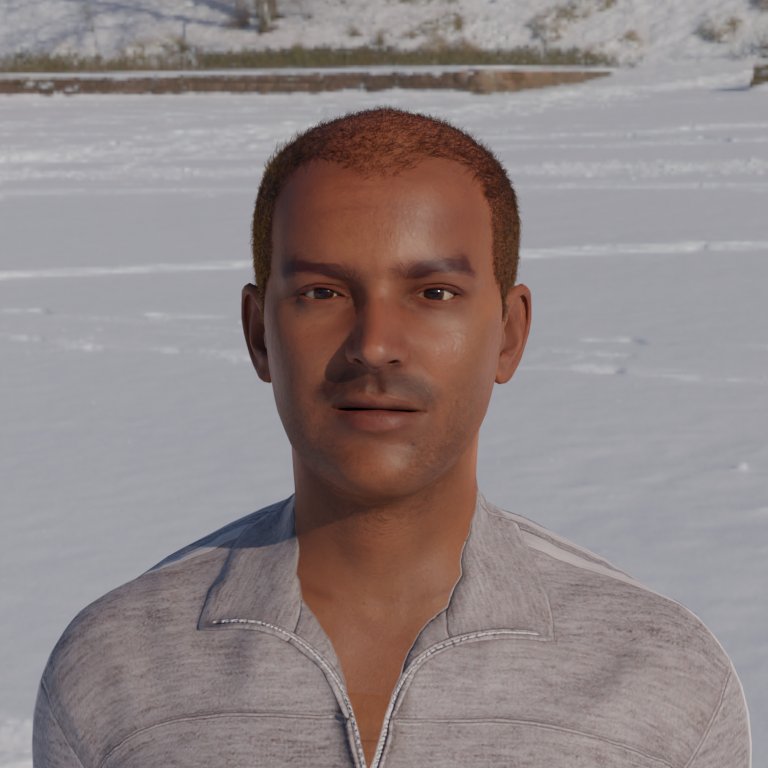} 
    \end{minipage}
    \hfill 
    \begin{minipage}{0.8\textwidth}
        \scriptsize 
        \begin{tabular}{l rrr | rrr | rrr}
            \setlength{\tabcolsep}{0pt}
            \renewcommand{\arraystretch}{3}
            & \multicolumn{9}{c}{\textbf{Thresholds: mm} $/$ \textbf{degrees}} \\
            \textbf{Method} & $2/20$ & $3/30$ & $4/40$ & $2/20$ & $3/30$ & $4/40$ & $2/20$ & $3/30$ & $4/40$ \\
            \cline{2-10}
            & \multicolumn{3}{c}{\textbf{Precision} ($\uparrow$)} & \multicolumn{3}{c}{\textbf{Recall} ($\uparrow$)} & \multicolumn{3}{c}{\textbf{F-score} ($\uparrow$)} \\
            \hline
            NeuralHDHair~\cite{wu2022neuralhdhair} & 6.38 & 13.38 & 20.47 & 7.85 & 17.39 & 27.23 & 7.04 & 15.12 & 23.37 \\
            HairStep~\cite{zheng2023hairstep} & 4.29 & 9.02 & 13.72 & 7.02 & 17.73 & 31.59 & 5.32 & 11.96 & 19.13 \\
            Ours & \textbf{63.73} & \textbf{86.23} & \textbf{94.16} & \textbf{44.22} & \textbf{67.34} & \textbf{80.17} & \textbf{52.21} & \textbf{75.63} & \textbf{86.60} \\
        \end{tabular}
    \end{minipage}
    \end{minipage}

    \caption{\textbf{Quantitative comparison.} We provide the quantitative comparison for each example in DiffLocks evaluation set. When the hairsytle is curly or balding or the image is not in frontal view, our method achieves significant improvement.}
    \label{fig:image-quantitative}
\end{figure*}

\paragraph{Quantitative.} We provide extended quantitative comparisons of HairStep and NeuralHDHair on both DiffLocks evaluation dataset and Yuksel dataset~\cite{yuksel2009hair}. We calculate precision, recall and F-score using 3D ground-truth strands similar to previous methods~\cite{wu2024monohair,sklyarova2023neural_haircut} as shown in~\cref{tab:quantitative_suppmat} and~\cref{fig:image-quantitative}. We provide per-example results to complement the aggregate quantitative metrics presented in the main paper. 
Since HairStep and NeuralHDHair are trained primarily on frontal views, and their training hairstyles lack diversity, they tend to perform poorly on the DiffLocks dataset which contains a range of hairstyles(curly, balding, combed-back, and afro-like) together with images that deviate slightly from the frontal view.

Considering that the baselines models were trained trained on the USC-HairSalon dataset~\cite{hu2015single}, while our training data aligns more closely with the distribution of the evaluation dataset (rendering manner and strands distribution on scalp), our results will be better aligned with the distribution of ground truth, leading to higher evaluation metrics. Thus, we also perform a quantitative evaluation on Yuksel synthetic dataset~\cite{yuksel2009hair}.
We show the metrics for the different hairstyles separately for a more comprehensive analysis.
HairStep and NeuralHDHair both achieved high F-score in straight hairsytle, but for curly hairstyle, their method struggle to reconstruct it accurately and both precision, recall and F-score are greatly reduced. 
In contract, our method still performs well on curly hairstyle, recovering plausible geometry even on the backside of the head.

\subsection{Additional in-the-wild results}
\begin{figure*}[t]

	\centering
	\begin{tikzpicture}[ remember picture, >={Stealth[inset=1pt,length=8pt,angle'=30,round]} ]

     \def\xshift{2.9}
     \def\yshift{-3.5}
     \def\w{0.18}
     \def\firstshift{-0.2}

     \def\cropBottom{0.1cm}

    \node[] (origin) at (0,0) {};

    \definetrim{cooper_4_crop}{0.0 0.2 0.0 0.0}

    \node[inner sep=0pt, anchor=south] (cooper_4_rgb) at (\firstshift+\xshift*0, 0) {\adjincludegraphics[width=\w\textwidth, trim={{0.1\width} {0.1\height} {0.1\width} 0}, clip]{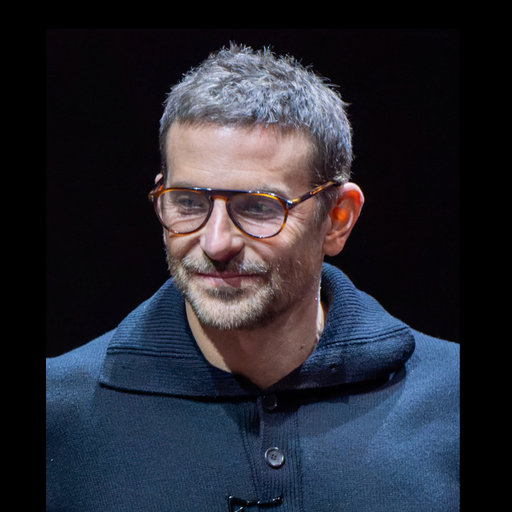}};
    \node[inner sep=0pt, anchor=south] (cooper_4_front) at (\xshift*1, 0) {\adjincludegraphics[width=\w\textwidth, trim={{0.1\width} {0.1\height} {0.1\width} 0}, clip]{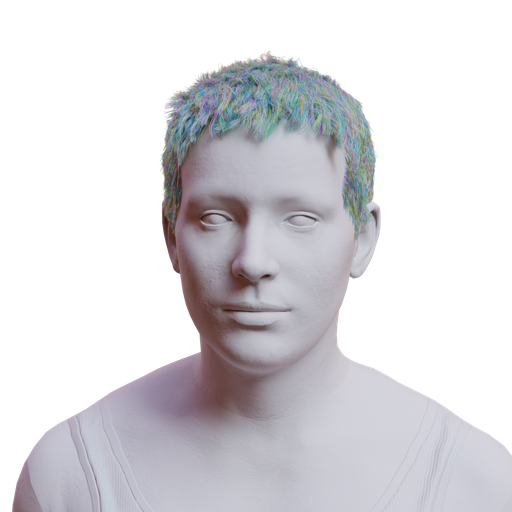}};
    \node[inner sep=0pt, anchor=south] (cooper_4_left) at (\xshift*2, 0) {\adjincludegraphics[width=\w\textwidth, trim={{0.1\width} {0.1\height} {0.1\width} 0}, clip]{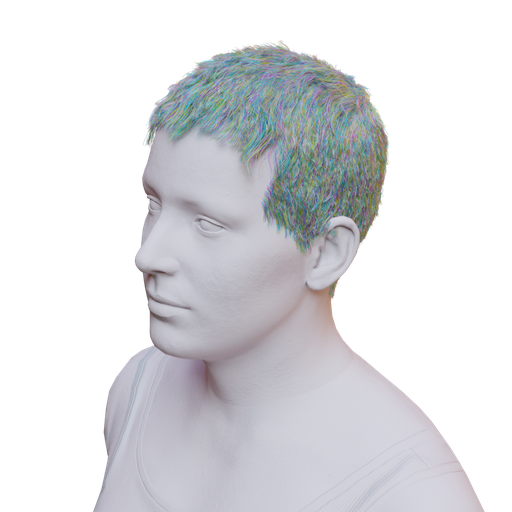}};
    \node[inner sep=0pt, anchor=south] (cooper_4_back) at (\xshift*3, 0) {\adjincludegraphics[width=\w\textwidth, trim={{0.1\width} {0.1\height} {0.1\width} 0}, clip]{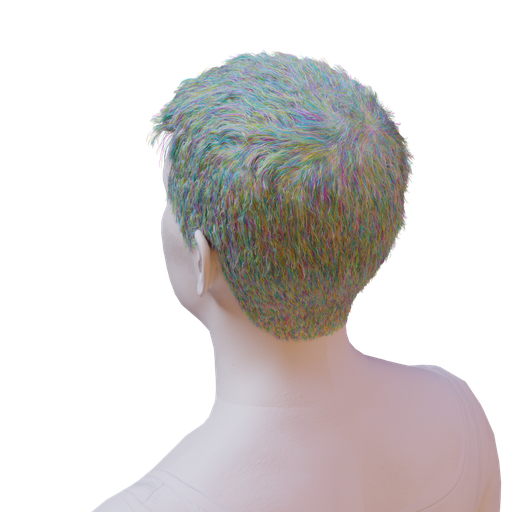}};

    \node[inner sep=0pt, anchor=south] (portman_3_rgb) at (\firstshift+\xshift*0, \yshift) {\adjincludegraphics[width=\w\textwidth, trim={{0.1\width} {0.1\height} {0.1\width} 0}, clip]{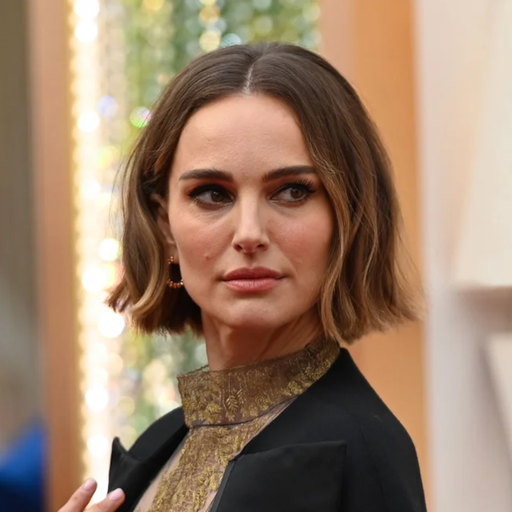}};
    \node[inner sep=0pt, anchor=south] (portman_3_front) at (\xshift*1, \yshift) {\adjincludegraphics[width=\w\textwidth, trim={{0.1\width} {0.1\height} {0.1\width} 0}, clip]{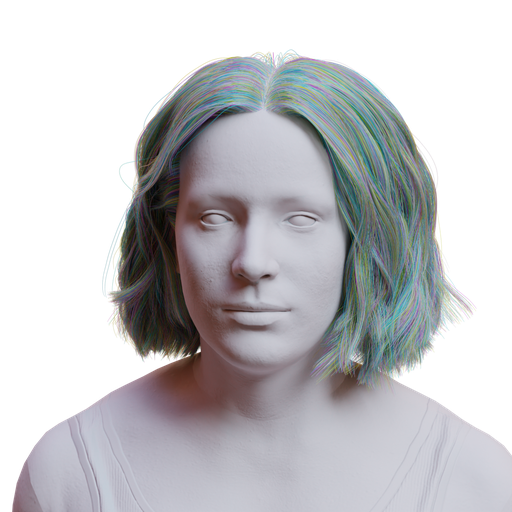}};
    \node[inner sep=0pt, anchor=south] (portman_3_front) at (\xshift*2, \yshift) {\adjincludegraphics[width=\w\textwidth, trim={{0.1\width} {0.1\height} {0.1\width} 0}, clip]{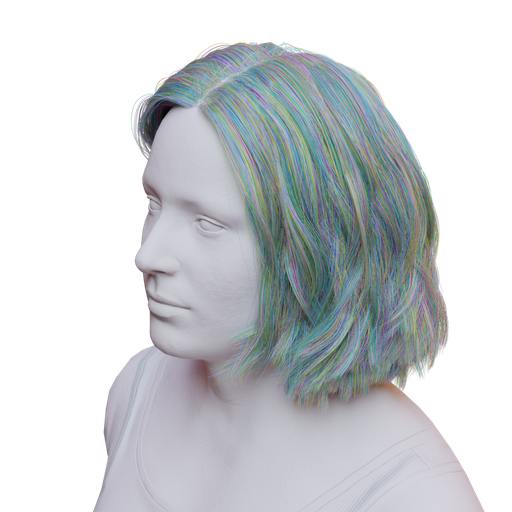}};
    \node[inner sep=0pt, anchor=south] (portman_3_front) at (\xshift*3, \yshift) {\adjincludegraphics[width=\w\textwidth, trim={{0.1\width} {0.1\height} {0.1\width} 0}, clip]{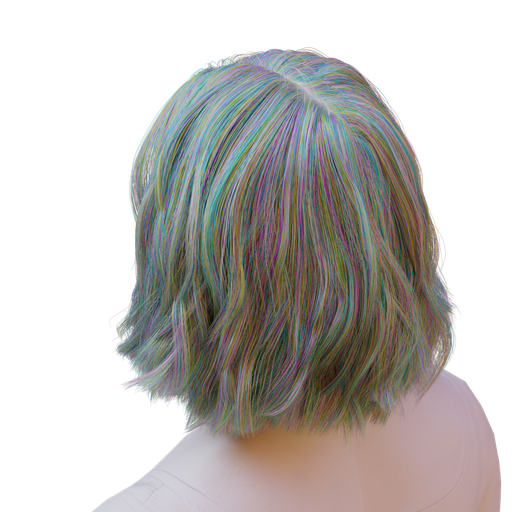}};
    \node[inner sep=0pt, anchor=south] (olsen_1_rgb) at (\firstshift+\xshift*0, \yshift*2+0.2) {\adjincludegraphics[width=\w\textwidth, trim={{0.05\width} {0.1\height} {0.05\width} 0}, clip]{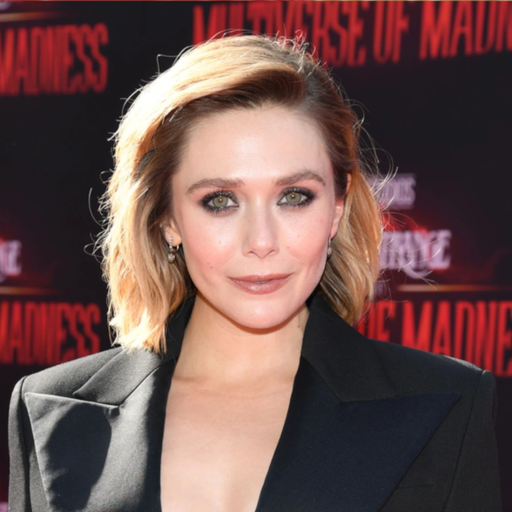}};
    \node[inner sep=0pt, anchor=south] (olsen_1_front) at (\xshift*1, \yshift*2+0.2) {\adjincludegraphics[width=\w\textwidth, trim={{0.05\width} {0.1\height} {0.05\width} 0}, clip]{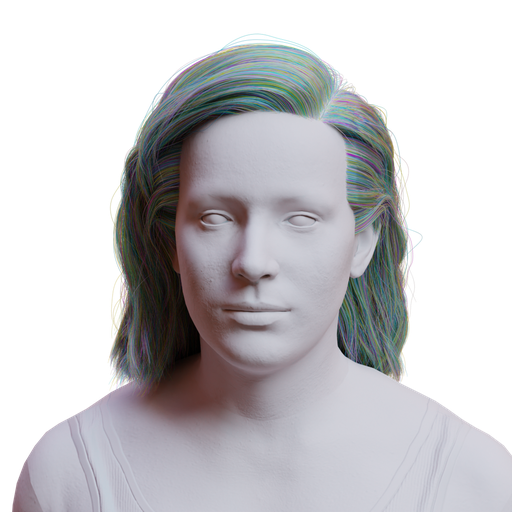}};
    \node[inner sep=0pt, anchor=south] (olsen_1_left) at (\xshift*2, \yshift*2) {\adjincludegraphics[width=\w\textwidth, trim={{0.05\width} {0.1\height} {0.05\width} 0}, clip]{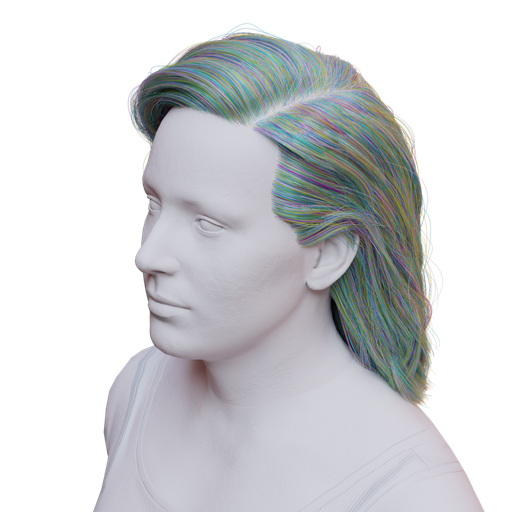}};
    \node[inner sep=0pt, anchor=south] (olsen_1_back) at (\xshift*3, \yshift*2+0.2) {\adjincludegraphics[width=\w\textwidth, trim={{0.05\width} {0.1\height} {0.05\width} 0}, clip]{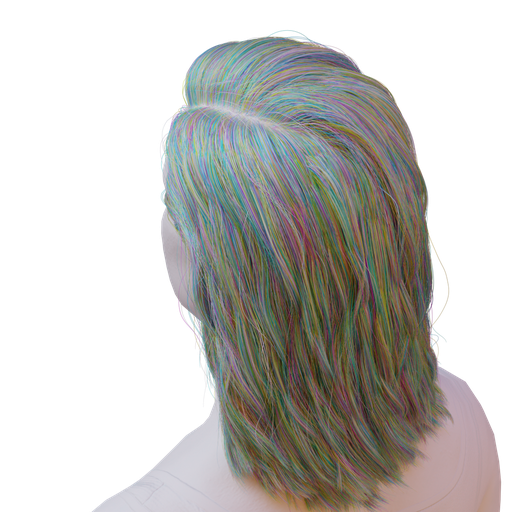}};

    \node[inner sep=0pt, anchor=south] (olsen_2_rgb) at (\firstshift+\xshift*0, \yshift*3) {\adjincludegraphics[width=\w\textwidth, trim={{0.05\width} {0.0\height} {0.05\width} 0}, clip]{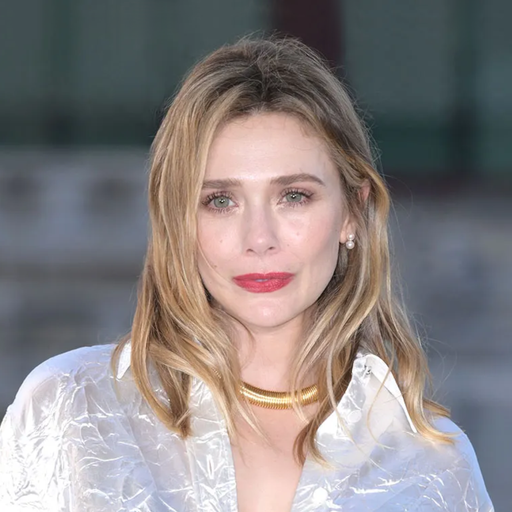}};
    \node[inner sep=0pt, anchor=south] (olsen_2_front) at (\xshift*1, \yshift*3) {\adjincludegraphics[width=\w\textwidth, trim={{0.05\width} {0.0\height} {0.05\width} 0}, clip]{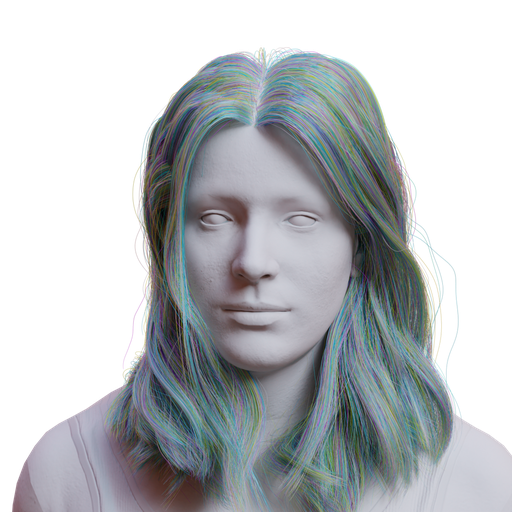}};
    \node[inner sep=0pt, anchor=south] (olsen_2_left) at (\xshift*2, \yshift*3) {\adjincludegraphics[width=\w\textwidth, trim={{0.05\width} {0.0\height} {0.05\width} 0}, clip]{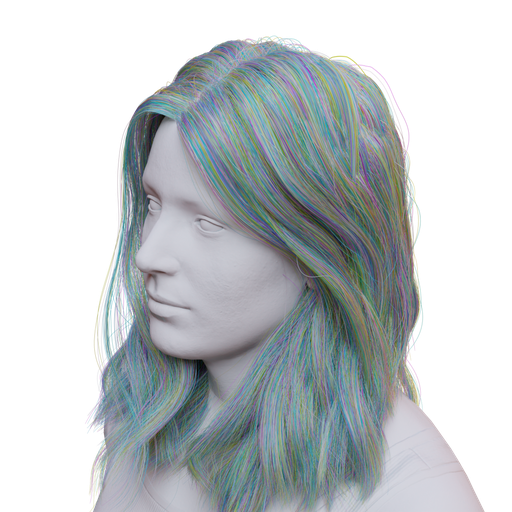}};
    \node[inner sep=0pt, anchor=south] (olsen_2_back) at (\xshift*3, \yshift*3) {\adjincludegraphics[width=\w\textwidth, trim={{0.05\width} {0.0\height} {0.05\width} 0}, clip]{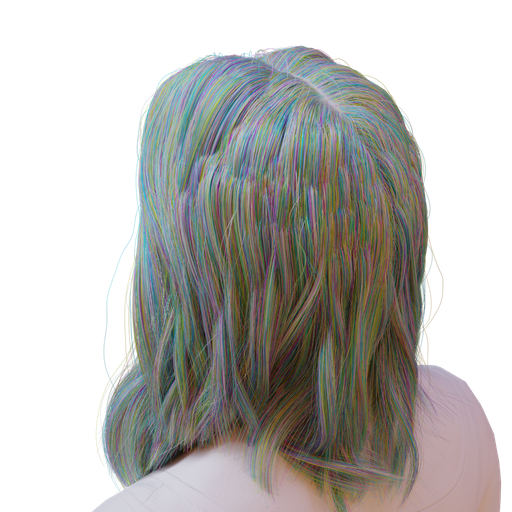}};

    \node[inner sep=0pt, anchor=south] (stone_1_rgb) at (\firstshift+\xshift*0, \yshift*4) {\adjincludegraphics[width=\w\textwidth, trim={{0.05\width} {0.1\height} {0.05\width} 0}, clip]{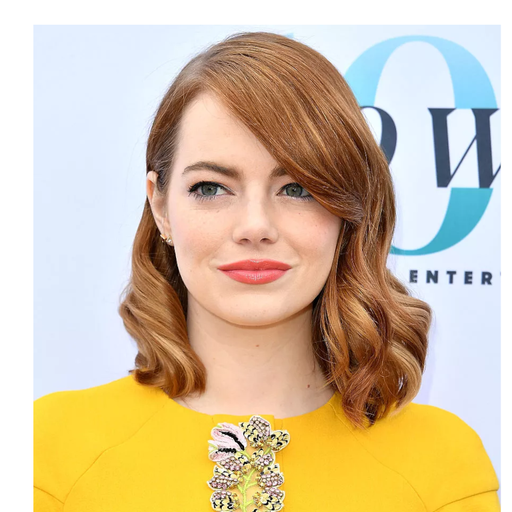}};
    \node[inner sep=0pt, anchor=south] (stone_1_front) at (\xshift*1, \yshift*4) {\adjincludegraphics[width=\w\textwidth, trim={{0.05\width} {0.1\height} {0.05\width} 0}, clip]{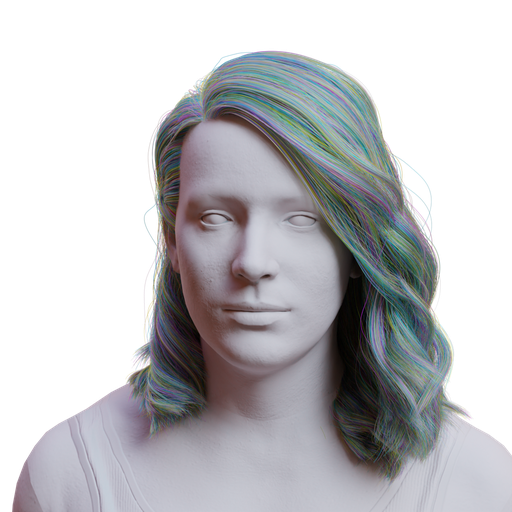}};
    \node[inner sep=0pt, anchor=south] (stone_1_left) at (\xshift*2, \yshift*4) {\adjincludegraphics[width=\w\textwidth, trim={{0.05\width} {0.1\height} {0.05\width} 0}, clip]{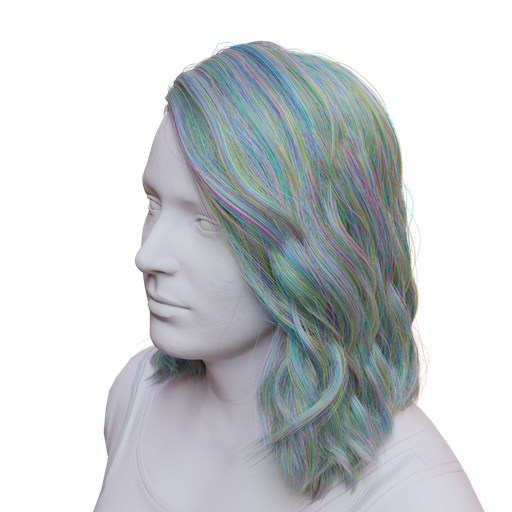}};
    \node[inner sep=0pt, anchor=south] (stone_1_back) at (\xshift*3, \yshift*4) {\adjincludegraphics[width=\w\textwidth, trim={{0.05\width} {0.1\height} {0.05\width} 0}, clip]{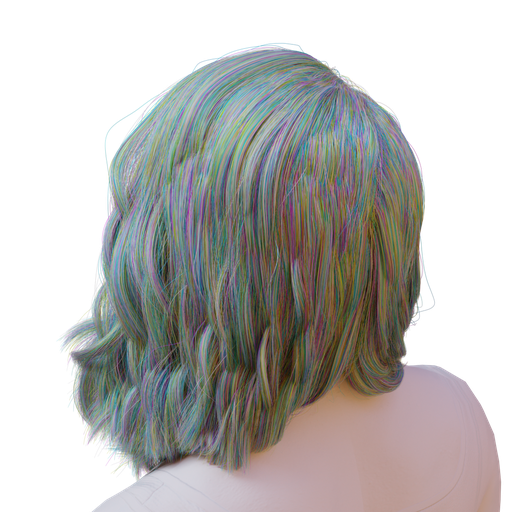}};

    \node[inner sep=0pt, anchor=south] (medium_6_rgb) at (\firstshift+\xshift*0, \yshift*5) {\adjincludegraphics[width=\w\textwidth, trim={{0.05\width} {0.1\height} {0.05\width} 0}, clip]{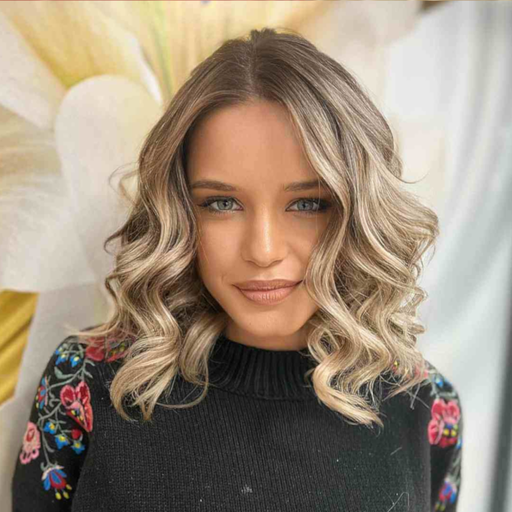}};
    \node[inner sep=0pt, anchor=south] (medium_6_front) at (\xshift*1, \yshift*5) {\adjincludegraphics[width=\w\textwidth, trim={{0.05\width} {0.1\height} {0.05\width} 0}, clip]{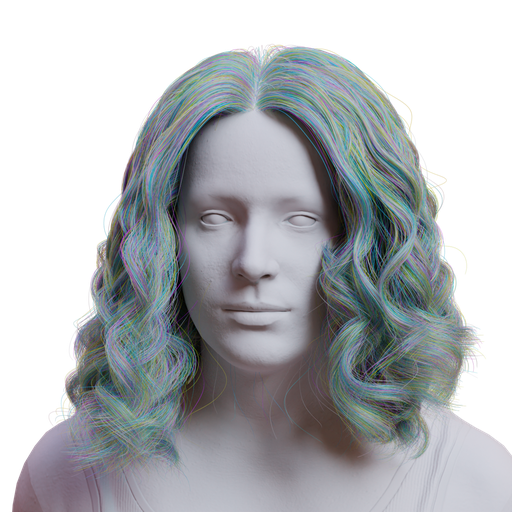}};
    \node[inner sep=0pt, anchor=south] (medium_6_left) at (\xshift*2, \yshift*5) {\adjincludegraphics[width=\w\textwidth, trim={{0.05\width} {0.1\height} {0.05\width} 0}, clip]{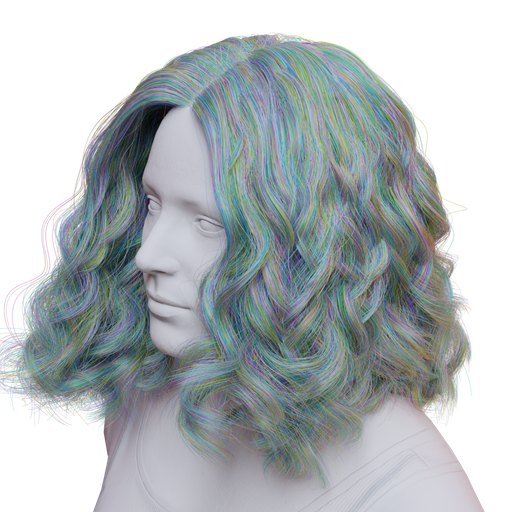}};
    \node[inner sep=0pt, anchor=south] (medium_6_back) at (\xshift*3, \yshift*5) {\adjincludegraphics[width=\w\textwidth, trim={{0.05\width} {0.1\height} {0.05\width} 0}, clip]{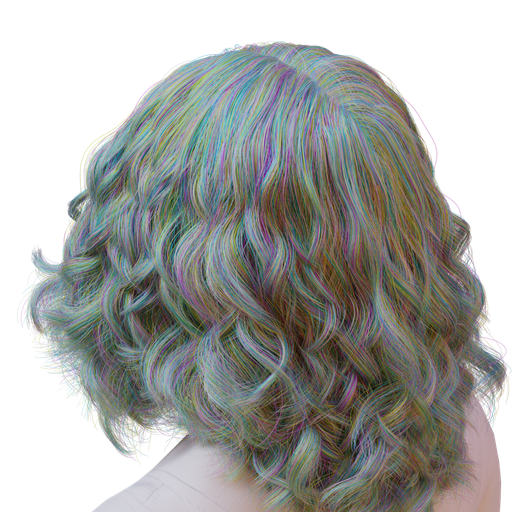}};



    \node[font=\footnotesize\selectfont, align=left, below = -0.0cm of medium_6_rgb] {RGB input};
    \node[font=\footnotesize\selectfont, align=left, below = -0.0cm of medium_6_front] {Front};
    \node[font=\footnotesize\selectfont, align=left, below = -0.0cm of medium_6_left] {Side};
    \node[font=\footnotesize\selectfont, align=left, below = -0.0cm of medium_6_back] {Back};

	\end{tikzpicture}
	\vspace{-0.in}
	\caption{More results of in-the-wild reconstruciton of hairstyles.
      }
    \vspace{-0.0in}
	\label{fig:qualitative_results_suppl}
\end{figure*}

Lastly, to evaluate the robustness and effectiveness of our method, we provide additional reconstructions from in-the-wild single images. As shown in~\cref{fig:qualitative_results_suppl}, our method can robustly reconstruct a large variety of hairstyles, achieving high-quality and realistic results.

\begin{figure*}
    \centering
    \resizebox{0.95\linewidth}{!}{%
    \begin{tabular}{l rrr | rrr | rrr || rrr | rrr | rrr}
        \setlength{\tabcolsep}{0pt}
        & \multicolumn{9}{c}{\textbf{Straight}} & \multicolumn{9}{c}{\textbf{Curly}} \\
        \textbf{Method} & $2 / 20$ & $3 / 30$ & $4 / 40$ & $2 / 20$ & $3 / 30$ & $4 / 40$ & $2 / 20$ & $3 / 30$ & $4 / 40$ & $2 / 20$ & $3 / 30$ & $4 / 40$ & $2 / 20$ & $3 / 30$ & $4 / 40$ & $2 / 20$ & $3 / 30$ & $4 / 40$ \\
        \cline{2-19}
        & \multicolumn{3}{c}{\textbf{Precision}} & \multicolumn{3}{c}{\textbf{Recall}} & \multicolumn{3}{c}{\textbf{F-score}}& \multicolumn{3}{c}{\textbf{Precision}} & \multicolumn{3}{c}{\textbf{Recall}} & \multicolumn{3}{c}{\textbf{F-score}} \\
        \hline

        NeuralHDHair~\cite{wu2022neuralhdhair}	&	\textbf{72.58}	&	\textbf{86.85}	&	\textbf{92.07}	&	45.28	&	63.57	&	74.55	&	55.77	&	73.41	&	82.39 &	23.24	&	43.29	&	55.45	&	17.92	&	38.65	&	54.73	&	20.23	&	40.84	&	55.09 \\
        
 HairStep~\cite{zheng2023hairstep}	&	64.05	&	78.95	&	84.52	&	\textbf{56.22}	&	\textbf{73.37}	&	\textbf{82.17}	&	\textbf{59.88}	&	\textbf{76.06}	&	\textbf{83.33} &	17.71	&	33.02	&	43.67	&	8.11	&	16.55	&	24.14	&	11.13	&	22.05	&	31.08 \\
  Ours	&	57.38	&	76.15	&	84.49	&	38.54	&	54.84	&	65.74	&	46.11	&	63.77	&	73.94 &	\textbf{29.83}	&	\textbf{60.46}	&	\textbf{79.09}	&	\textbf{31.41}	&	\textbf{61.69}	&	\textbf{77.84}	&	\textbf{30.60}	&	\textbf{61.07}	&	\textbf{78.46} \\

    \end{tabular} 
    }
    \captionof{table}{Quantitative comparison with \cite{wu2022neuralhdhair,zheng2023hairstep} on Yuksel dataset~\cite{yuksel2009hair}. We evaluate our method on straight and curly hair separately.  Our method achieves superior results on curly hair. Our method performs slightly worse on straight hair due to the diffusion model, which introduces perturbations to enhance the realism of the generated hairstyles.}
    \label{tab:quantitative_suppmat}
\end{figure*}

\end{document}